%% file: 00_main_arxiv.tex
\documentclass[10pt,twocolumn,letterpaper]{article}

\usepackage{iccv}
\usepackage{times}
\usepackage{epsfig}
\usepackage{graphicx}
\usepackage{amsmath}
\usepackage{amssymb}
\usepackage{subcaption}
\usepackage{afterpage}
\usepackage{floatpag}

\usepackage{booktabs}
\usepackage[dvipsnames]{xcolor}
\usepackage{multirow}
\usepackage{pifont}
\usepackage{comment}
\usepackage{caption}
\usepackage{authblk}
\usepackage[symbol]{footmisc}

\usepackage[page]{appendix} 

\newcommand{\cmark}{\ding{51}}%
\newcommand{\xmark}{\ding{55}}%
\newcommand{\greencmark}{{\color{Green}\cmark}}
\newcommand{\redxmark}{{\color{red}\xmark}}

\usepackage[pagebackref=true,breaklinks=true,letterpaper=true,colorlinks,bookmarks=false]{hyperref}

\iccvfinalcopy 

\def\iccvPaperID{9268} 

\input{arxiv/constants}

\begin{document}

\title{\datasetname: A High-Fidelity Dataset of 3D Indoor Scenes}
\renewcommand*{\Authsep}{~~~~~}
\renewcommand*{\Authand}{~~~~~}
\renewcommand*{\Authands}{~~~~~}

\author[]{Chandan Yeshwanth$^\ast$}
\author[]{Yueh-Cheng Liu$^\ast$}
\author[]{Matthias Nie{\ss}ner}
\author[]{Angela Dai}

\affil[]{Technical University of Munich}

\twocolumn[{%
\renewcommand\twocolumn[1][]{#1}%
\maketitle

    \begin{center}
            \vspace{-0.7cm}
            \captionsetup{type=figure}
            \includegraphics[width=0.9\textwidth]{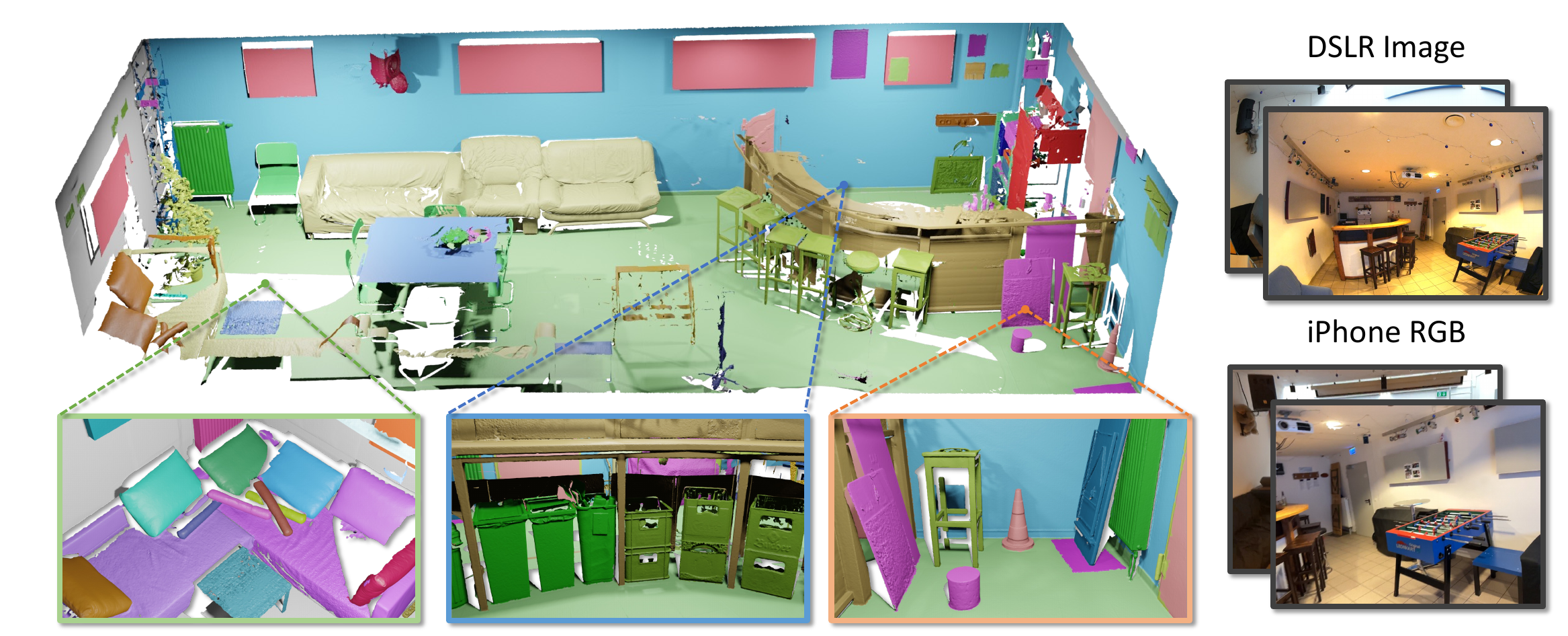}
            \vspace{-0.1cm}
            \captionof{figure}{\OURS{} contains \numscenes{} high-resolution 3D reconstructions of indoor scenes with dense semantic and instance annotations, along with corresponding high-quality DSLR images and iPhone RGB-D sequences. The long-tail and multi-labeled annotations enable fine-grained semantic understanding, while the high-quality and commodity RGB images enable benchmarking of novel view synthesis methods at scale.}
    \end{center}
}]

\begin{abstract}
\vspace{-0.4cm}

We present \OURS{}, a large-scale dataset that couples together capture of high-quality and commodity-level geometry and color of indoor scenes.
Each scene is captured with a high-end laser scanner at sub-millimeter resolution, along with registered \DSLRResMP{}-megapixel images from a DSLR camera, and RGB-D streams from an iPhone.
Scene reconstructions are further annotated with an open vocabulary of semantics, with label-ambiguous scenarios explicitly annotated for comprehensive semantic understanding. 
\OURS{} enables a new real-world benchmark for novel view synthesis, both from high-quality RGB capture, and importantly also from commodity-level images, in addition to a new benchmark for 3D semantic scene understanding that comprehensively encapsulates diverse and ambiguous semantic labeling scenarios.
Currently, \OURS{} contains \numscenes{} scenes, \numDSLRImages{} captured DSLR images, and over \numiPhoneFrames{} iPhone RGBD frames.

\end{abstract}

\footnotetext[1]{Equal contribution.}
\footnotetext[2]{Project page: \url{https://cy94.github.io/scannetpp/}}

\input{arxiv/01_introduction}
\input{arxiv/02_related_work}

\input{arxiv/03_method}
\input{arxiv/04_experiment}
\input{arxiv/05_conclusion}

\section*{Acknowledgements}
This work was supported by the Bavarian State Ministry of Science and the Arts and coordinated by the Bavarian Research Institute for Digital Transformation (bidt),   the German Research Foundation (DFG) Grant ``Learning How to Interact with Scenes through Part-Based Understanding,'' the ERC Starting Grant Scan2CAD (804724), the German Research Foundation (DFG) Grant ``Making Machine Learning on Static and Dynamic 3D Data Practical,'' and the German Research Foundation (DFG) Research Unit ``Learning and Simulation in Visual Computing.''
We thank Ben Mildenhall for helpful discussions and advice on NeRF capture.

{\small
\bibliographystyle{ieee_fullname}
\bibliography{00_main_arxiv}
}
\clearpage

\renewcommand{\appendixpagename}{\Large{Appendix}} 

\begin{appendices}
In this supplemental material we provide further details of data collection in Section \ref{sec:app-data-collection}, description of the NVS and semantic understanding benchmarks in Section \ref{sec:app-benchmark}, details of semantic annotation and qualitative results of baselines in Section \ref{sec:app-semantic-understanding} and qualitative results of NVS on iPhone data in Section \ref{sec:app-nvs-iphone}.
\input{arxiv/06_appendix}
\end{appendices}

\end{document}

%% file: arxiv/constants.tex
\newcommand{\OURS}{ScanNet++}
\newcommand{\datasetname}{\OURS}
\newcommand{\numscenes}{460}
\newcommand{\DSLRResMP}{33}

\newcommand{\iphoneDurationHours}{17.4}
\newcommand{\numiPhoneFrames}{3.7M}

\newcommand{\numDSLRImages}{280,000}

\newcommand{\totalFloorArea}{15,000}

\newcommand{\numTrainScenes}{360}
\newcommand{\numValScenes}{50}
\newcommand{\numTestScenes}{50}

\newcommand{\avgScansPerScene}{4.85}
\newcommand{\totalScans}{1858}

%% file: arxiv/01_introduction.tex
\section{Introduction}
\label{sec:intro}

\begin{table*}[thb!]
  \vspace{2mm}
  \centering
  \setlength{\tabcolsep}{3 pt}
  \resizebox{0.8\textwidth}{!}{
\begin{tabular}{ c |  c  |  c |  c | c  c | c c  | c c  }
\hline
 \multirow{2}{*}{Dataset} & \multirow{2}{*}{Num. scenes}  & Average & \multirow{2}{*}{Total scans} & \multicolumn{2}{|c|}{RGB (MP)} & \multicolumn{2}{|c|}{Depth} &  \multirow{2}{*}{Dense semantics} & \multirow{2}{*}{NVS}  \\
 
 & & {scans/scene} & & Commodity & DSLR & LR & HR  & &  \\
 \hline 
 LLFF \cite{mildenhall2019llff} & 35 & -  & - & 12 & \redxmark & \redxmark & \redxmark & \redxmark & \greencmark \\
 DTU \cite{jensen2014large} & 124 & - & - & 1.9 & \redxmark & \redxmark & \greencmark & \redxmark & \greencmark  \\
 BlendedMVS \cite{yao2020blendedmvs} & 113 & -  & - & \redxmark & \redxmark & \redxmark & \redxmark & \greencmark & \redxmark  \\
 ScanNet \cite{dai2017scannet} & 1503  & - & -& 1.25  & \redxmark & \greencmark & \redxmark & \greencmark & \redxmark  \\
 Matterport3D \cite{chang2017matterport3d} & 90\footnotemark[3]   & - &- & 1.3  & \redxmark & \greencmark & \redxmark & \greencmark & \redxmark  \\
 Tanks and Temples \cite{knapitsch2017tanks} &  21  & 10.57 &  74 &  \redxmark & 8  & \redxmark & \greencmark &\redxmark & \greencmark  \\
 ETH3D \cite{eth3d} & 25 & 2.33 &  42 &  \redxmark & 24 & \redxmark & \greencmark & \redxmark & \redxmark   \\
 ARKitScenes \cite{baruch2021arkitscenes} &  1004 &  3.16 &  3179 &  3  & \redxmark & \greencmark & \greencmark &  \redxmark   & \redxmark  \\ 
 \textit{\datasetname{}} (ours) &  \numscenes{} &  \avgScansPerScene{} & \totalScans{} & 2.7  & 33  & \greencmark & \greencmark &  \greencmark & \greencmark \\\hline
\end{tabular}
  }
  \caption{Comparison of datasets in terms of RGB and geometry. \OURS{} surpasses existing datasets in terms of resolution, quality, density of semantic annotations, and coverage of laser scans. While the quality of the reconstructed geometry in ARKitScenes is similar to ours, we additionally capture DSLR data to support the novel view synthesis task (NVS) and provide dense semantic annotations.
}
  \label{tab:comparison}
\end{table*}
Reconstruction and understanding of 3D scenes is fundamental to many applications in computer vision, including robotics, autonomous driving, mixed reality and content creation, among others.
The last several years have seen a revolution in representing and reconstructing 3D scenes with groundbreaking networks such as neural radiance fields (NeRFs) \cite{mildenhall2020nerf}.
NeRFs optimize complex scene representations from an input set of posed RGB images with a continuous volumetric scene function to enable synthesis of novel image views, with recent works achieving improved efficiency, speed, and scene regularization \cite{sun2022direct,fridovich2022plenoxels,yu2021plenoctrees,martin2021nerf,liu2020neural,barron2021mip,barron2022mip,chen2022tensorf,muller2022instant}.
Recent works have even extended the photometric-based formulation to further optimize scene semantics based on 2D semantic signal from the input RGB images \cite{zhi2021place,vora2021nesf,fu2022panoptic,kundu2022panoptic,siddiqui2022panoptic}.

Notably, such radiance field scene representations focus on individual per-scene optimization, without learning generalized priors for view synthesis.
This is due to the lack of large-scale datasets which would support learning such general priors.
As shown in Table~\ref{tab:comparison}, existing datasets either contain a large quantity of scenes that lack high-quality color and geometry capture, or contain a very limited number of scenes with high-quality color and geometry.
We propose to bridge this divide with \OURS{}, a large-scale dataset that contains both high-quality color and geometry capture coupled with commodity-level data of indoor environments. 
We hope that this inspires future work on generalizable novel view synthesis with semantic priors.

\OURS{} contains \numscenes{} scenes covering a total floor area of \totalFloorArea{}$m^2$, with each scene captured by a Faro Focus Premium laser scanner at sub-millimeter resolution with an average distance of 0.9$mm$ between points in a scan, DSLR camera images at \DSLRResMP{}-megapixels, and RGB-D video from an iPhone 13 Pro.
All sensor modalities are registered together to enable seamless interaction between geometric and color modalities, as well as commodity-level and high-end data capture.
Furthermore, as semantic understanding and reconstruction can be seen as interdependent, each captured scene is additionally densely annotated with its semantic instances.
Since semantic labeling can be ambiguous in many scenarios, we collect annotations that are both open-vocabulary and explicitly label semantically ambiguous instances, with more than 1000 unique classes annotated.

\OURS{} thus supports new benchmarks for novel view synthesis and 3D semantic scene understanding, enabling evaluation against precise real-world ground truth not previously available. 
This enables comprehensive, quantitative evaluation of state-of-the-art methods with a general and fair evaluation across a diversity of scene scenarios, opening avenues for new improvement.

For novel view synthesis, we also introduce a new task of view synthesis from commodity sensor data to match that of high-quality DSLR ground truth capture, which we believe will push existing methodologies to their limits.
In contrast to existing 3D semantic scene understanding benchmarks, we explicitly take into account label ambiguities for more accurate, comprehensive semantics.

\medskip
To summarize, our main contributions are:
\begin{itemize}
\vspace{-0.1cm}
\item We present a new large-scale and high-resolution indoor dataset  with 3D reconstructions, high-quality RGB images, commodity RGB-D video, and semantic annotations covering label ambiguities.
\vspace{-0.1cm}
\item Our dataset enables optimizing and benchmarking novel view synthesis on large-scale real-world scenes from both high-quality DSLR and commodity-level iPhone images.  Instead of sampling from the scanning trajectory for testing ground-truth images, we provide a more challenging setting where testing images are captured independently from the scanning trajectory. 
\vspace{-0.1cm}

\item 
Our 3D semantic data enables training and benchmarking a comprehensive view of semantic understanding that handles possible label ambiguities inherent to semantic labeling tasks.

\end{itemize}

\footnotetext[3]{90 buildings.}

%% file: arxiv/02_related_work.tex
\section{Related Work}
Deep learning methods for  3D semantic understanding and novel view synthesis require large-scale, diverse datasets to generalize. We review existing datasets proposed for both tasks and compare them with \OURS{}.

\subsection{Semantic Understanding of 3D Indoor Scenes}
Early datasets for 3D semantic understanding, such as NYUv2~\cite{silberman2011indoor} and SUN RGB-D~\cite{song2015sun}, comprise short RGB-D sequences with low resolution and limited annotations. ScanNet~\cite{dai2017scannet} was the first dataset to provide 3D reconstructions and annotations at scale, consisting of 1503 RGB-D sequences of 707 unique scenes recorded with an iPad mounted with a Structure sensor. Due to the lower-resolution commodity-level geometric capture, small objects and details are difficult to recognize and annotate.
More recently, the ScanNet200 benchmark~\cite{rozenberszki2022language} was proposed on top of the ScanNet dataset for recognition of 200 annotated classes. However, the performance on long-tail classes is also limited by the geometric resolution of ScanNet.
Similarly, Matterport3D~\cite{chang2017matterport3d} consists of low-resolution reconstructions from panoramic RGB-D images and semantic annotations.
ARKitScenes~\cite{baruch2021arkitscenes} improves upon these datasets in the resolution of ground truth geometry from laser scans. However, rather than dense semantic labels, ARKitScenes only provides bounding box annotations for only 17 object classes.

In comparison to these datasets, \OURS{} includes both high-resolution 3D geometry provided by the laser scanner and high-quality color capture, along with long-tail fine-grained semantic annotations with multi-labeling to disambiguate regions that may belong to multiple classes.

\subsection{Novel View Synthesis}
Novel view synthesis (NVS) methods have primarily been evaluated on outside-in and forward-facing images. The LLFF~\cite{mildenhall2019llff} dataset contains 35 handheld cellphone captures of small scenes with images sharing the same viewing direction (\ie, forward-facing). NeRF~\cite{mildenhall2020nerf} and its successors \cite{liu2020neural,barron2021mip} built synthetic datasets of object-centric, outside-in images.

Meanwhile, datasets that were originally proposed for multi-view stereo such as DTU~\cite{jensen2014large}, BlendedMVS~\cite{yao2020blendedmvs}, and Tanks and Temples~\cite{knapitsch2017tanks} are now also used for novel view synthesis. Although Tanks and Temples has high-quality RGB images, it only consists of 7 training scenes and 14 test scenes, lacking scale and diversity of scenes.

Since ScanNet~\cite{dai2017scannet} contains RGB-D scans of a wide variety of indoor scenes, some NeRF methods \cite{liu2020neural,xu2022point} also use it for NVS. However, the data is not ideal for novel view synthesis since it was captured with commodity iPad RGB cameras, hence suffering from  high motion blur and limited field-of-view. Additionally, as ScanNet is not designed for the NVS task, testing poses must be subsampled from the training camera trajectory, resulting in biased evaluation. 

In contrast to existing NVS datasets, \OURS{} provides higher-quality images for many diverse real-world scenes without constraints over the camera poses. 
Our testing images are captured independent of the camera poses for training, reflecting more practical scenarios. 
In addition to benchmarking NVS methods that are optimized on a per-scene level, we believe that the scale and diversity of \OURS{} will encourage research on NVS methods that generalize over multiple scenes \cite{yu2021pixelnerf, wang2021ibrnet, chen2021mvsnerf, johari2022geonerf, zhang2022nerfusion, suhail2022generalizable,liu2022neural}.

%% file: arxiv/03_method.tex
\begin{figure}[thb!]
\centering
\includegraphics[width=\linewidth]{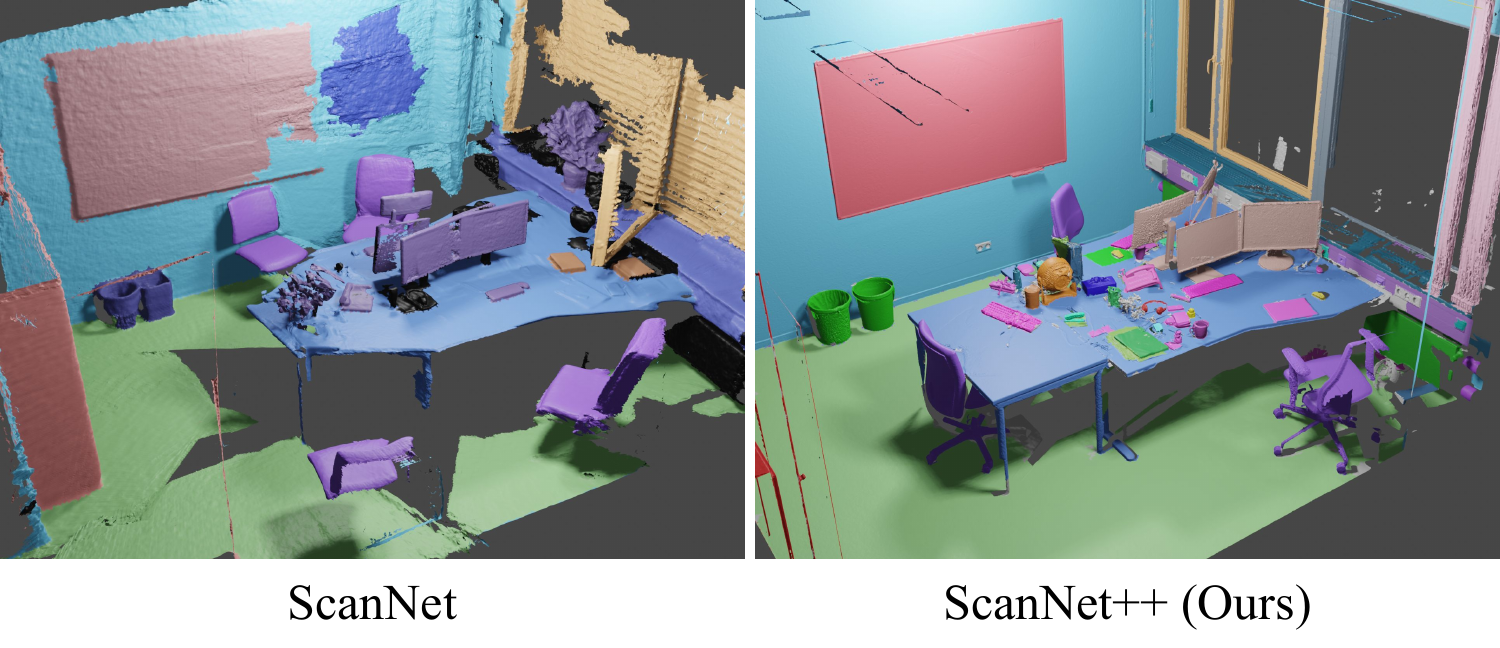}
\vspace{-6.5mm}
\caption{Comparison of a 3D reconstruction and semantic annotation on a scene from ScanNet~\cite{dai2017scannet} and a similar scene from \OURS{}.}
\label{fig:comparison_scannet}
\vspace{-5mm}
\end{figure}

\section{Data Acquisition and Processing}
We record three modalities of data for each scene using the laser scanner, DSLR camera, and iPhone RGB-D videos. The whole capture process takes around 30 minutes on average for a scene, and upwards of 2 hours for larger scenes. In the following, we discuss the capture process of each sensor.
\subsection{Laser Scans}
We acquire point clouds of the scenes using the Faro Focus Premium laser scanner. Each scan contains about 40 million points. We use multiple scanner positions per scene, 4 scans for a medium-sized room on average, and increase proportionately based on the size and complexity of the scene, in order to obtain maximum coverage of the surface of the scene.
We use Poisson reconstruction on the point clouds~\cite{kazhdan2006poisson,kazhdan2013screened} to produce mesh surface representations for each scene. 
For computational tractability, we run Poisson reconstruction \cite{kazhdan2006poisson,kazhdan2013screened} on overlapping chunks of points, trimming the resulting meshes of their overlap regions and merging them together. Finally, we use Quadric Edge Collapse~\cite{garland1997surface} to obtain a simplified mesh for the ease of visualization and annotation.

\subsection{DSLR Images}

\begin{figure}[htb!]
\centering
\includegraphics[width=0.8\linewidth]{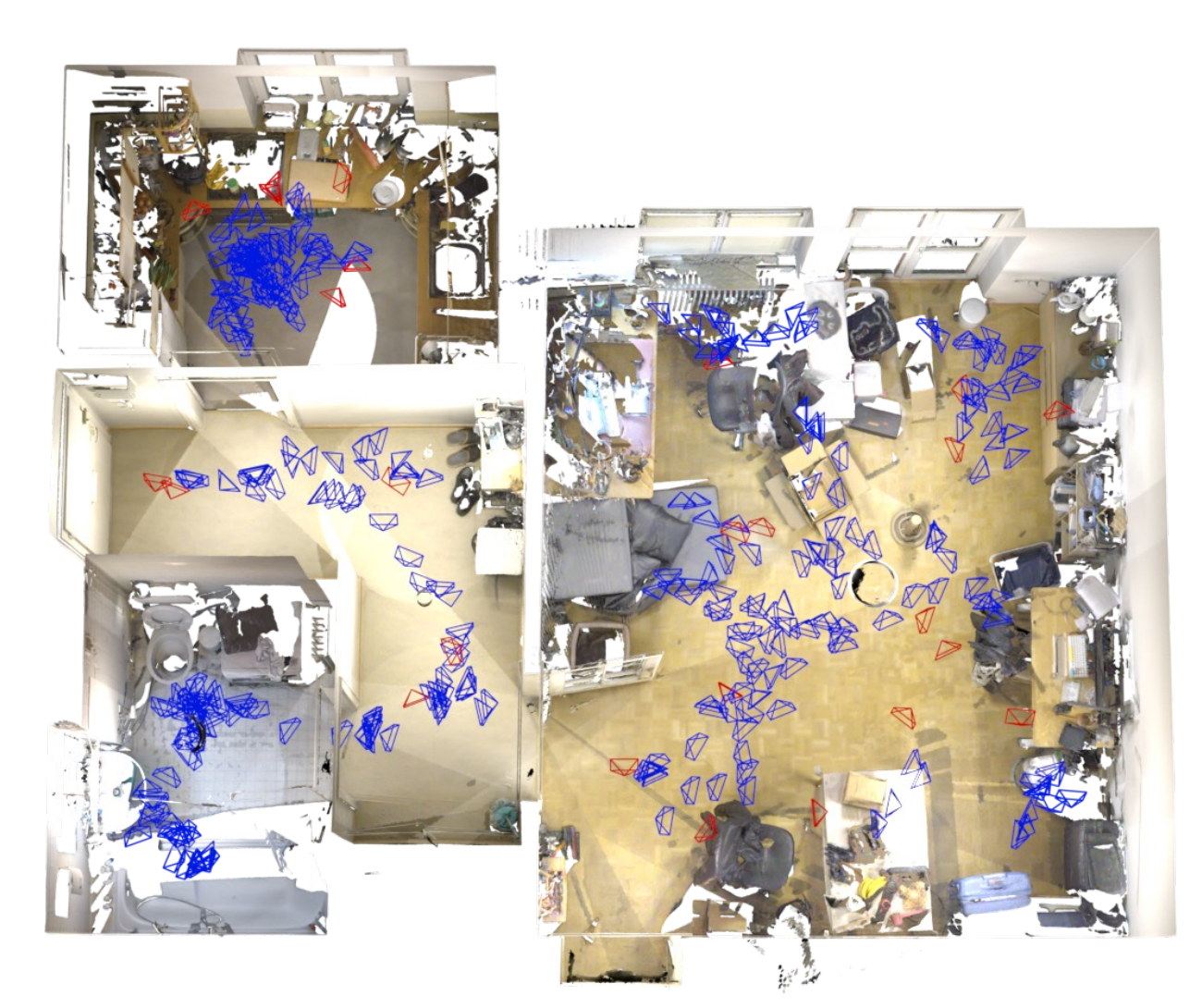}
\vspace{-2mm}
\caption{DSLR camera poses for novel view synthesis. Training poses (blue) form a continuous and dense trajectory at a standard capture height, while test poses (red) are outside this trajectory at varying heights and angles.}
\label{fig:nvs_poses}
\end{figure}

Novel view synthesis (NVS) works rely on photometric error as supervision. Hence, the ground truth data for NVS must have fixed lighting, wide field-of-view, and sharp images. Accordingly, we take static images with a Sony Alpha 7 IV DSLR camera with a fisheye lens. These wide-angle images are beneficial for registration of the images with each other to obtain camera poses, especially since indoor scenes can have large textureless regions (e.g., walls,  cupboards). For a medium-sized room, we capture around 200 images for training and scale up to proportionately for larger scenes. Instead of using held-out views that are subsampled from the camera trajectory for evaluation, we capture an additional set of 15-25 novel images per scene to obtain challenging, realistic testing images for novel view synthesis. An example of these poses is shown in Fig. \ref{fig:nvs_poses}.

Tab.~\ref{tab:dslrtraj} shows the average distances from train/test poses to the nearest train poses (excluding query pose) from ScanNet~\cite{dai2017scannet} and \OURS{}.
For ScanNet, we subsample held-out views as testing images from the camera trajectory, following Point-NeRF \cite{xu2022point} and NeRFusion \cite{zhang2022nerfusion}. Our train/test poses differ notably in both translation and orientation.

\begin{table}[thb!]
\centering
\footnotesize
\begin{tabular}{l | c | c | c }
\hline
Dataset & Split & Distance (m) & Rotation (deg.) \\
\hline
\multirow{2}{*}{ScanNet} & train & 0.04 & 3.25 \\
 & test & 0.04 &  3.09 \\
 \hline
\multirow{2}{*}{\OURS{}} & train & 0.07 & 7.21 \\
 & test & \textbf{0.40} & \textbf{42.69} \\
\hline
\end{tabular}
\caption{Distance to the nearest train camera pose. Evaluating with novel poses that have large translation and rotation difference makes \OURS{} more challenging for NVS compared to existing datasets like 
 ScanNet~\cite{dai2017scannet}.} \label{tab:dslrtraj}
\end{table}

\subsection{iPhone Images and LiDAR}

We capture the RGB and LiDAR depth stream provided by iPhone 13 Pro using a custom iOS application. Unlike the manually controlled DSLR scanning process, we use the default iPhone camera settings (auto white balance, auto exposure, and auto-focus) to reflect the most common capture scenarios. RGB images are captured at a resolution of 1920 $\times$ 1440, and LiDAR depth images at 256 $\times$ 192, both recorded at 60 FPS synchronously. For a medium-sized room, we record the RGB-D video for around two minutes, yielding \iphoneDurationHours{} hours of video in the whole dataset.

\begin{figure}[htb!]
\centering
\includegraphics[width=\linewidth]{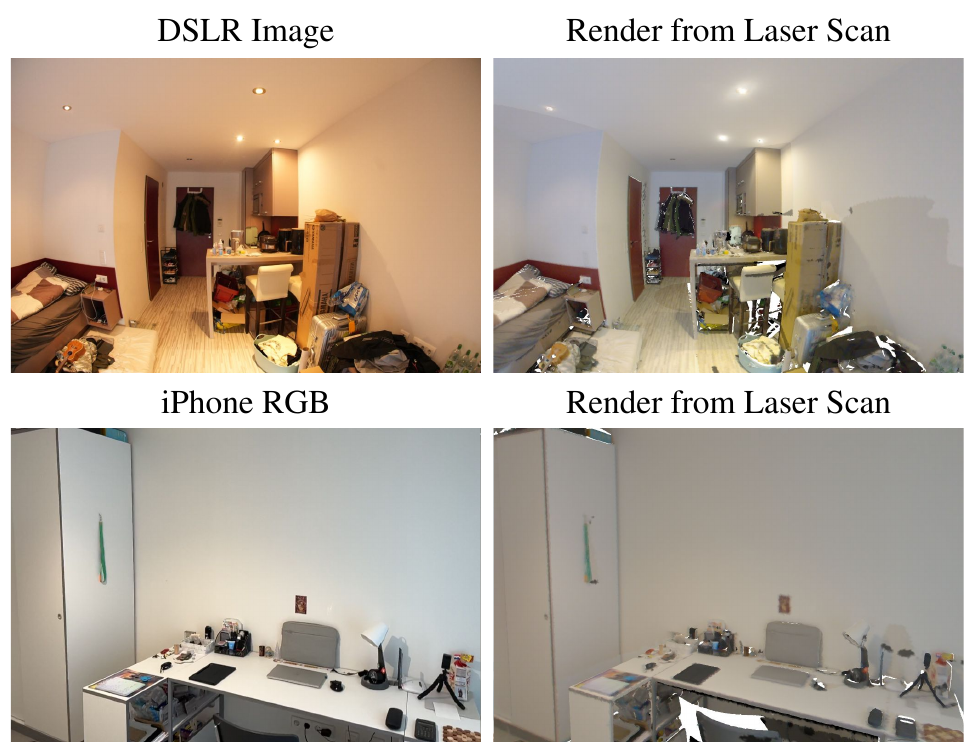}
\vspace{-6mm}
\caption{Examples of the alignment between DSLR, iPhone, and the scanner in \OURS{}. We obtain accurate alignment of all 3 sensors into the same coordinate system, empowering research across three modalities.}
\label{fig:align}
\vspace{-3mm}

\end{figure}

\begin{figure}[thb!]
\centering
\includegraphics[width=\linewidth]{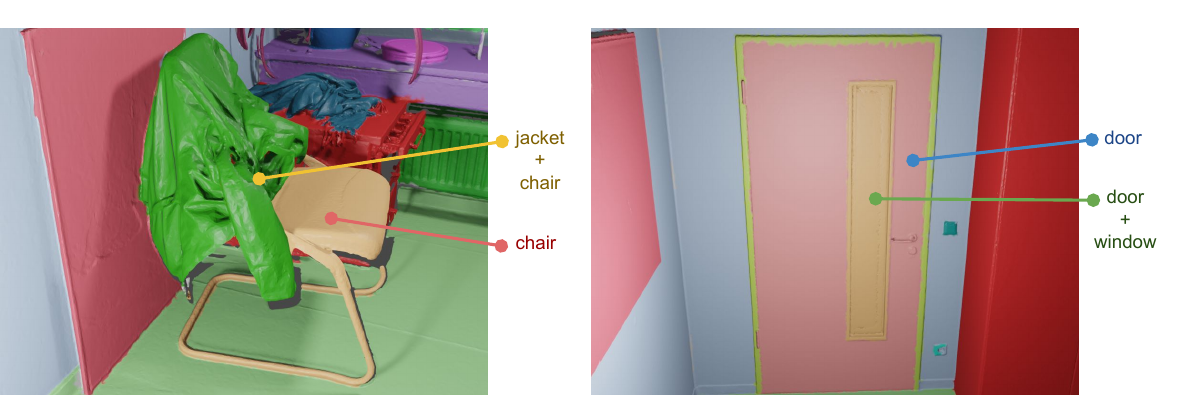}
\vspace{-6mm}
\caption{Examples of multi-label annotation. The part of the chair covered by the jacket is annotated as both jacket and chair. The small window in the door is annotated as both door and window.}
\label{fig:mutlilabel_example}
\vspace{-4mm}
\end{figure}

\subsection{Registration and Alignment}
We leverage COLMAP~\cite{schoenberger2016mvs,schoenberger2016sfm} to register the DSLR and iPhone images with the laser scan, obtaining poses for both sets of images in the same coordinate system as the scan. To do this, we first render pseudo images from the laser scan and include them in the COLMAP Structure-from-Motion (SfM) pipeline. Once the rendered images are registered with the real images, we can then transform the SfM poses into the same coordinate system as the laser scans and recover the metric scale. Additionally, we refine the camera poses with dense photometric error guided by the geometry of the laser scan \cite{zhou2014color,eth3d}. 
For iPhone images, we filter out iPhone frames as unreliably registered when the average difference between iPhone depth and rendered laser scan depth is  $>0.3$m. 
Examples of the obtained DSLR and iPhone alignment are shown in Fig.~\ref{fig:align}.

\subsection{Semantic Annotation}
\begin{figure*}[t]
\vspace{-0.3cm}
\centering
\includegraphics[width=\textwidth, trim={0.0cm 12cm 0 0},clip]{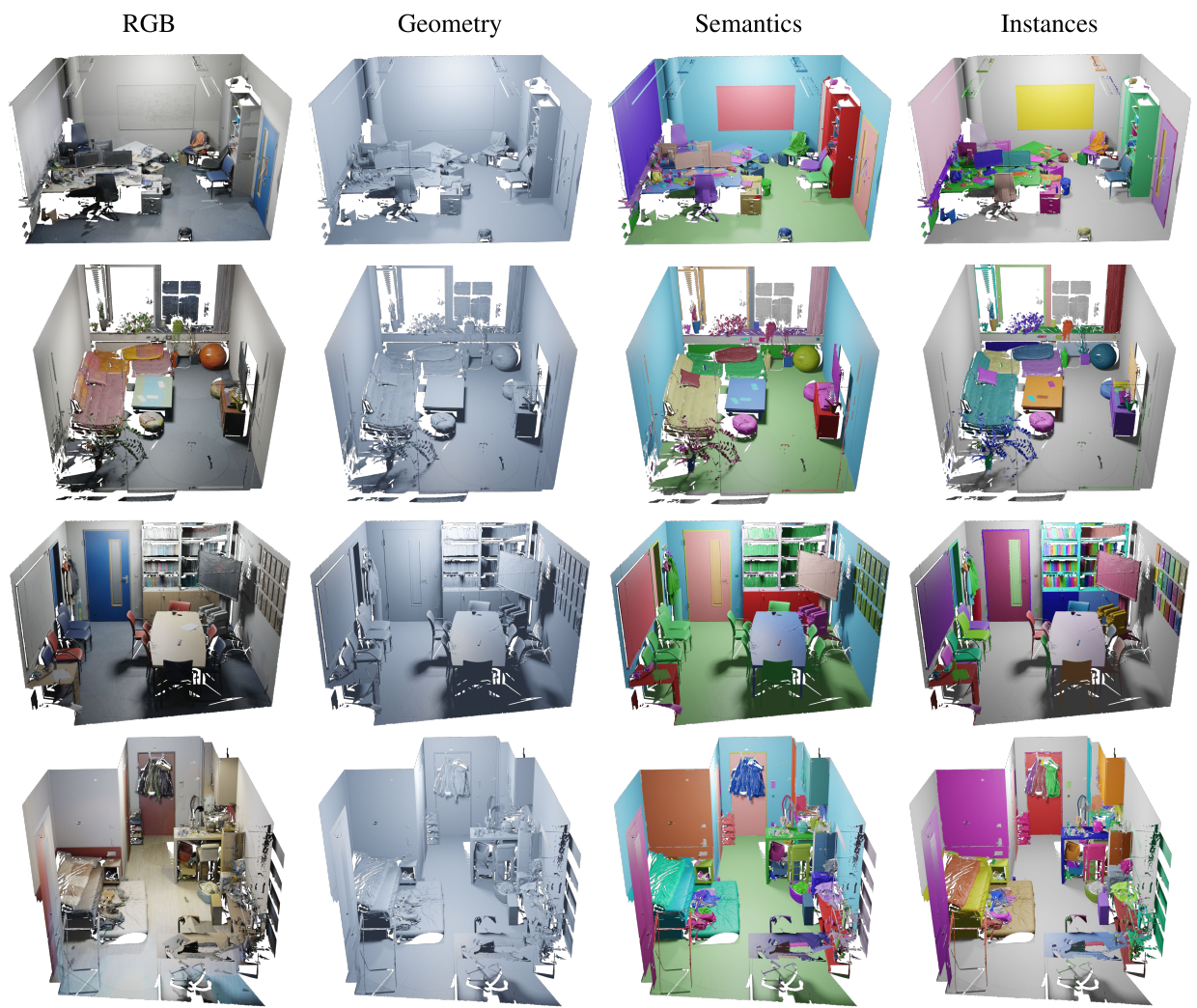}
\caption{3D reconstructions of laser scans are shown with and without color, and with semantic annotations and instance labels. The high-resolution meshes and open-vocabulary annotation allow us to annotate semantic labels in fine detail and close to 100\% completion for every scene. }
\label{fig:rgb_sem}
\end{figure*}

Semantic labels are applied onto an over-segmentation \cite{felzenszwalb2004efficient} of the decimated mesh. The segments are annotated in a 3D web interface with free-text instance labels \cite{dai2017scannet}, giving more than 1000 unique labels. The annotation process takes about 1 hour per scene on average. Examples of the semantic and instance labels, along with the colored mesh and geometry, are shown in Fig.~\ref{fig:rgb_sem}.

Additionally, in contrast to existing datasets such as ScanNet~\cite{dai2017scannet}, we allow multiple labels on each mesh segment, enabling us to capture different kinds of label ambiguity such as occlusion and part-whole relations. Examples are shown in Fig.~\ref{fig:mutlilabel_example}.

\subsection{Benchmark}
To accompany our dataset, we will release an online benchmark for both novel view synthesis and 3D scene understanding tasks.
In total, we collect \numscenes{} scenes with each scene containing a high-fidelity annotated 3D mesh, high-resolution DSLR images, and an iPhone RGB-D video. 
The dataset contains a wide variety of scenes, including apartments, classrooms, large conference rooms, offices, storage rooms, scientific laboratories, and workshops among others. 
For evaluation, we split the dataset into \numTrainScenes{}, \numValScenes{} and \numTestScenes{} training, validation and test scenes respectively following the same scene type distribution.
The dataset will be made public and aims at benchmarking novel view synthesis for both DSLR and commodity iPhone data, as well as 3D semantic and instance segmentation through an online public evaluation website. Following ScanNet~\cite{dai2017scannet}, labels of the test set will remain hidden.

%% file: arxiv/04_experiment.tex
\begin{figure*} 
\centering
\includegraphics[width=\textwidth,trim={0 7.3cm 0 0},clip]{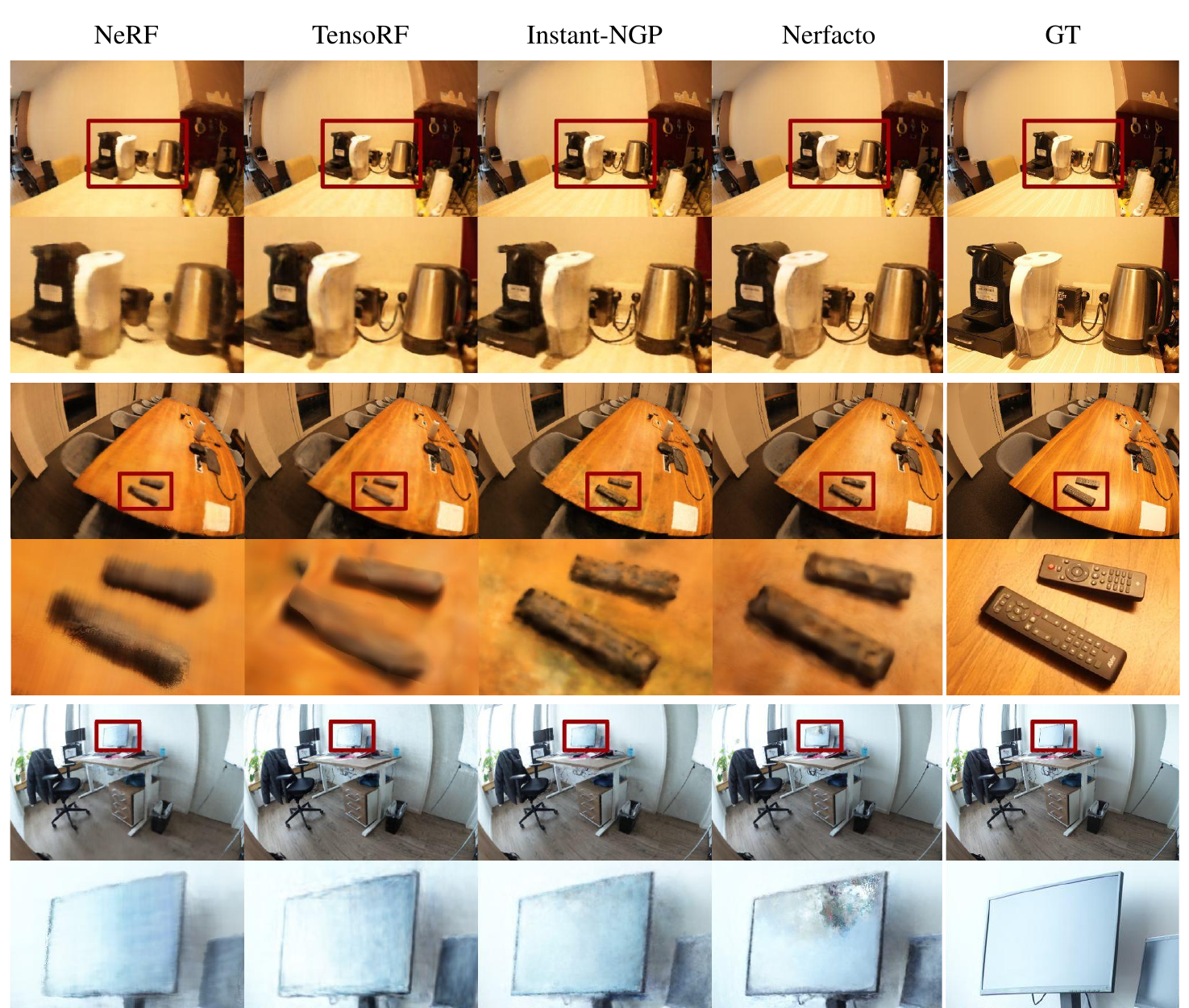}
\caption{Comparison of different novel view synthesis methods on \OURS{}. Note that existing methods achieve remarkable re-rendering results while at the same time still leaving significant room for improvement in future works.}
\label{fig:nvs_results}
\end{figure*}

\section{Experiments}

\begin{figure*}[thb!]
\setlength\tabcolsep{2pt}
\centering

\begin{subfigure}{\textwidth}
\begin{tabular}{cccc}
 PointNet++ &
 KPConv &
 MinkowskiNet &
 GT \\
 \includegraphics[width=0.25\linewidth]{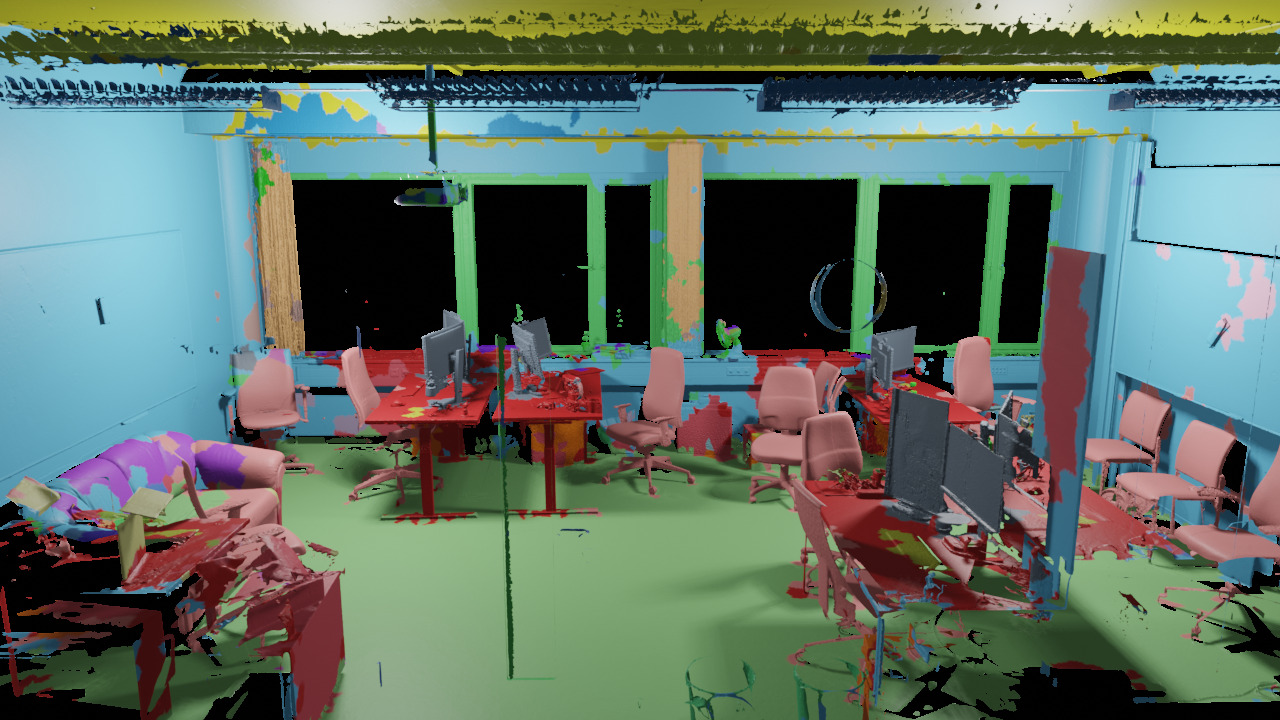} &
 \includegraphics[width=0.25\linewidth]{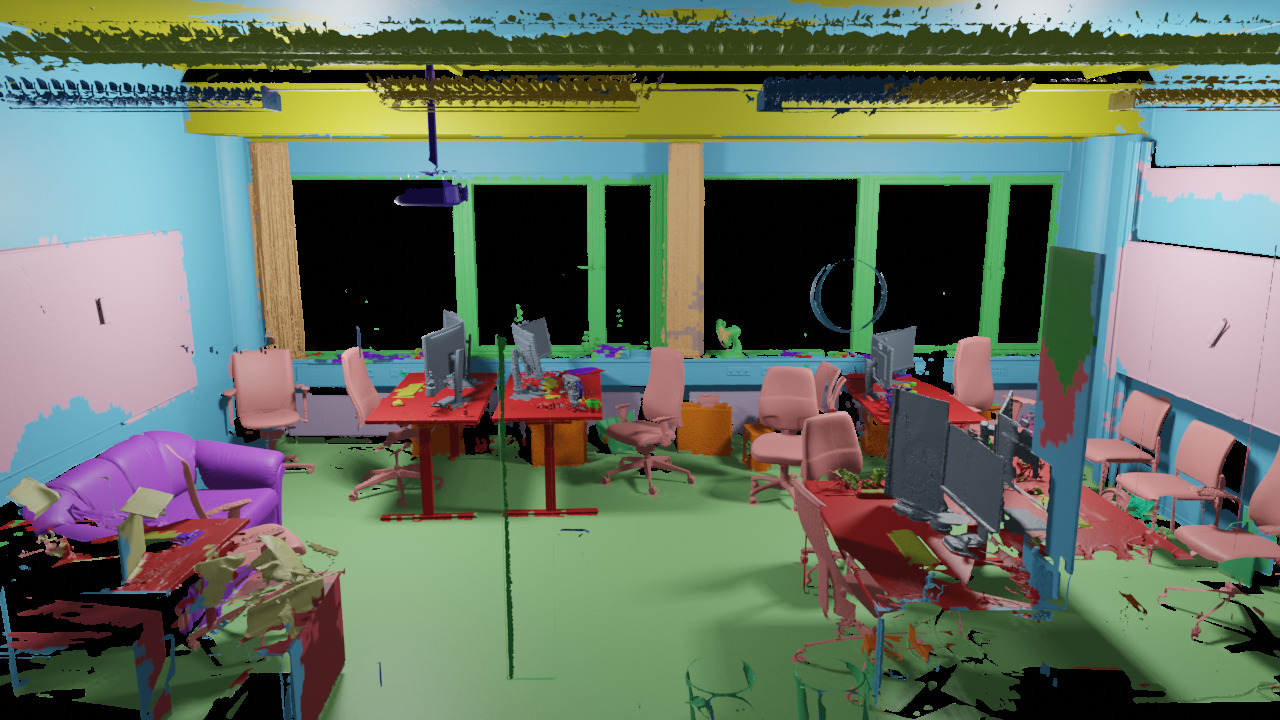} &
 \includegraphics[width=0.25\linewidth]{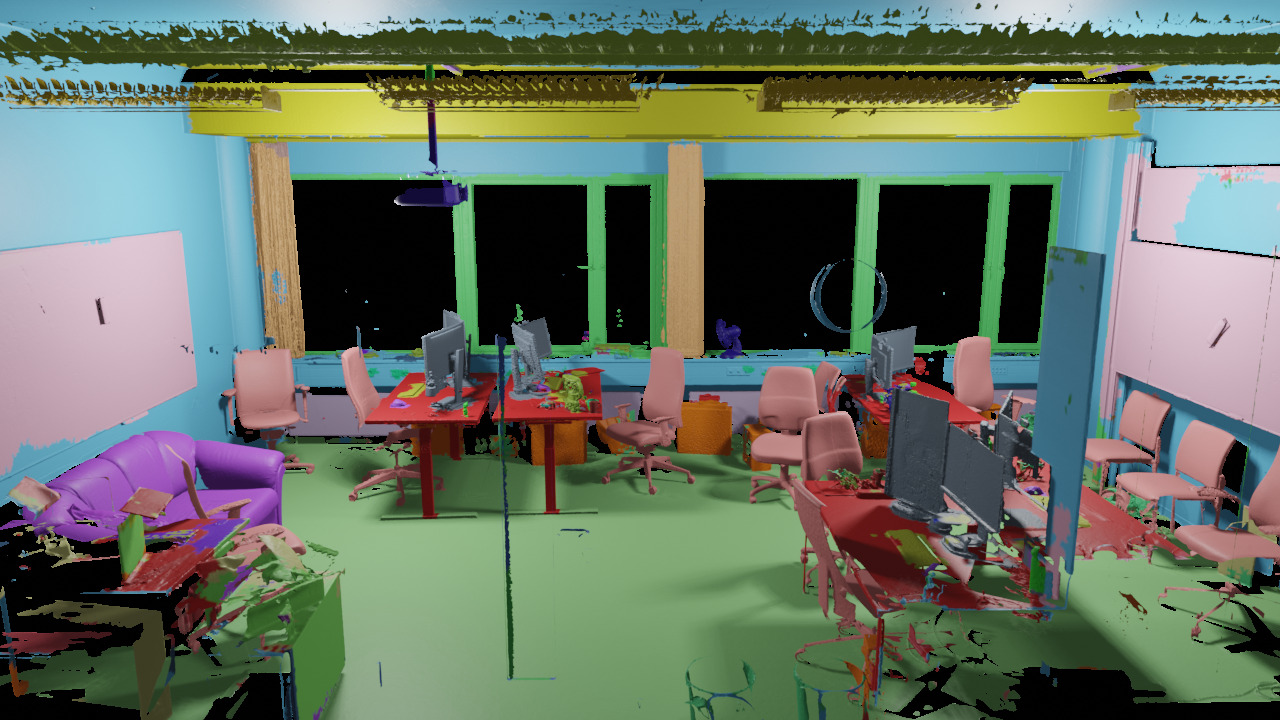} &
 \includegraphics[width=0.25\linewidth]{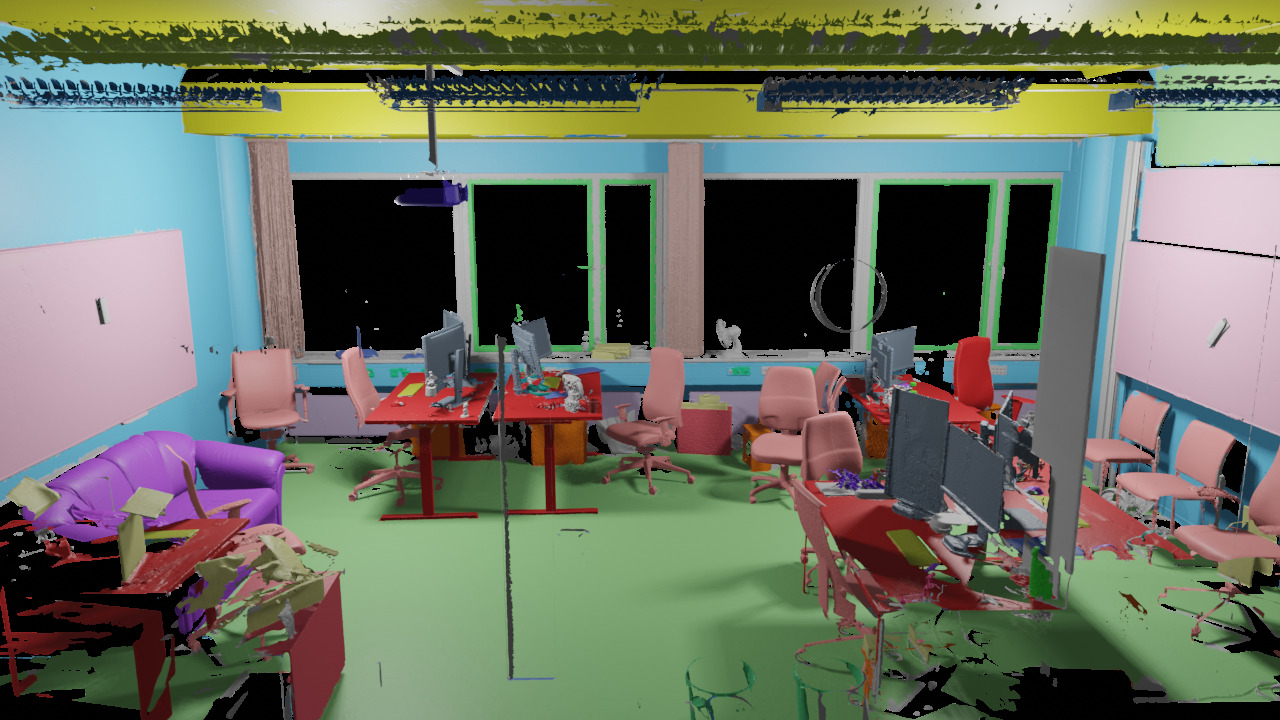} \\
 
\includegraphics[width=0.25\linewidth]{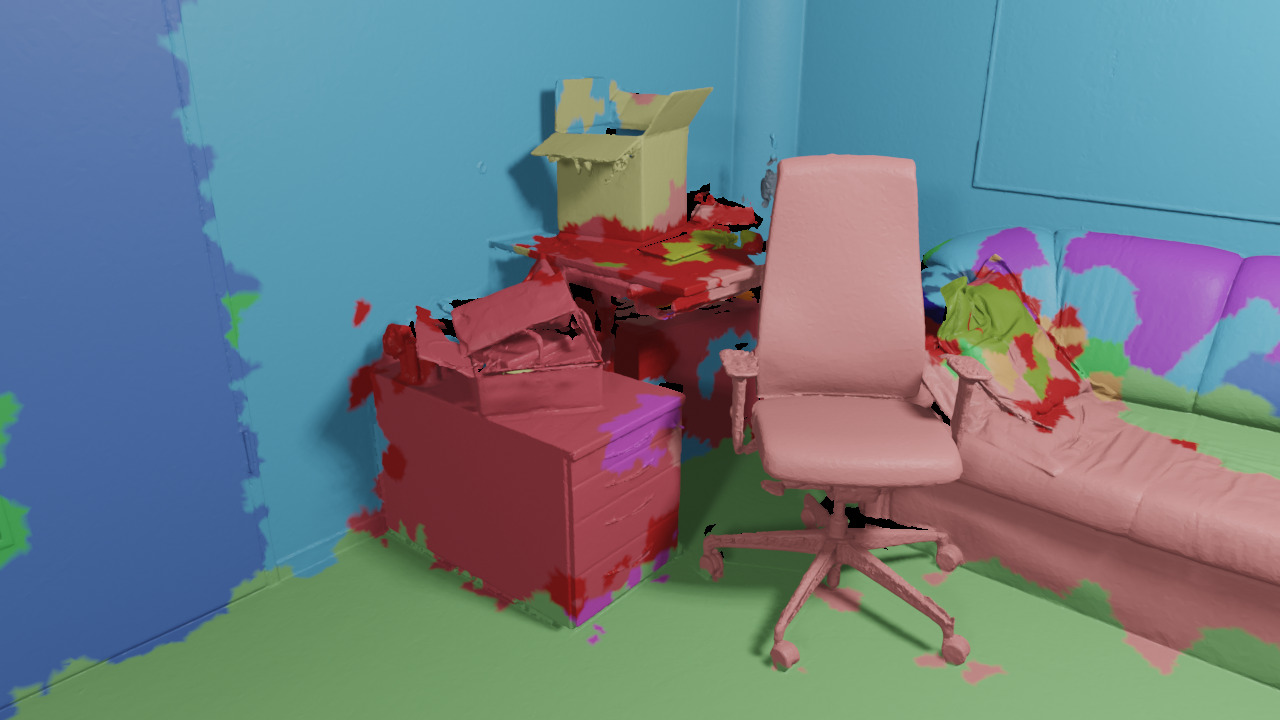} &
 \includegraphics[width=0.25\linewidth]{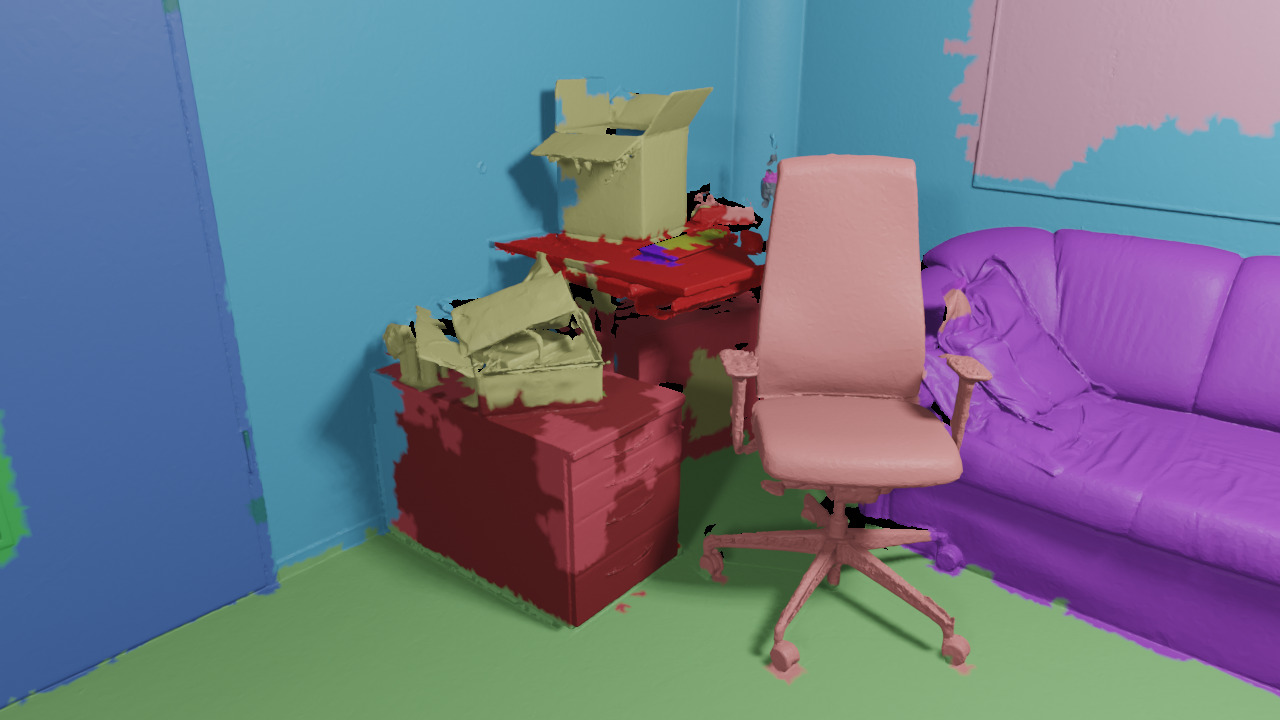} &
 \includegraphics[width=0.25\linewidth]{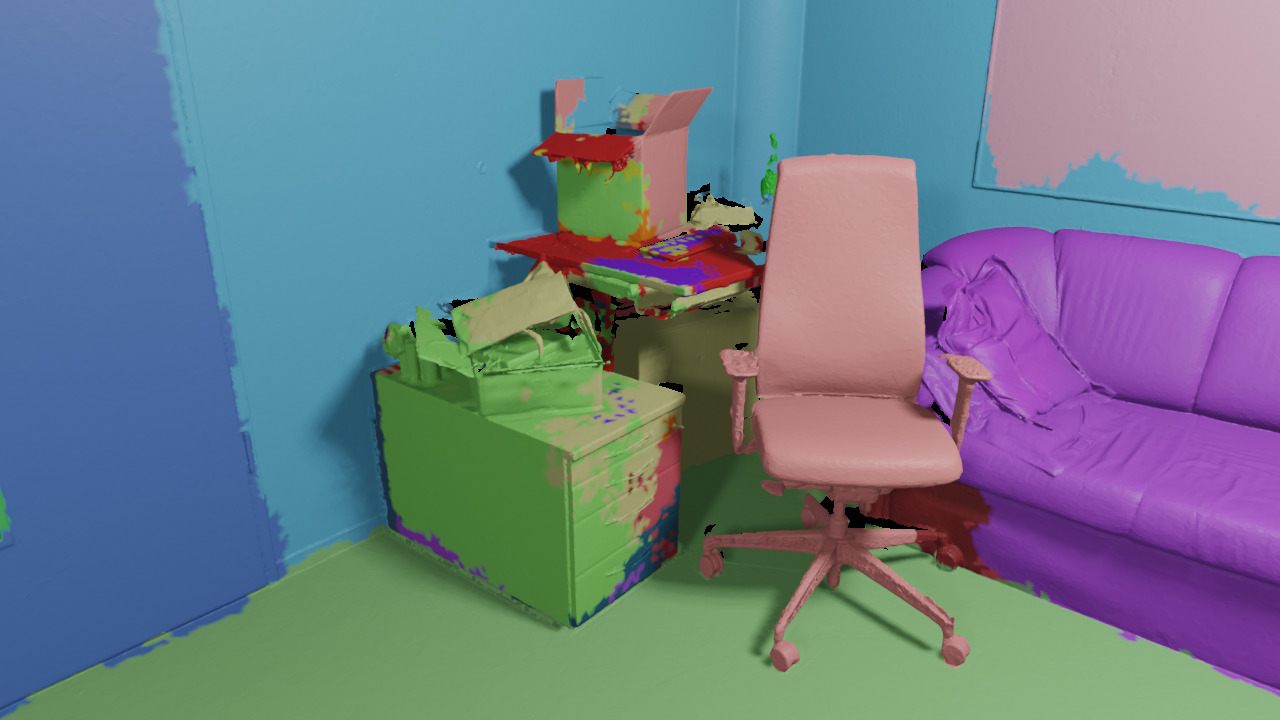} &
 \includegraphics[width=0.25\linewidth]{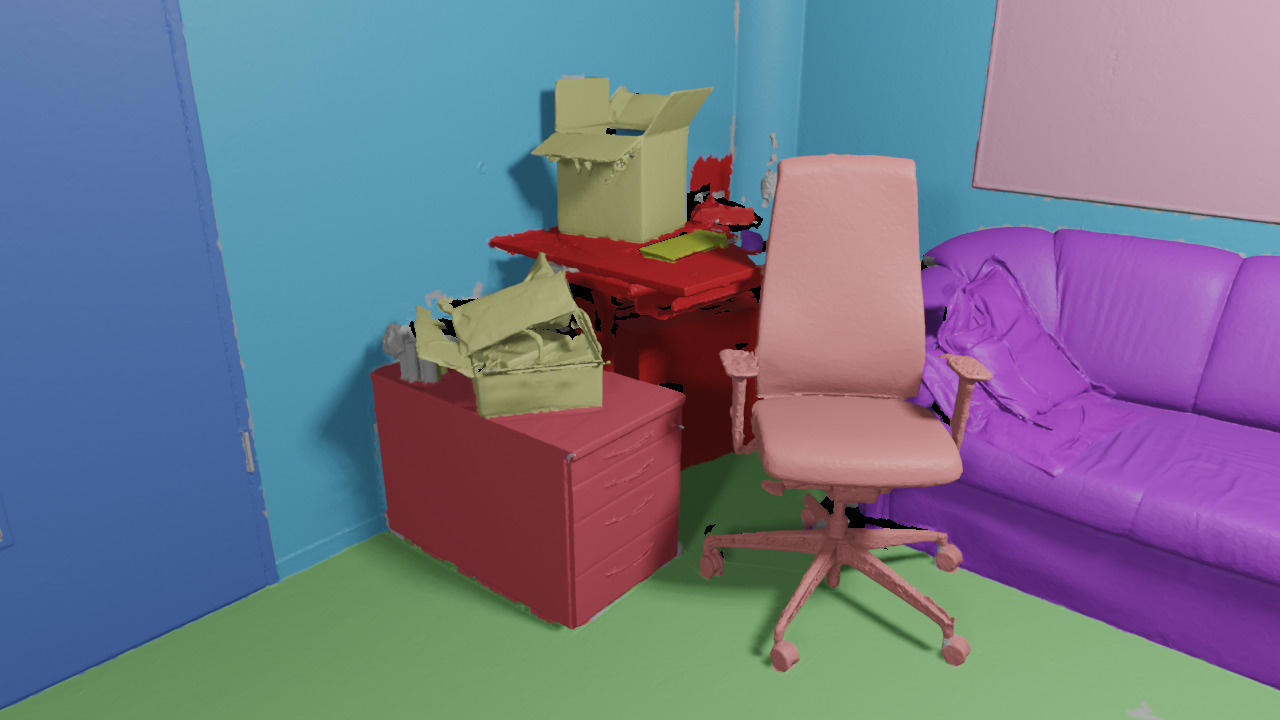} \\
\end{tabular}
\caption{3D semantic segmentation baselines. We show results of point-based PointNet++ and KPConv, and sparse-voxel based MinkowskiNet. These methods perform well on distinct objects such as chairs and cabinets, but fail to handle small objects and ambiguity such as a whiteboard on a wall.}
\end{subfigure}

\hspace{1cm}

\begin{subfigure}{\linewidth}
\begin{tabular}{cccc}
 PointGroup &
 HAIS &
 SoftGroup &
 GT \\

\includegraphics[width=0.25\linewidth]{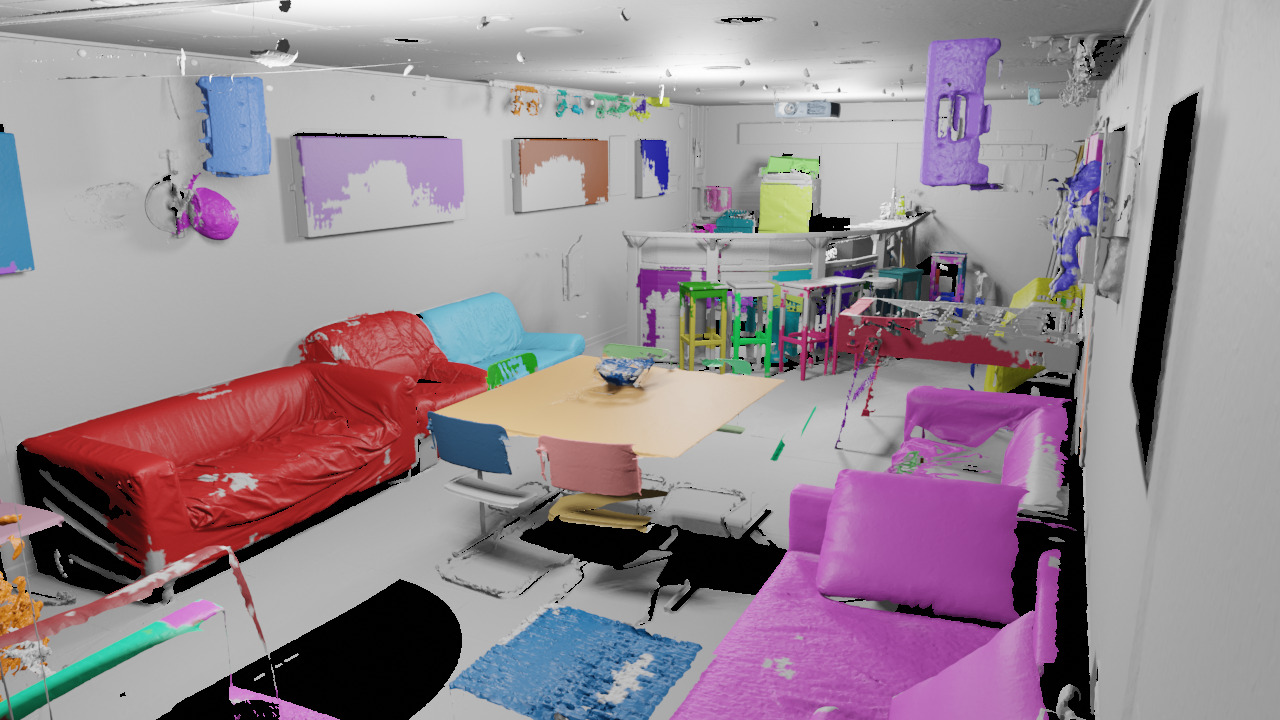} &
 \includegraphics[width=0.25\linewidth]{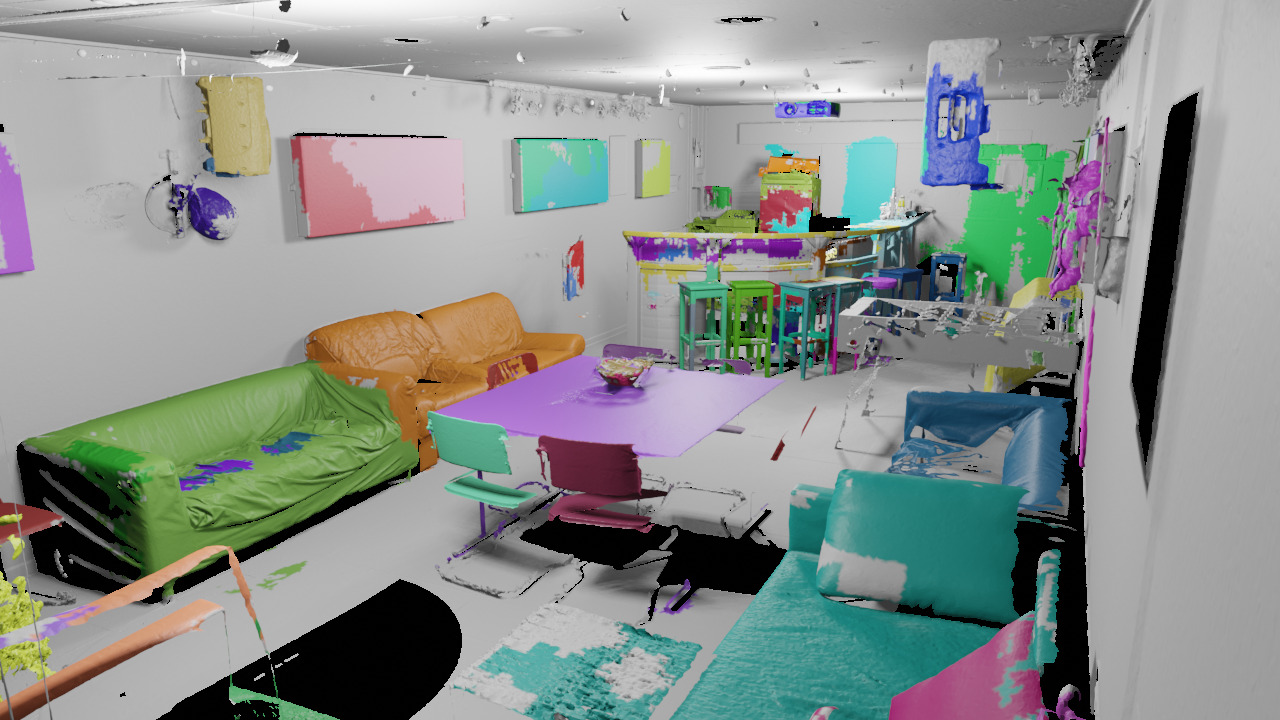} &
 \includegraphics[width=0.25\linewidth]{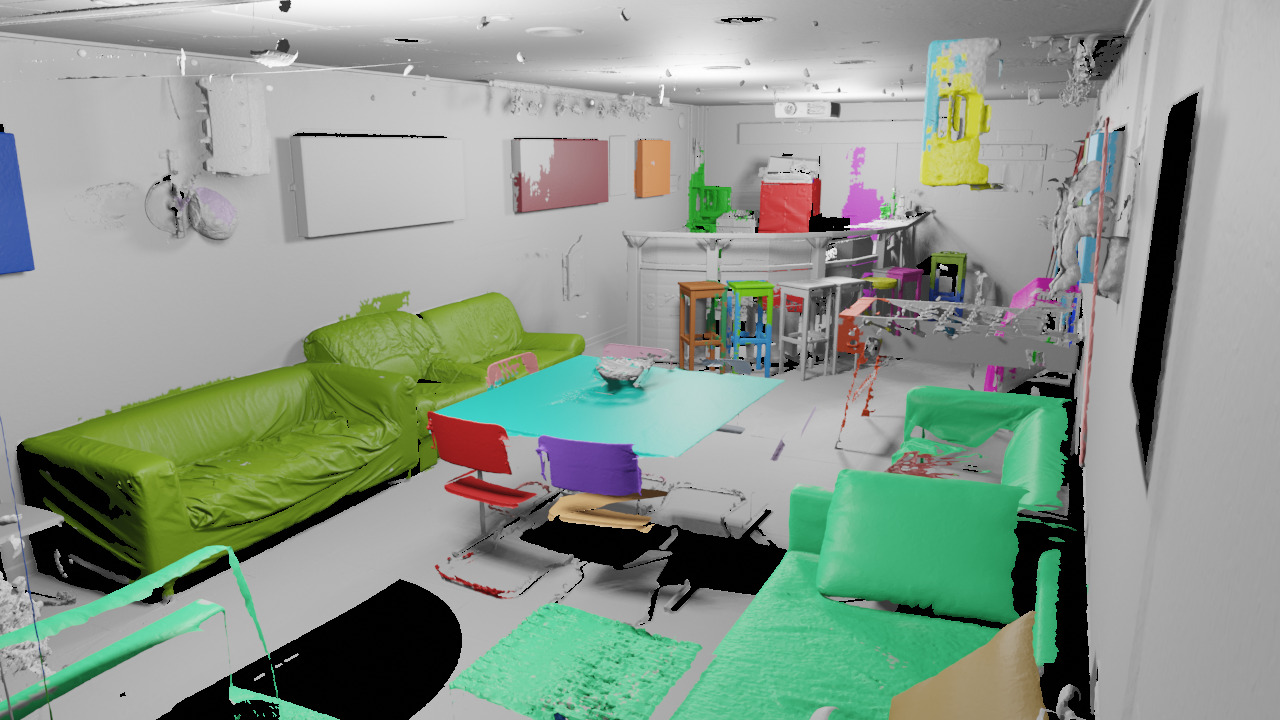} &
 \includegraphics[width=0.25\linewidth]{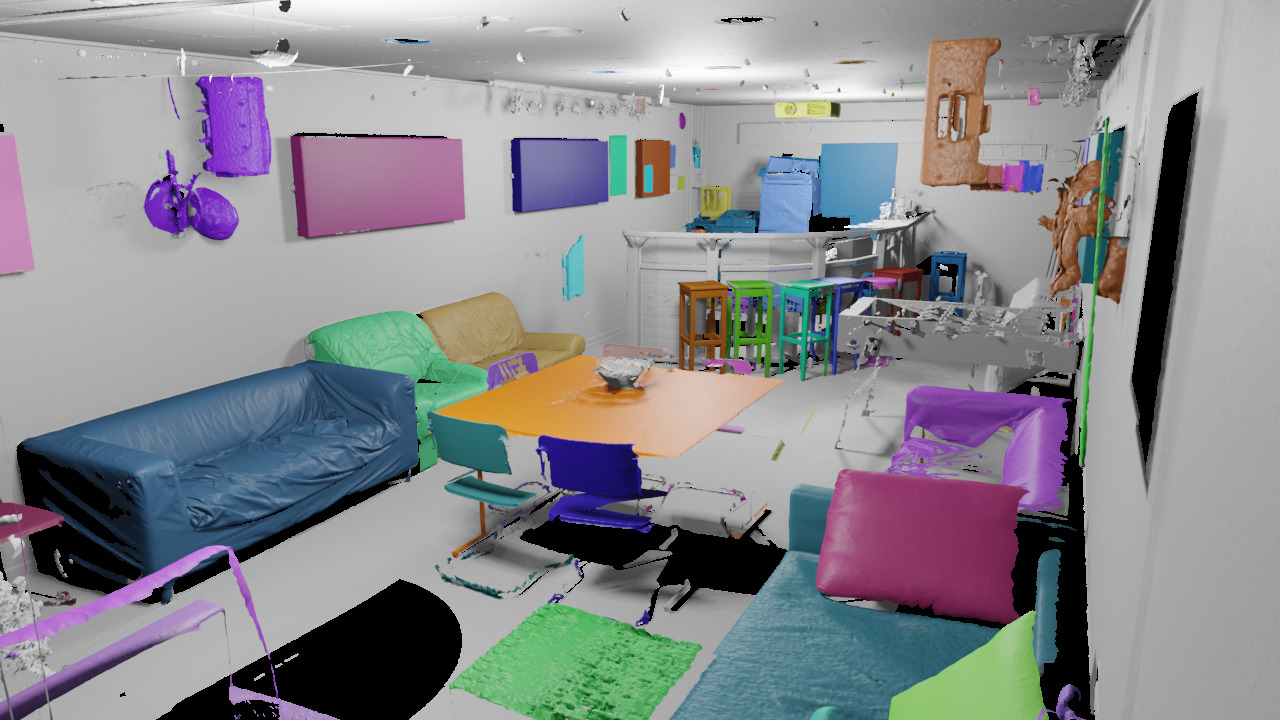} \\
 
\includegraphics[width=0.25\linewidth]{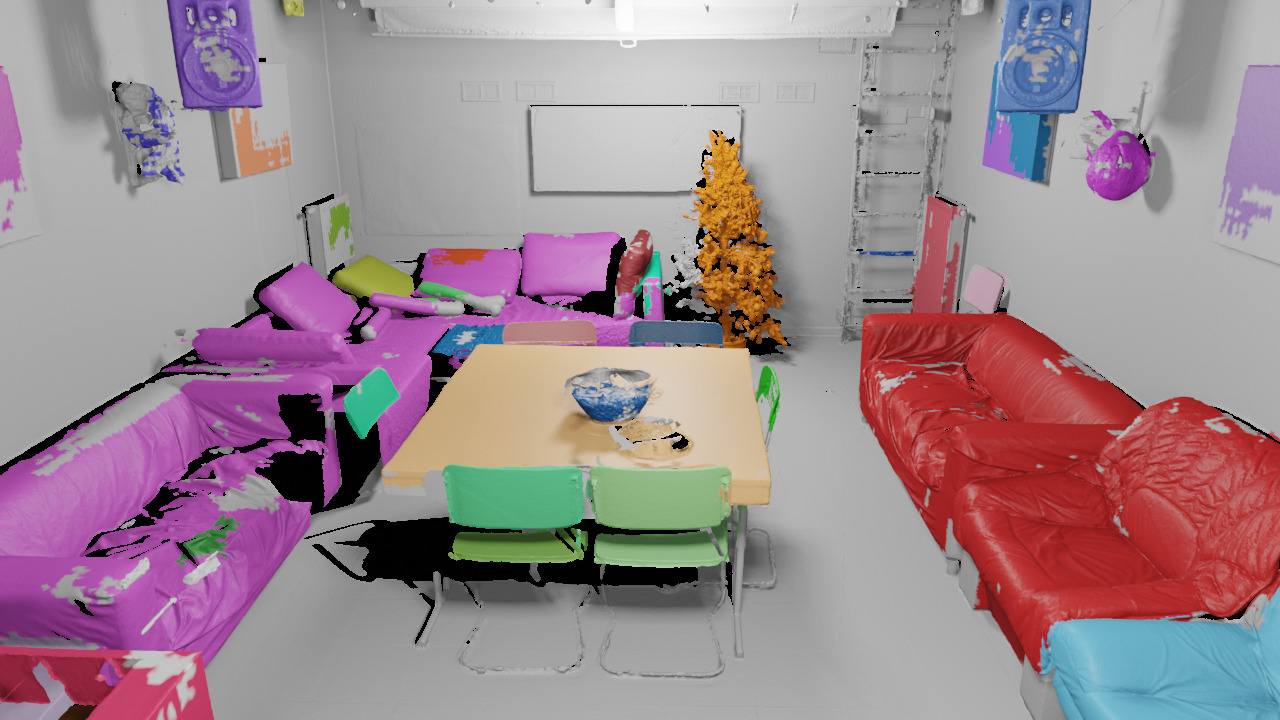} &
 \includegraphics[width=0.25\linewidth]{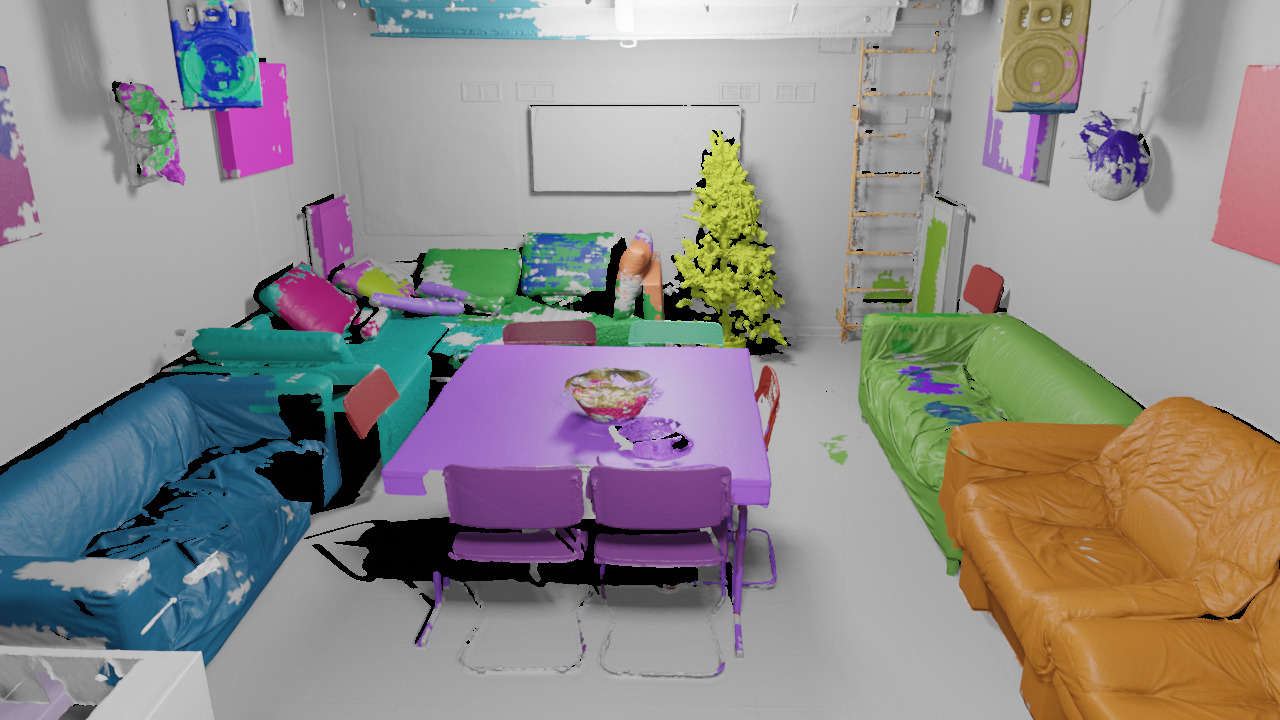} &
 \includegraphics[width=0.25\linewidth]{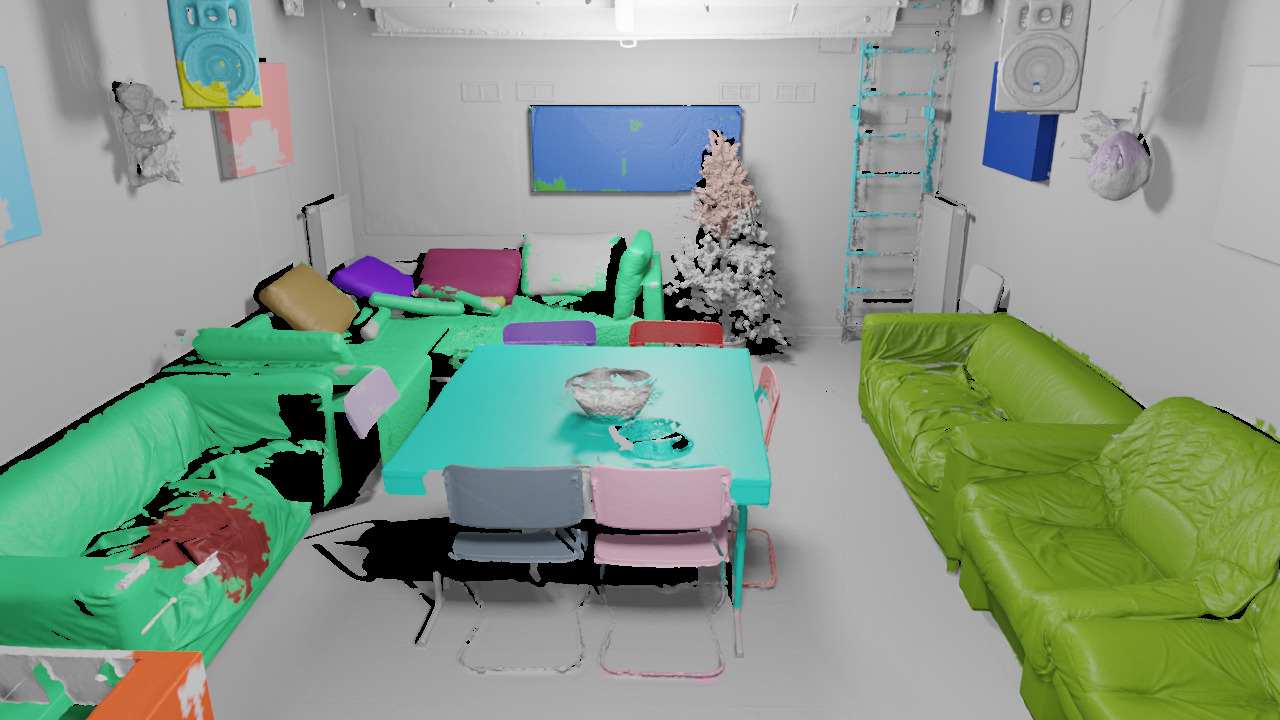} &
 \includegraphics[width=0.25\linewidth]{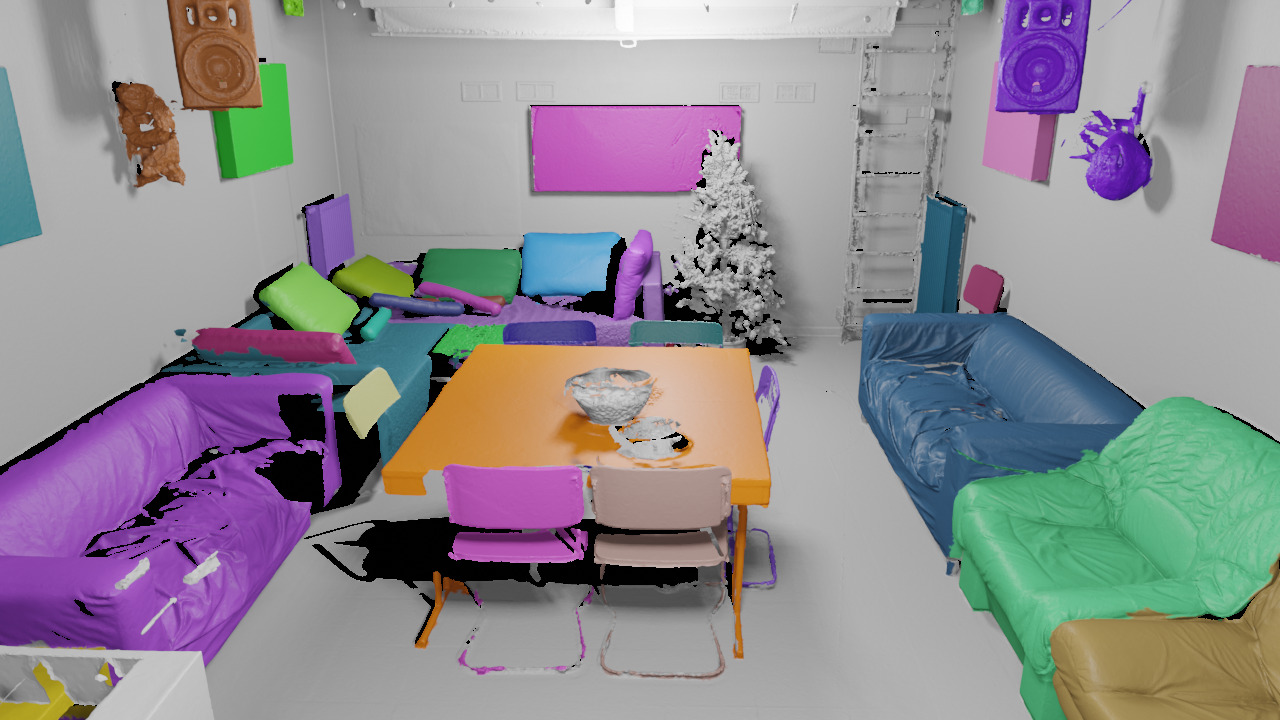} \\
\end{tabular}
\caption{3D instance segmentation baselines. We show results of PointGroup which groups points by semantic label, HAIS which groups incomplete fragments, and SoftGroup which combines bottom-up and top-down methods. These methods can recognize large distinct instances, but tend to combine nearby instances and perform poorly on small objects.}
\end{subfigure}

\caption{Qualitative results of 3D semantic and instance segmentation methods on the validation set of \OURS{}.}
\label{fig:semantic-qual}
\end{figure*}

\subsection{Novel View Synthesis} We evaluate the novel view synthesis task using two types of data as input, high-quality DSLR images and commodity  RGB images. For both experiments, we show results of NeRF \cite{mildenhall2020nerf} and its state-of-the-art variants \cite{muller2022instant,chen2022tensorf,nerfstudio} on the validation scenes. The evaluation metrics we used are PSNR, LPIPS and SSIM.

\paragraph{DSLR Data} We leverage the training images (varying from 200 to 2k images depending on the scene size) for training and compare synthesized views against the testing images. Quantitative and qualitative results are shown in Tab.~\ref{tab:nvs} and Fig.~\ref{fig:nvs_results} respectively. \OURS{} DSLR data has a wide field-of-view and consistent brightness across frames within a scene. Therefore, it is well-suited to NeRF-like methods that rely on the photometric error as supervision. On the other hand, \OURS{} is challenging for novel view synthesis since it contains large and diverse scenes, and many glossy and reflective materials for which view-dependent effects are hard to model. As shown in Fig.~\ref{fig:nvs_results} (2nd row), all methods fail to model the light reflected on the table.

In general, NeRF~\cite{mildenhall2020nerf} fails to reconstruct fine-grained details and tends to generate over-smoothed results while TensoRF~\cite{chen2022tensorf} and Instant-NGP~\cite{muller2022instant} are able to produce sharper results. However, TensoRF produces striped pattern artifacts for testing poses that differ greatly from the training poses, possibly due to the sampled tri-plane coordinates not being observed during training. Similarly, Instant-NGP outputs have floater artifacts. Among these, Nerfacto~\cite{nerfstudio}, which brings together components from multiple state-of-the-art NeRF methods \cite{muller2022instant,barron2022mip,barron2021mip,martin2021nerf}, performs the best and produces the sharpest renderings. However, it can overfit to view-dependent effects, as seen on the monitor screen in Fig.~\ref{fig:nvs_results}.

To summarize, novel view synthesis methods in real-world environments have much room for improvement, especially when reconstructing small objects and handling strong view-dependent effects.
\begin{table}[thb!]
\begin{center}
\begin{tabular}{l | c c c}
\hline
Method & PSNR $\uparrow$ & SSIM $\uparrow$ & LPIPS $\downarrow$ \\
\hline
NeRF \cite{mildenhall2020nerf} & 24.11 & 0.833 & 0.262 \\
Instant-NGP  \cite{muller2022instant}  & 24.67 & 0.846 & 0.221 \\
TensoRF      \cite{chen2022tensorf}    & 24.32 & 0.843 & 0.240 \\
Nerfacto     \cite{nerfstudio}         & \textbf{25.02} & \textbf{0.858} & \textbf{0.180} \\
\hline
\end{tabular}
\end{center}
\vspace{-4mm}
\caption{Novel view synthesis on \datasetname{} test images.}
\label{tab:nvs}
\vspace{-3mm}
\end{table}

\vspace{-0.3cm}

\paragraph{iPhone Data}
To benchmark the task of generating high-quality results by training only on commodity sensor data, we show results of training on iPhone video frames and use the DSLR images as ground truth for novel view evaluation. We perform color correction to compensate for color differences between the iPhone and DSLR captures.

\begin{table}[thb!]
\begin{center}
\begin{tabular}{l | c c c}
\hline
Method & PSNR $\uparrow$ & SSIM $\uparrow$ & LPIPS $\downarrow$ \\
\hline
NeRF \cite{mildenhall2020nerf}         & 16.29 & 0.747 & 0.368  \\
Instant-NGP  \cite{muller2022instant}  & 15.18 & 0.681 & 0.408  \\
TensoRF      \cite{chen2022tensorf}    & 15.90 & 0.721 & 0.412 \\
Nerfacto     \cite{nerfstudio}         & \textbf{17.70} & \textbf{0.755} & \textbf{0.300}  \\
\hline
\end{tabular}
\end{center}
\vspace{-0.3cm}
\caption{Novel view synthesis trained on iPhone video and evaluated on the DSLR testing set of \datasetname{}. Compared to the DSLR result in Tab.~\ref{tab:nvs}, training NVS with iPhone data is more challenging due to the motion blur, varying brightness, and less field-of-view.}
\label{tab:nvs-iphone}
\vspace{-2.5mm}
\end{table}

Results are shown in Tab.~\ref{tab:nvs-iphone}, and are significantly worse than those from training on DSLR images. This is mainly due to motion blur and varying brightness of the frames in the captured video. Additionally, blurriness and a small field-of-view can cause misalignments in  the structure-from-motion (SfM) camera poses.

Therefore, to perform NVS on consumer-grade data without a controlled scanning process, an NVS method should be robust to noisy camera poses, blurriness, and brightness changes. 

\vspace{-4mm}
\paragraph{Generalization Across Scenes}

\begin{table}[thb!]
\begin{center}

\begin{tabular}{l | c c c}
\hline
Method & PSNR $\uparrow$ & SSIM $\uparrow$ & LPIPS $\downarrow$ \\
\hline
Nerfacto            & 25.02 & 0.858 & 0.180 \\
Nerfacto + pix2pix  & \textbf{25.42} & \textbf{0.869} &	\textbf{0.156} \\
\hline
\end{tabular}
\end{center}
\vspace{-0.3cm}
\caption{We apply pix2pix~\cite{pix2pix2017} on the output of Nerfacto. This general prior learned from \OURS{} improves rendering quality.}
\label{tab:nvs-general}
\end{table}

Since \OURS{} contains a large number of scenes, we show that it can be used to train general priors for novel view synthesis, thus improving over traditional single-scene training. The naive solution we consider is to train a 2D pix2pix ~\cite{pix2pix2017} across scenes by refining the per-scene Nerfacto outputs while freezing the Nerfacto model weights. As shown in Tab.~\ref{tab:nvs-general}, the general prior learned from \OURS{} can improve the performance of Nerfacto.

\subsection{3D Semantic Understanding}

We evaluate semantic and instance segmentation methods on the 5\% decimated meshes by predicting labels on the vertices and comparing with ground truth labels. 
\vspace{-0.4cm}
\paragraph{3D Semantic Segmentation}
We compare 4 methods for 3D semantic segmentation on \OURS{}: point-based methods PointNet \cite{qi2017pointnet} and PointNet++~\cite{qi2017pointnet++} and KPConv~\cite{thomas2019kpconv}, and sparse-voxel-based MinkowskiNet~\cite{choy20194d} on the top 78 semantic classes. 
\vspace{-0.4cm}
\paragraph{3D Instance Segmentation}
We compare 3 methods for  3D instance segmentation on \OURS{}: PointGroup \cite{jiang2020pointgroup}, HAIS \cite{chen2021hierarchical} and SoftGroup \cite{vu2022softgroup} on 75 instances classes - semantic classes excluding wall, ceiling and floor. 

Quantitative results are shown in Tab.~\ref{tab:sem-inst-seg-results}, and qualitative results are shown in Fig. \ref{fig:semantic-qual}. All methods can distinguish large and well separated objects such as chairs and sofas, but perform poorly on ambiguous objects such as a whiteboard on a white wall and smaller objects. 

\begin{table}[htb!]
\centering

\parbox{.45\linewidth}{
    \begin{tabular}{l | c }
    \hline
    Method & mIoU \\
    \hline
    PointNet & 0.07  \\
    PointNet++ & 0.15 \\
    Minkowski & 0.28  \\
    KPConv       &   0.30 \\
    \hline
    \end{tabular}
}
\parbox{.45\linewidth}{
    \begin{tabular}{l | c }
    \hline
    Method & AP50  \\
    \hline
    PointGroup  & 0.148  \\ 
    HAIS   & 0.167 \\ 
    SoftGroup & 0.237 \\ 
    \hline
    \end{tabular}
}
\caption{Quantitative results of 3D semantic and instance segmentation baselines on \OURS{}.}
\label{tab:sem-inst-seg-results}
\vspace{-3mm}
\end{table}

%% file: arxiv/05_conclusion.tex
\section{Limitations and Future Work}
\OURS{} contains large-scale and high-quality DSLR captures which we believe will open up opportunities for novel view synthesis methods to generalize over multiple scenes and improve rendering quality \cite{yu2021pixelnerf, wang2021ibrnet, chen2021mvsnerf,suhail2022generalizable,liu2022neural,muller2022diffrf}. Further, the registered DSLR and semantic annotations allow the combination of radiance and semantic fields on \OURS{} \cite{zhi2021place,vora2021nesf,fu2022panoptic,kundu2022panoptic,siddiqui2022panoptic}. 
Nevertheless, there are some limitations of the dataset. Since we fix the DSLR brightness settings for each scene to ensure photometric consistency, some parts, such as light sources, may suffer from overexposure, while poorly-lit areas may be underexposed. Due to the expensive data collection process, \OURS{} cannot scale at the same rate as 2D datasets \cite{lin2014microsoft,bdd100k}.

\section{Conclusion}
We present \OURS{}, a large-scale dataset with high-fidelity 3D geometry and high-resolution RGB images of indoor scenes, and show how it enables challenging benchmarks for NVS and semantic understanding.
The high-quality DSLR capture allows benchmarking of NVS methods at scale and the development of generalized NVS methods, while the iPhone capture raises the challenging task of handling motion blur and noisy poses. 
Additionally, long-tail and multi-label annotations on the reconstructions enable fine-grained semantic understanding while accounting for label uncertainty. 
Registering all modalities into a single coordinate system allows multi-modal learning of semantics and the usage of semantic priors for novel view synthesis.
We hope the \OURS{} dataset and benchmark will open up new challenges and stimulate the development of new methods for NVS and semantic understanding.

%% file: arxiv/06_appendix.tex
\appendix

\section{Details of Data Collection}
\label{sec:app-data-collection}
\begin{figure}[thb!]
\centering
\includegraphics[width=0.85\linewidth]{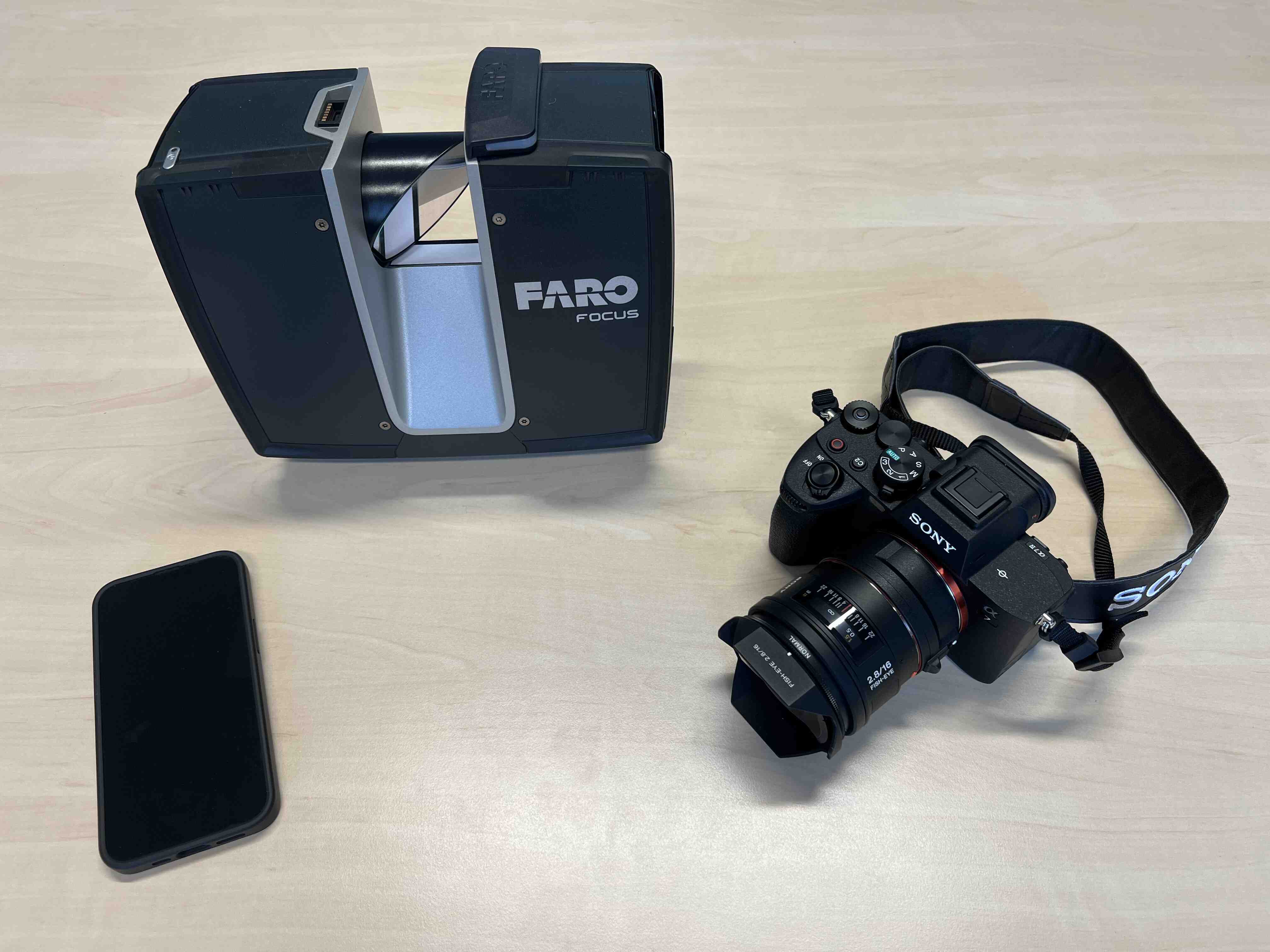}
\caption{Our hardware setup consists of a Faro Focus Premium laser scanner, Sony Alpha 7 IV DSLR camera, and iPhone 13 Pro with a LiDAR sensor.}
\label{fig:hardware}
\end{figure}

Our hardware setup is shown in Fig.~\ref{fig:hardware}. 
 We aim to capture large spaces as a single scene, rather than splitting them into separate rooms in order to provide more context as well as increase the complexity for downstream tasks. 
Captures from the three sensors are performed as close together in time as possible to avoid inconsistencies in lighting between the different modalities.

\vspace{-0.5cm}

\subsection{Laser Scan}
 We use the 1/4 resolution and 2x quality setting for the Faro scanner, which takes about 2 minutes for each scan. 
We estimate the normals of the point cloud as the cross product of  neighboring scan points in the corresponding 2D scan grid, and voxelize the points to a 1$mm$ resolution. 

Point clouds from different scans are first merged at a 1$mm$ resolution. The resulting point cloud is chunked into overlapping cubes of side $0.5m$. Poisson reconstruction \cite{kazhdan2013screened} is applied on each chunk with a depth of 9 (grid size of $2^9$) which gives a grid cell size of $0.5m / 512 < 1mm$.  
The chunk overlap is set to 150 grid cells. The resulting chunk meshes are first trimmed to match the original point cloud, and then clipped by 75 grid cells.  The vertices of these chunk meshes are finally joined with a $1mm$ threshold to form a single mesh. Quadric edge collapse is then applied \cite{garland1997surface} to the full resolution mesh, to obtain smaller meshes with $12.5\%$, $5\%$ and $1.5\%$ of the original number of faces.

\subsection{DSLR Images}
Our DSLR capture settings are optimized for training and evaluating of novel view synthesis methods. Specifically, we fix the white balance and exposure time to have consistent lighting throughout the scene. We use a wide field-of-view fisheye lens which provides larger overlap between images, and empirically improves both camera pose registration and novel view synthesis. Exposure time is set to 1/100s to avoid  flickering effects from indoor lights and minimize motion blur.

\subsection{iPhone}
In contrast to the DSLR capture, our iPhone recordings are performed in the default automatic mode of the iPhone, making novel view synthesis more challenging. We show the setting comparison between DSLR and iPhone in Tab.~\ref{tab:setting}.

\begin{figure}[thb!]
\centering
\includegraphics[width=1.0\linewidth]{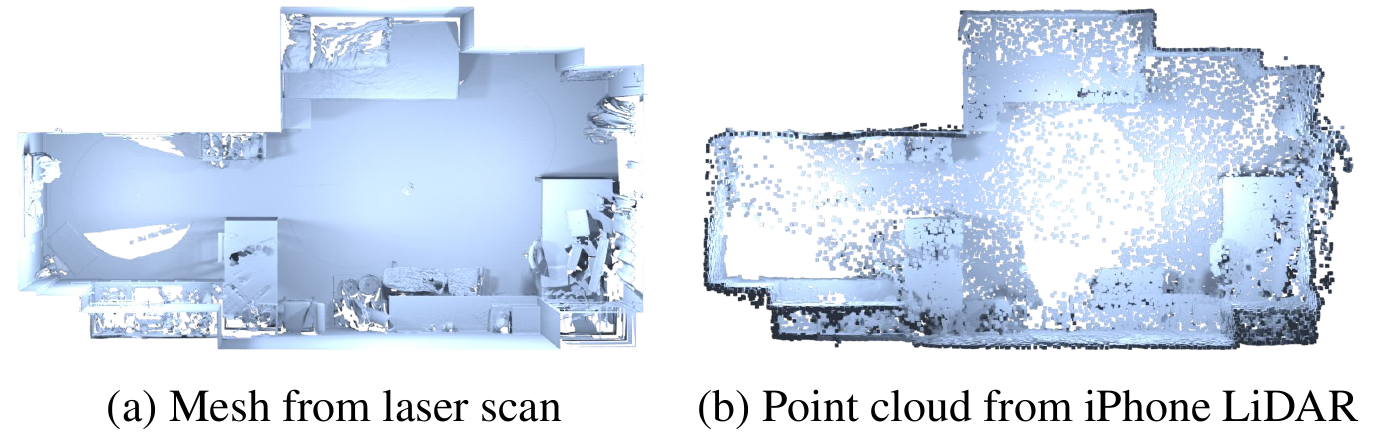}
\caption{Comparison of 3D reconstructions from the laser scan and low-resolution point cloud from the iPhone. Depth images from iPhone LiDAR are noisy and low resolution.}
\label{fig:scanner_iphone}
\end{figure}

We show the comparison of the point cloud generated from the iPhone depth map and 3D geometry obtained from the laser scanner in Fig.~\ref{fig:scanner_iphone}. The 3D geometry from the scanner is much cleaner and preserves finer details.

\begin{table}[thbp]
    \centering
    \begin{tabular}{l | c c}
    \hline
        Settings & DSLR &  iPhone\\
    \hline
        Auto white balance & \xmark & \cmark \\
        Auto focus & \xmark & \cmark \\
        Auto exposure & \xmark & \cmark \\
        Field of view (deg) & 180 & 71 \\
    \hline
    \end{tabular}
    \caption{Comparison of capture settings between DSLR and iPhone.}
    \label{tab:setting}
\end{table}

\vspace{-0.5cm}

\section{Benchmark}
\label{sec:app-benchmark}
The \OURS{} dataset will be made public, along with an online benchmark for the following tasks.

\subsection{Novel View Synthesis}
\begin{figure}[htb!]
\centering
\includegraphics[width=\linewidth]{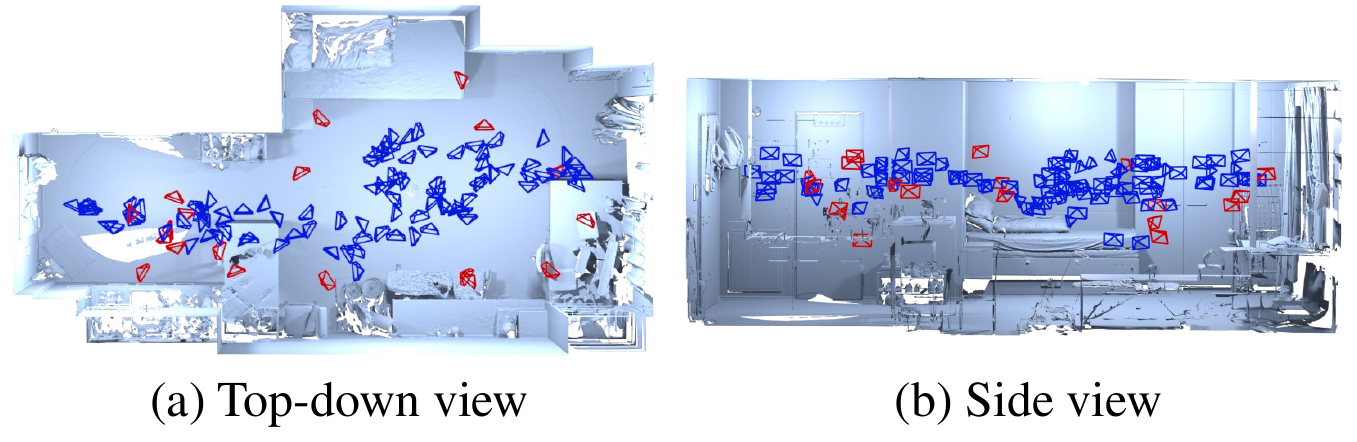}
\caption{
The unseen poses (red) of our DSLR capture for evaluation are challenging since they are very different from the training trajectory (blue) in terms of translation and orientation. 
}
\label{fig:nvs_train_test}
\end{figure}

Given a set of training images of a scene and unseen camera poses, a method must synthesize the views at the unseen poses. 
The unseen poses are captured at challenging viewpoints independently of the training trajectory (e.g., Fig.~\ref{fig:nvs_train_test}).
Evaluation metrics include PSNR, LPIPS and SSIM.

 To benchmark methods trained on iPhone data, we evaluate against DSLR images as ground truth. This setting is more challenging since the methods are expected to produce high-quality outputs based on commodity-level input.

\begin{figure*}[htb!]
\centering
\includegraphics[width=\textwidth]{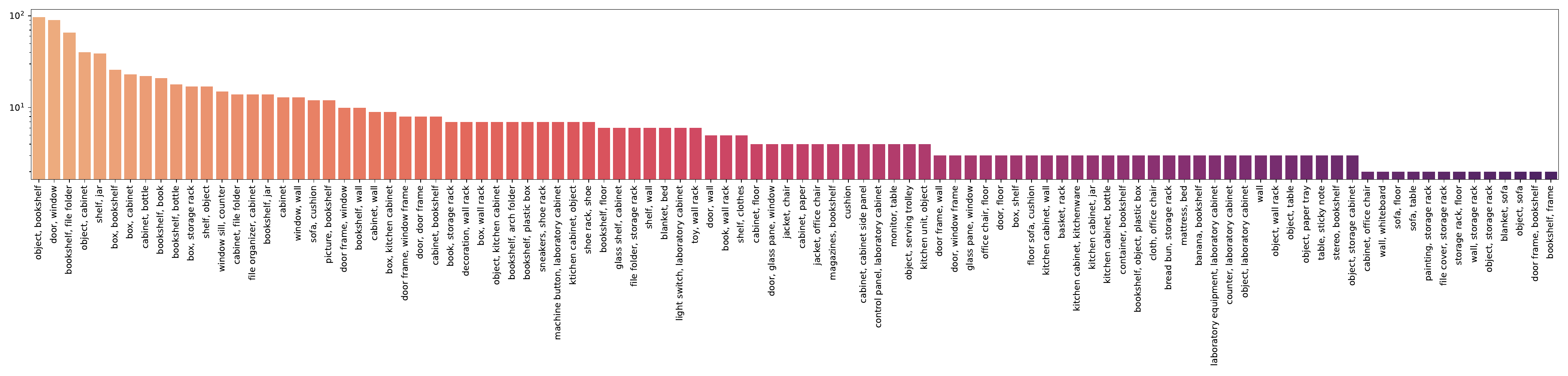}
\caption{Top 100 pairs of semantic classes that are multi-labeled together on a log scale.}
\label{fig:multi-label_count}
\end{figure*}

\begin{figure*}
\centering
\includegraphics[width=\textwidth]{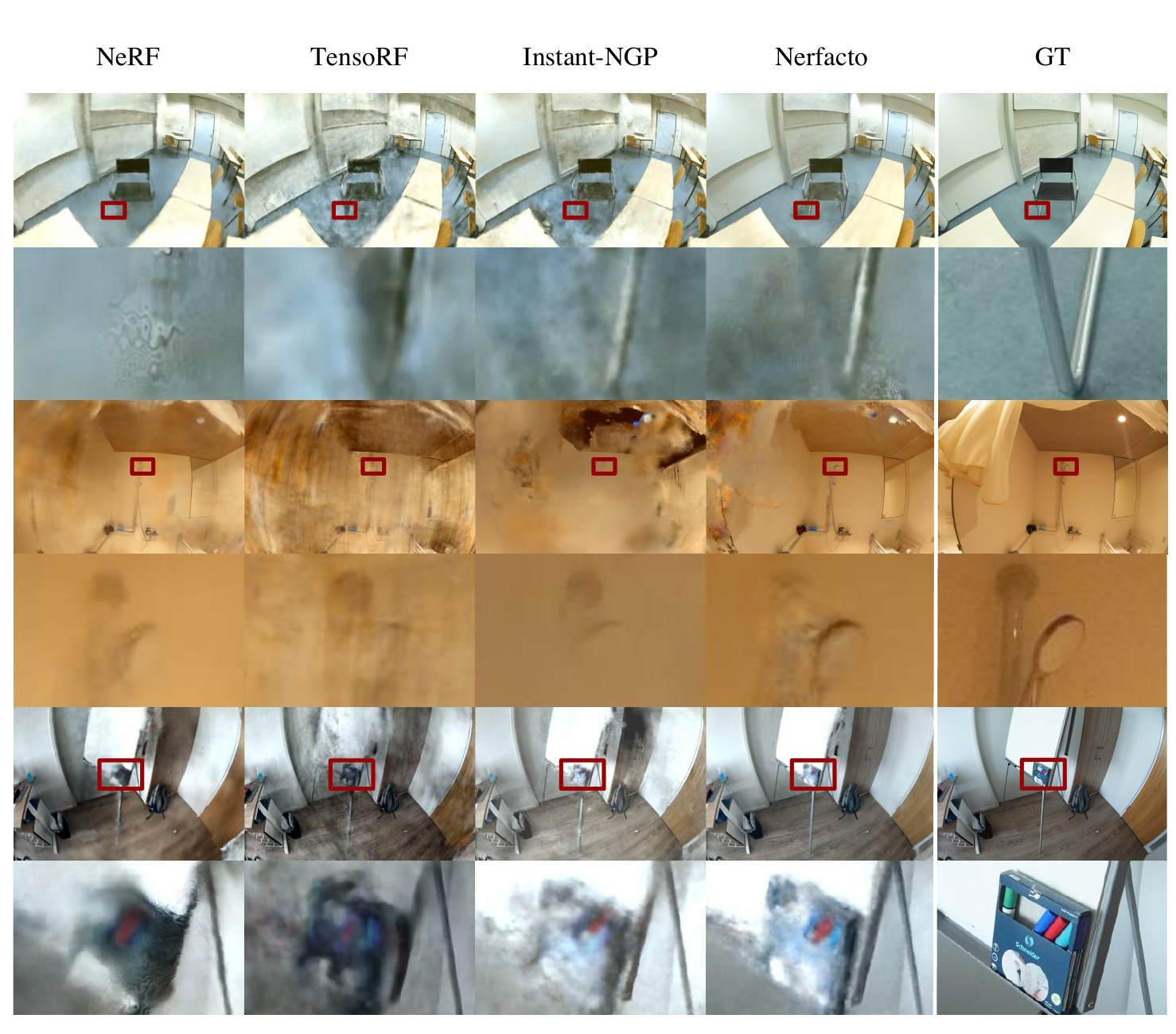}
\caption{Comparison of different novel view synthesis methods on iPhone data using the DSLR ground truth. Compared to DSLR results shown in the main paper, NVS on commodity iPhone data generates more artifacts and blurry results.}
\label{fig:iphone-nerf}
\end{figure*}

\subsection{3D Semantic Understanding}

Given colored meshes of scenes and posed RGB images, we evaluate predictions of 3D semantic segmentation methods on vertices against the ground-truth vertex labels; this is similar to the semantic benchmark on ScanNet~\cite{dai2017scannet}. Evaluation metrics include per-class intersection-over-union (IoU) and mean IoU. Similarly, 3D instance segmentation methods are evaluated on ground truth instance masks as well as semantic labels, and evaluation metrics include AP25, AP50 and AP similar to the ScanNet benchmark.

Our benchmark will include more than 100 frequent object classes for both tasks, and evaluation is performed against the multilabeled ground truth -- which is not possible in any prior 3D semantic scene understanding benchmark. Hence, we will allow submissions to provide more than one prediction per vertex. 

Following ScanNet~\cite{dai2017scannet}, we maintain a hidden test set and build an online public evaluation website. This website will provide for entries to the latest state-of-the-art methods to facilitate comparisons for new submissions. Importantly, the test set for novel view synthesis tasks does not overlap with the one for semantic understanding tasks, so that the input meshes of the latter cannot be used to guide the former.

\section{3D Semantic Understanding}
\label{sec:app-semantic-understanding}
\paragraph{Semantic Annotation}
Semantic annotation is performed using a web interface that allows the annotator to apply a free-text label to every mesh segment. Annotation of one scene takes about 1 hour on average, after which a verification pass is done by another annotator to fix incorrect  labels. A set of guidelines with reference images is provided to the annotators to perform consistent annotation over similar classes, such as shelf, cabinet, cupboard, wardrobe, bookshelf. Importantly, the guidelines describe common cases for multilabel annotation such as ``jacket on a chair'', ``bedsheet on a bed'' and so on. The distribution of the most frequent multilabeled classes is shown in  Fig.~\ref{fig:multi-label_count}.

\vspace{-0.4cm}

\begin{figure*}[thb!]
\thisfloatpagestyle{empty}
\setlength\tabcolsep{2pt}
\centering

\begin{subfigure}{\linewidth}
\begin{tabular}{cccc}
 PointNet++ &
 KPConv &
 MinkowskiNet &
 GT \\
 \includegraphics[width=0.25\linewidth]{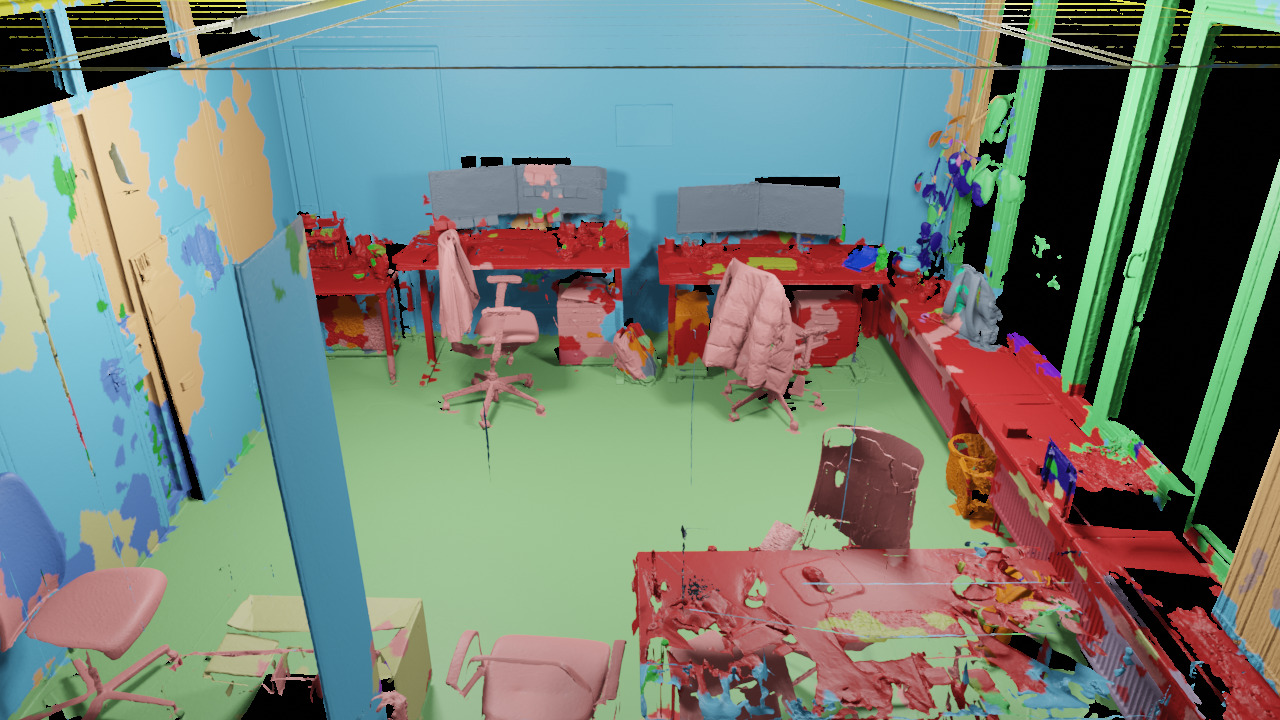} &
 \includegraphics[width=0.25\linewidth]{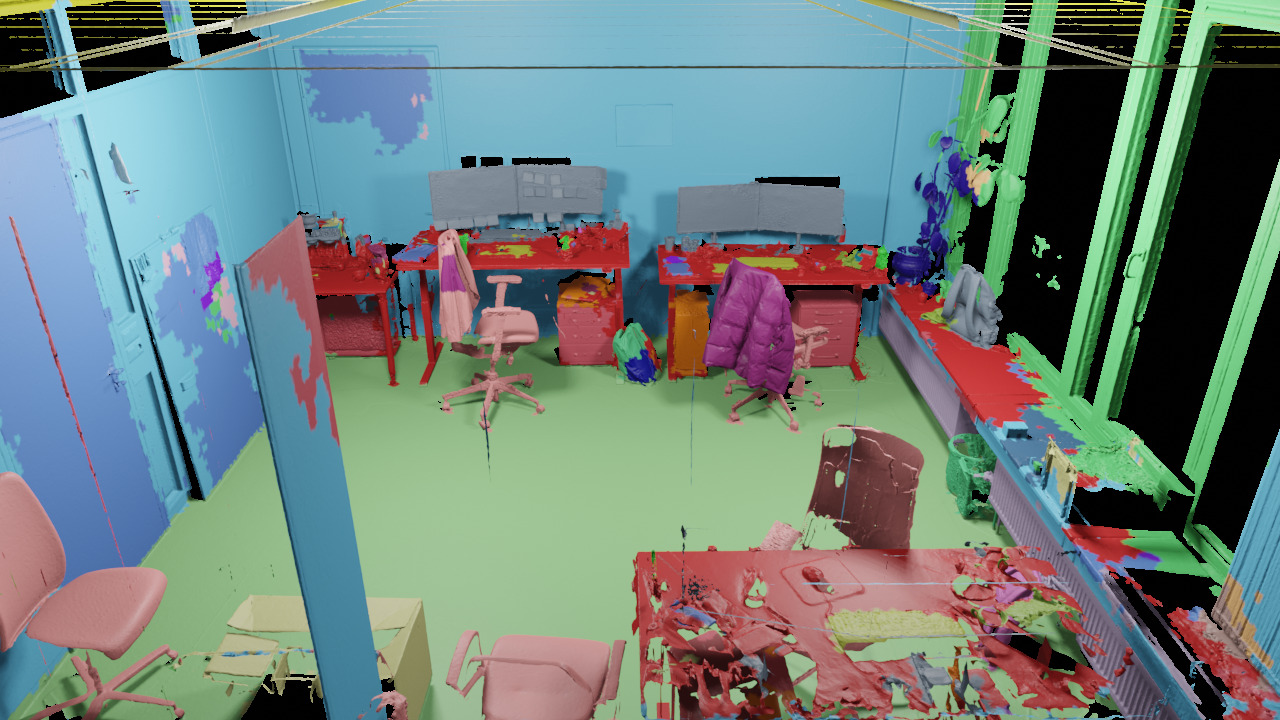} &
 \includegraphics[width=0.25\linewidth]{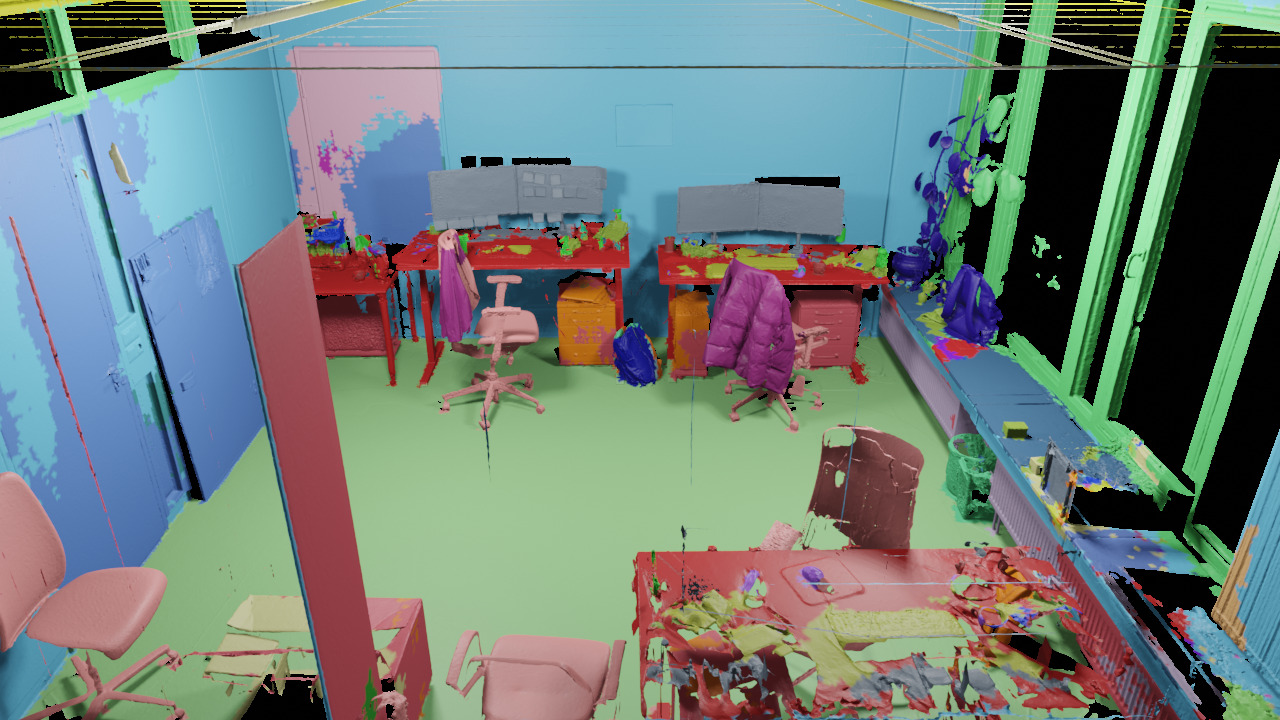} &
 \includegraphics[width=0.25\linewidth]{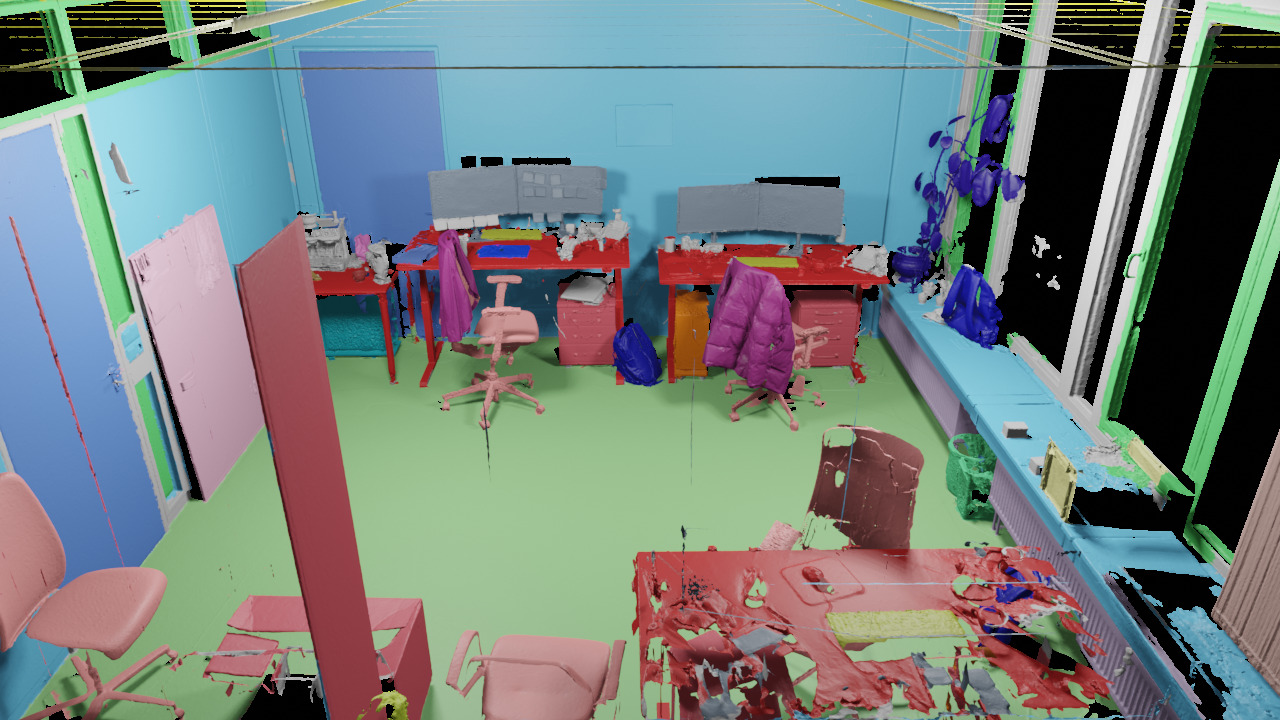} \\

  \includegraphics[width=0.25\linewidth]{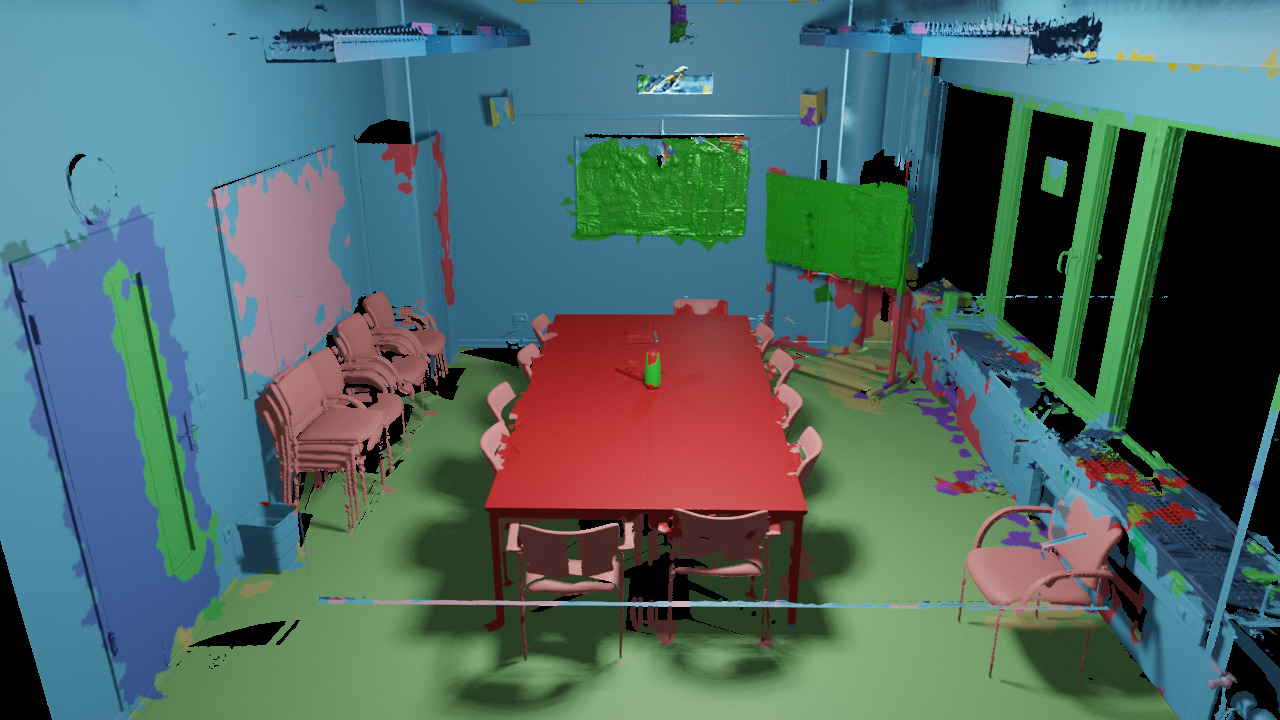} &
 \includegraphics[width=0.25\linewidth]{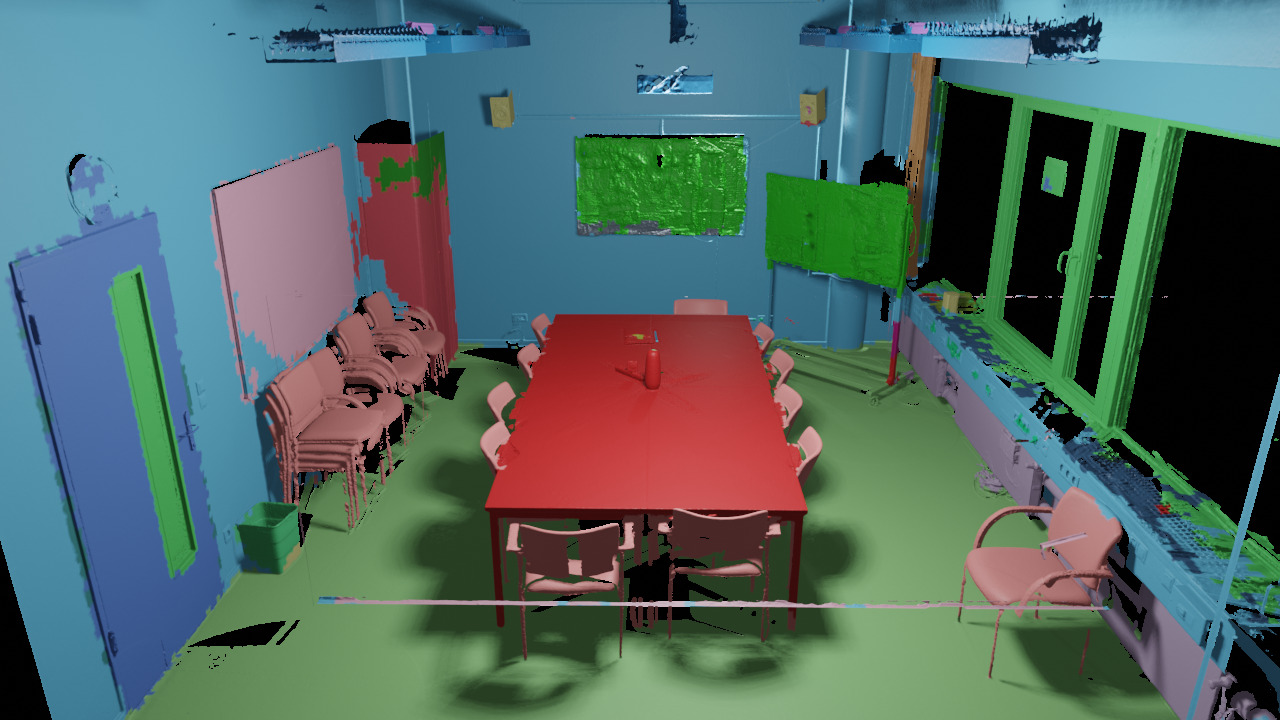} &
 \includegraphics[width=0.25\linewidth]{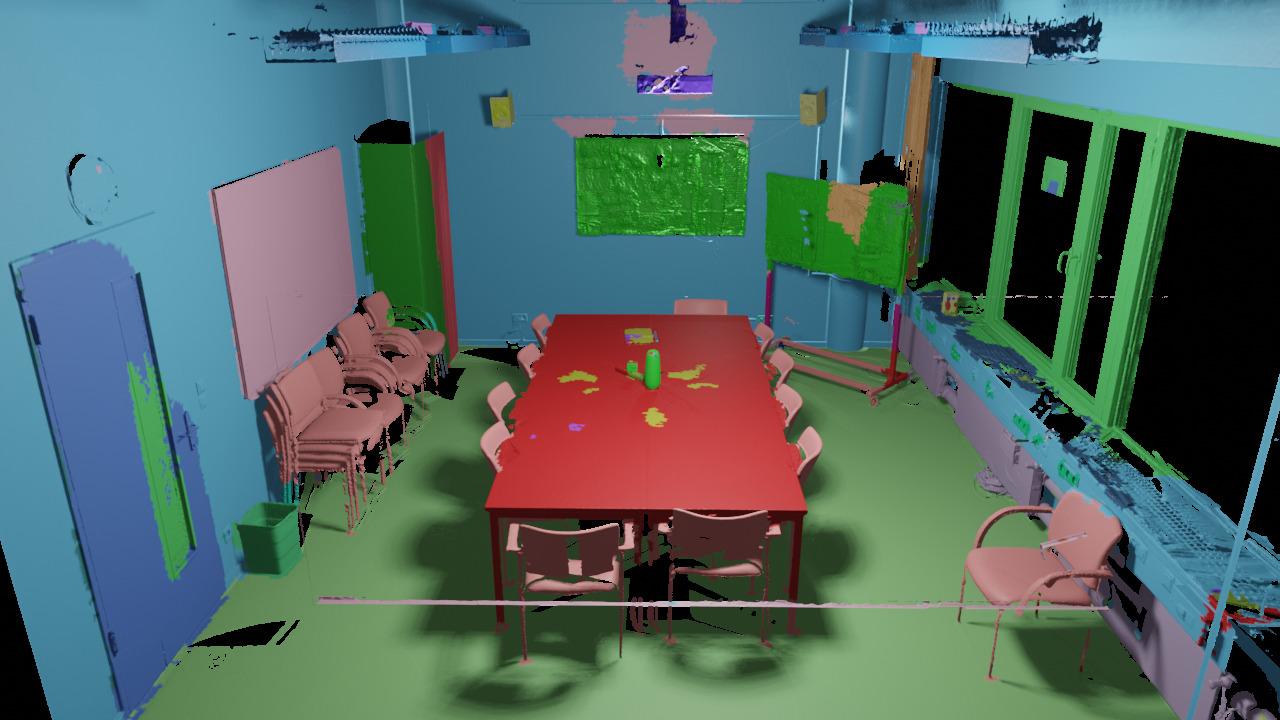} &
 \includegraphics[width=0.25\linewidth]{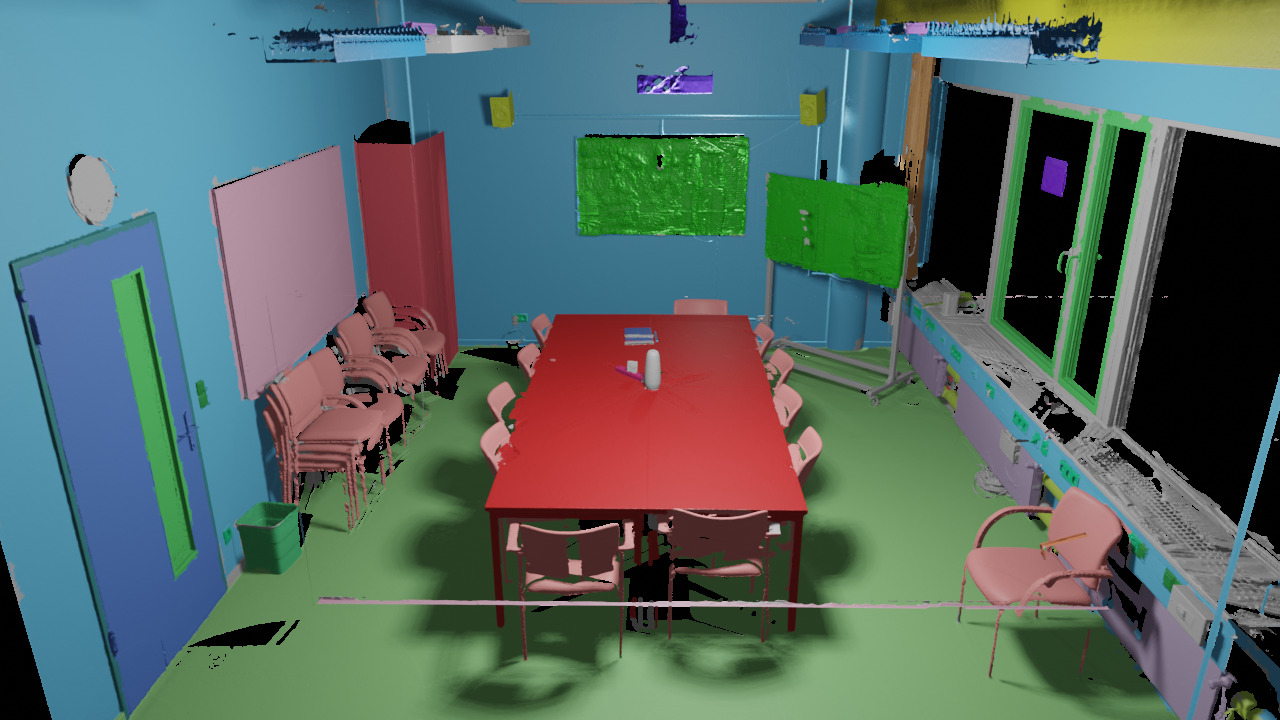} \\

 \includegraphics[width=0.25\linewidth]{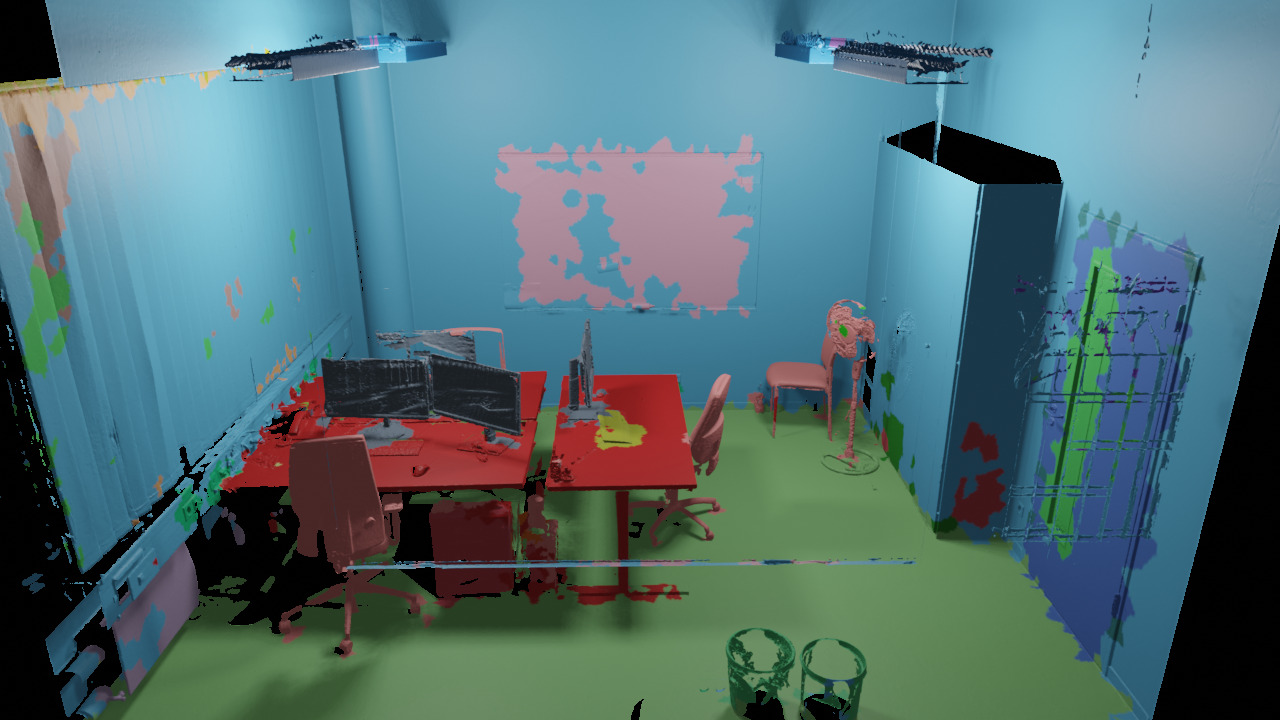} &
 \includegraphics[width=0.25\linewidth]{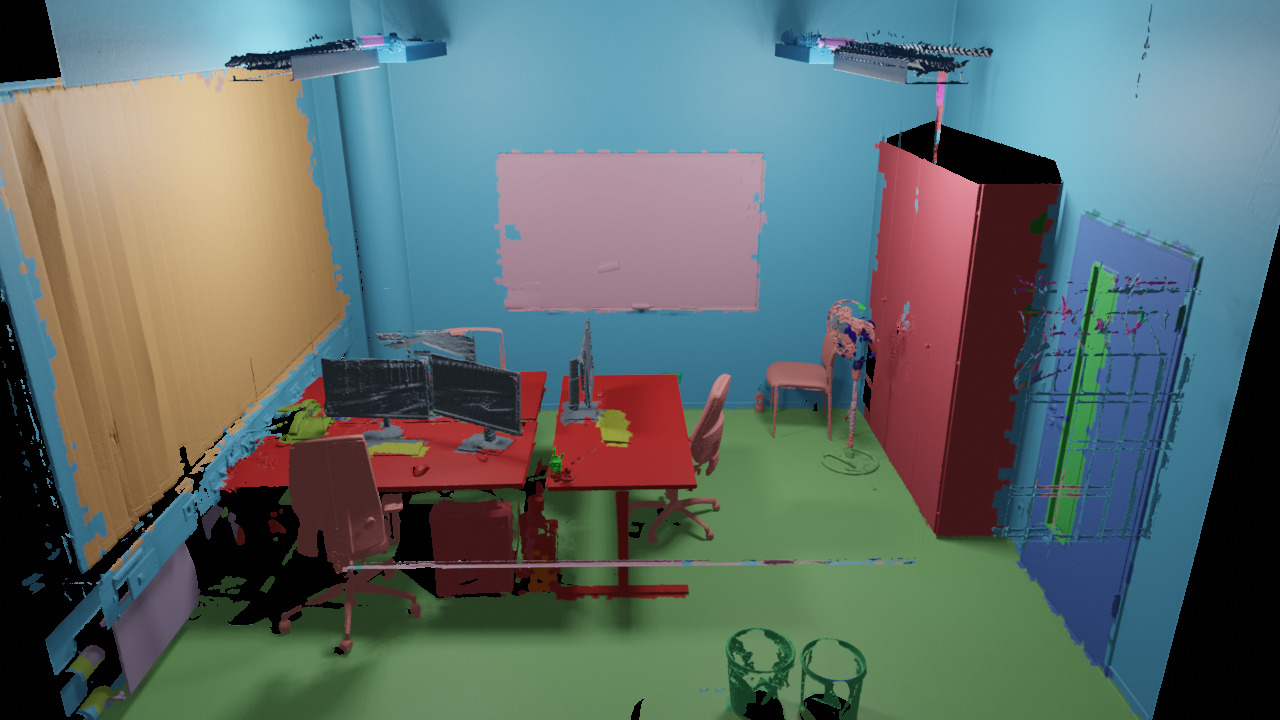} &
 \includegraphics[width=0.25\linewidth]{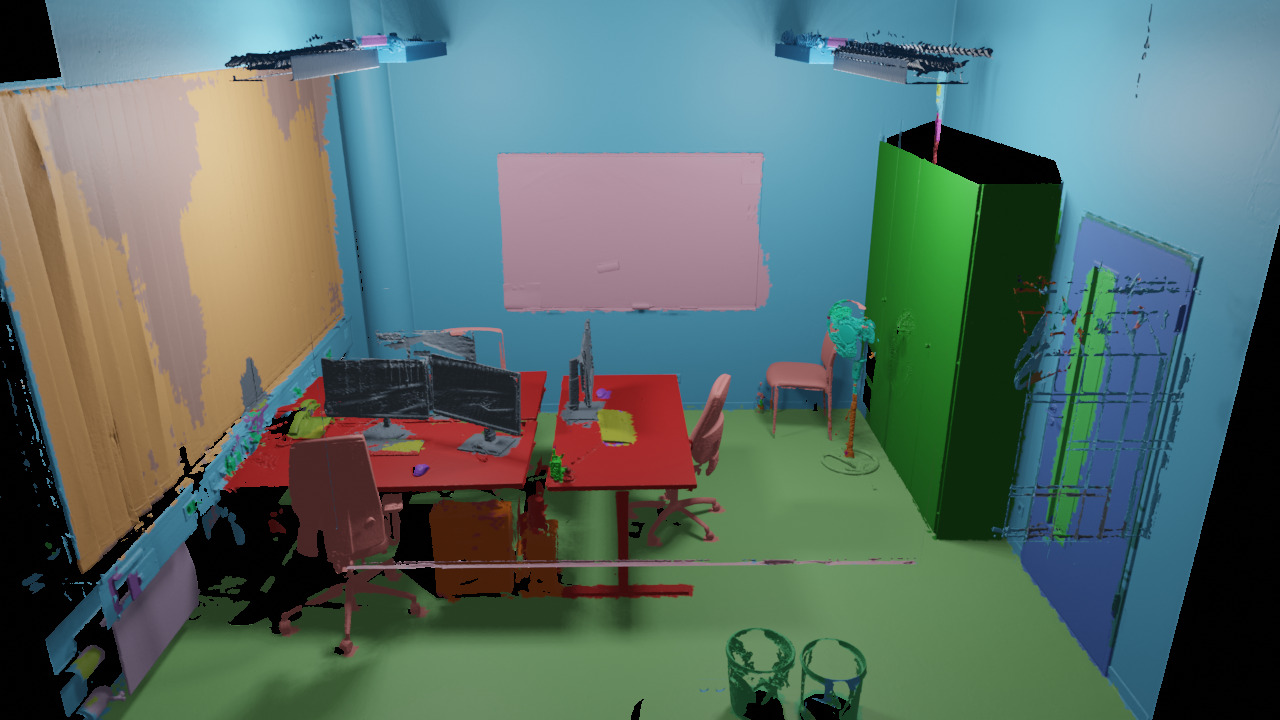} &
 \includegraphics[width=0.25\linewidth]{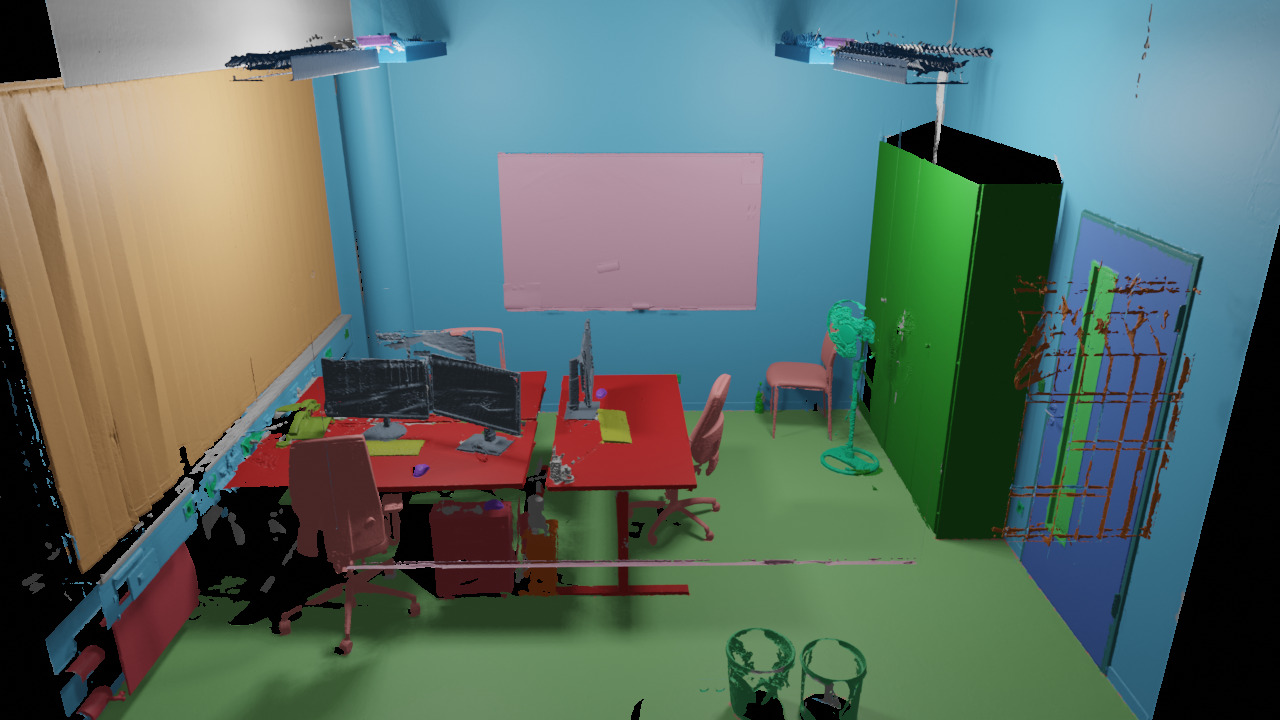} \\

 \includegraphics[width=0.25\linewidth]{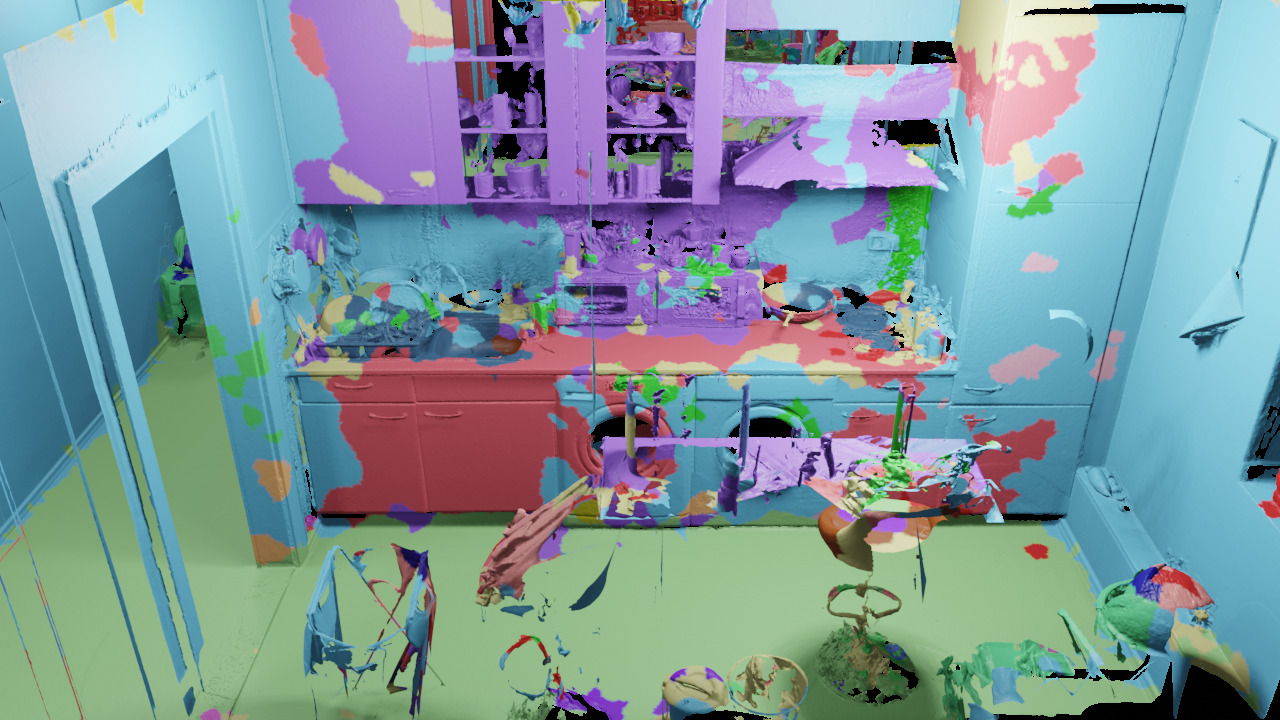} &
 \includegraphics[width=0.25\linewidth]{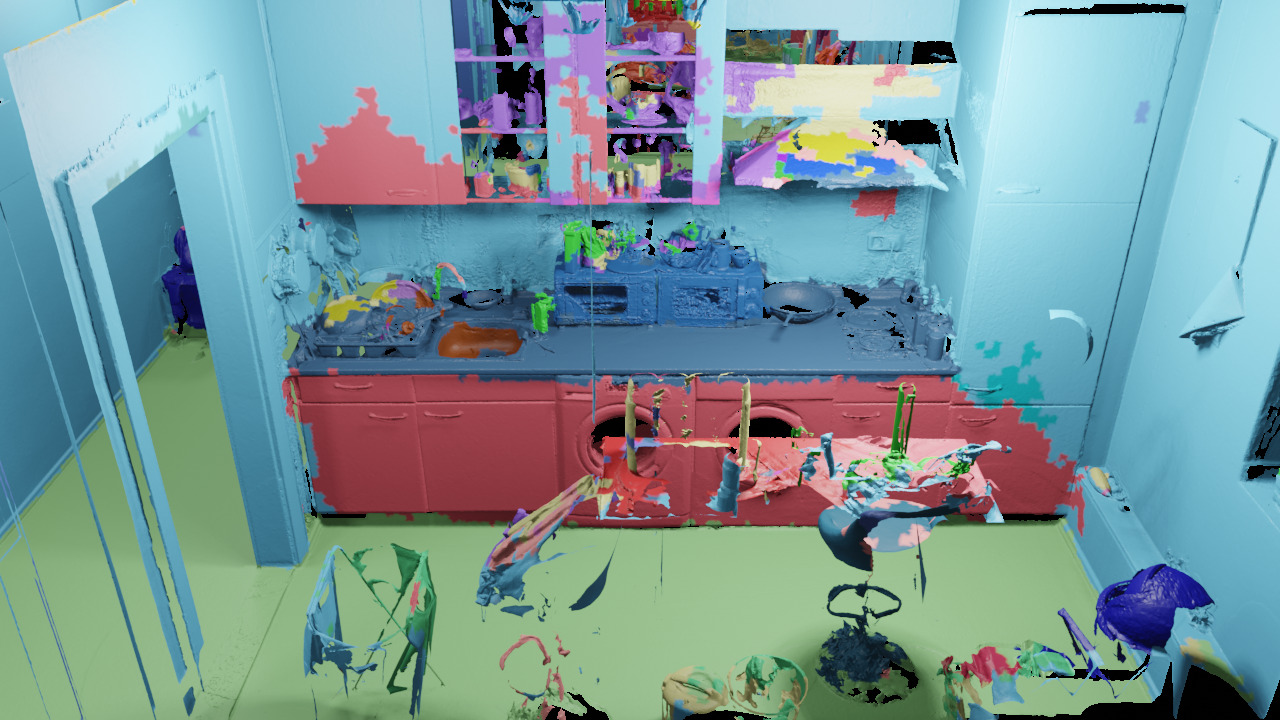} &
 \includegraphics[width=0.25\linewidth]{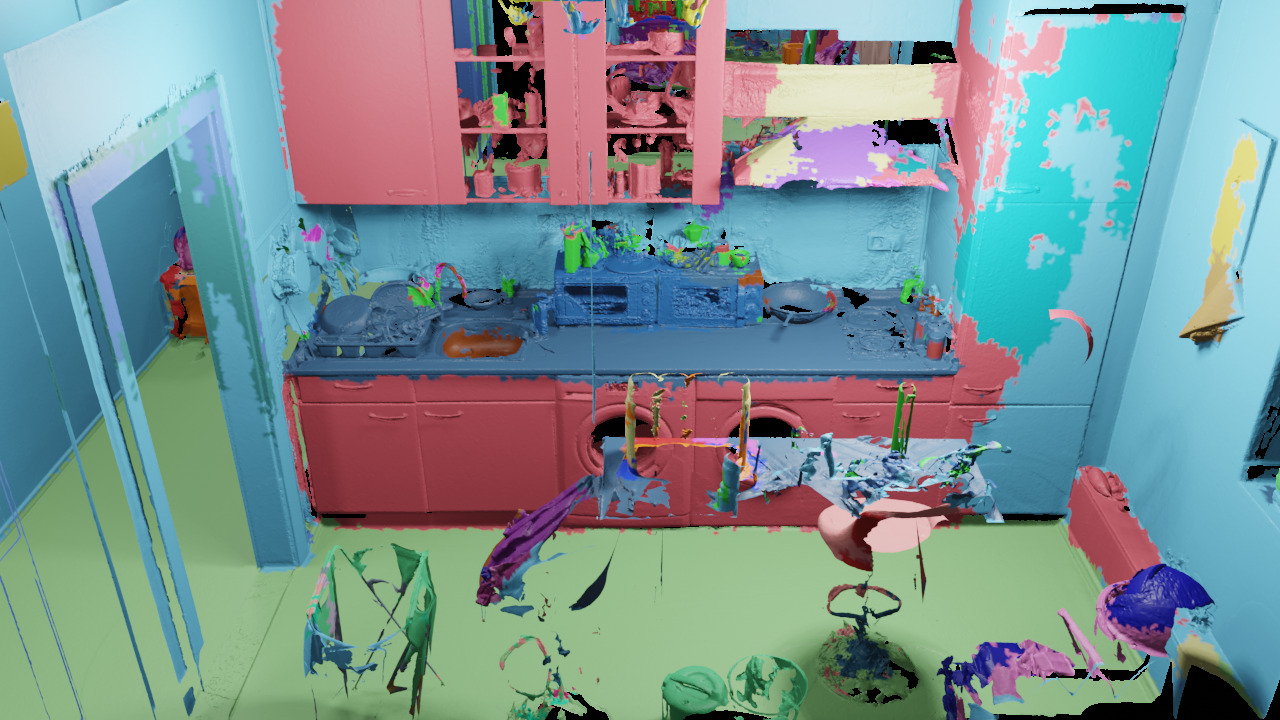} &
 \includegraphics[width=0.25\linewidth]{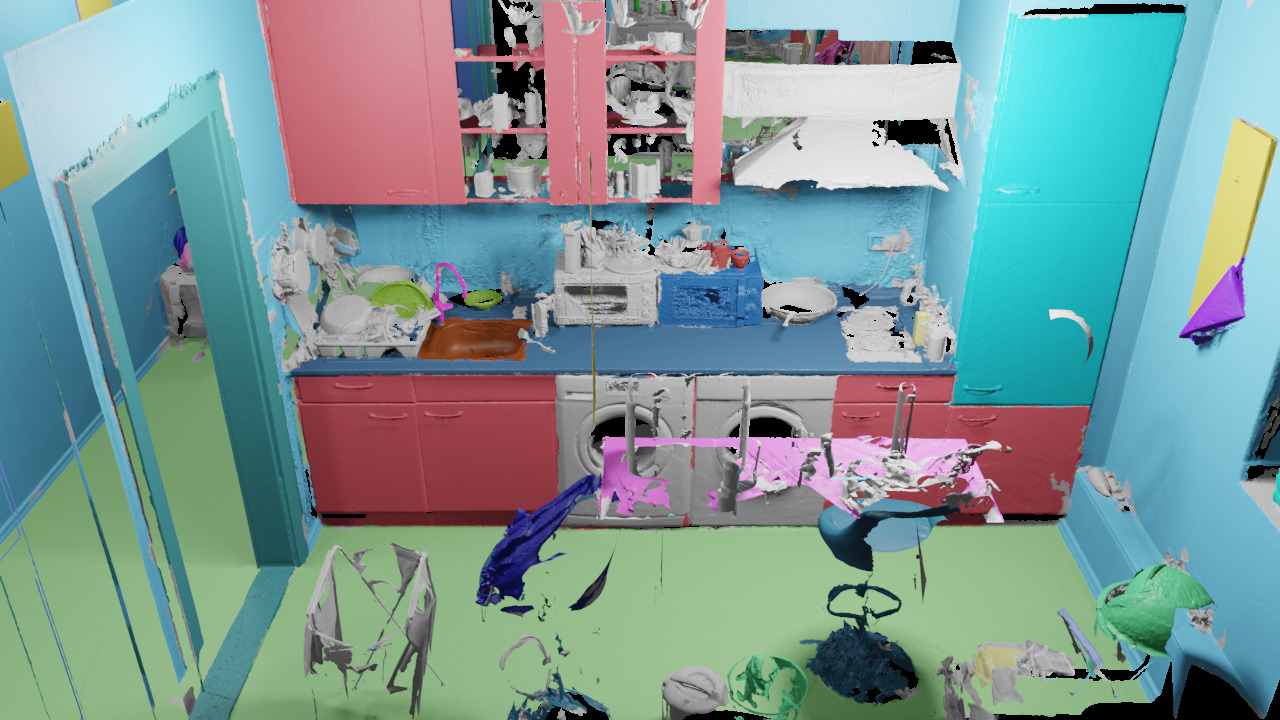} \\

\end{tabular}
\vspace{-0.2cm}
\caption{3D semantic segmentation baselines.}
\end{subfigure}

\vspace{0.2cm}

\begin{subfigure}{\linewidth}
\begin{tabular}{cccc}
 PointGroup &
 HAIS &
 SoftGroup &
 GT \\

\includegraphics[width=0.25\linewidth]{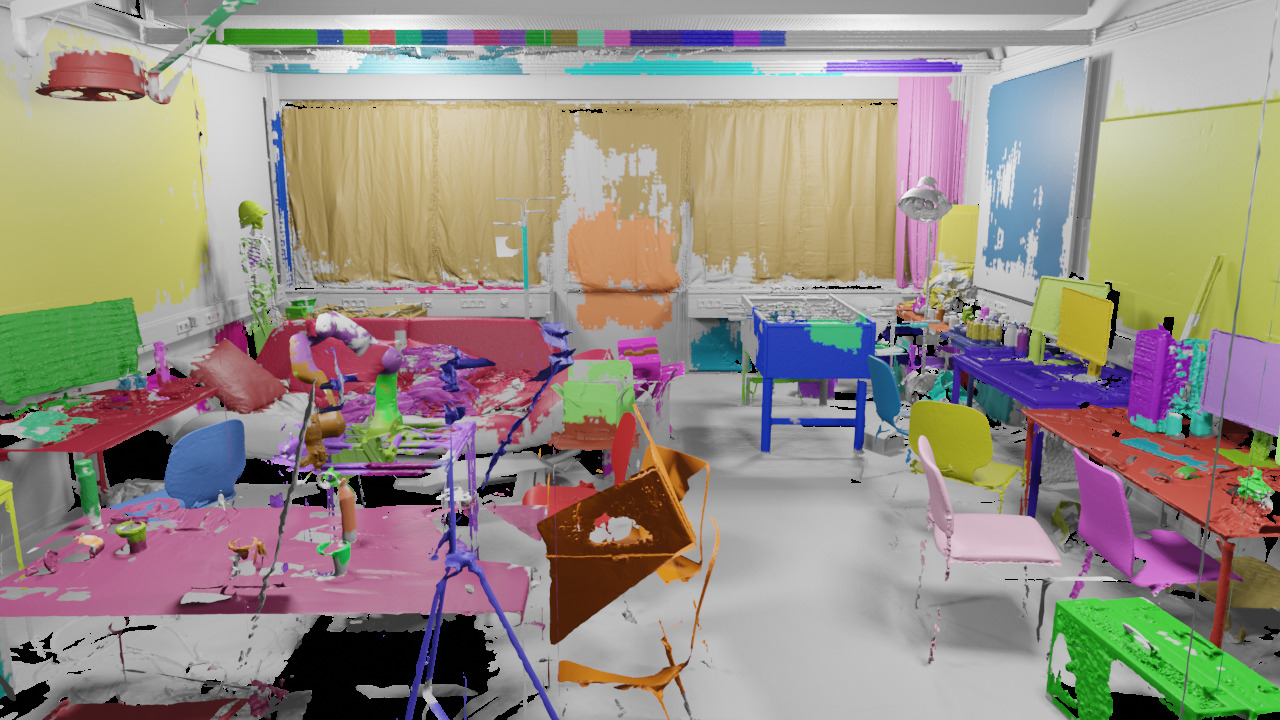} &
 \includegraphics[width=0.25\linewidth]{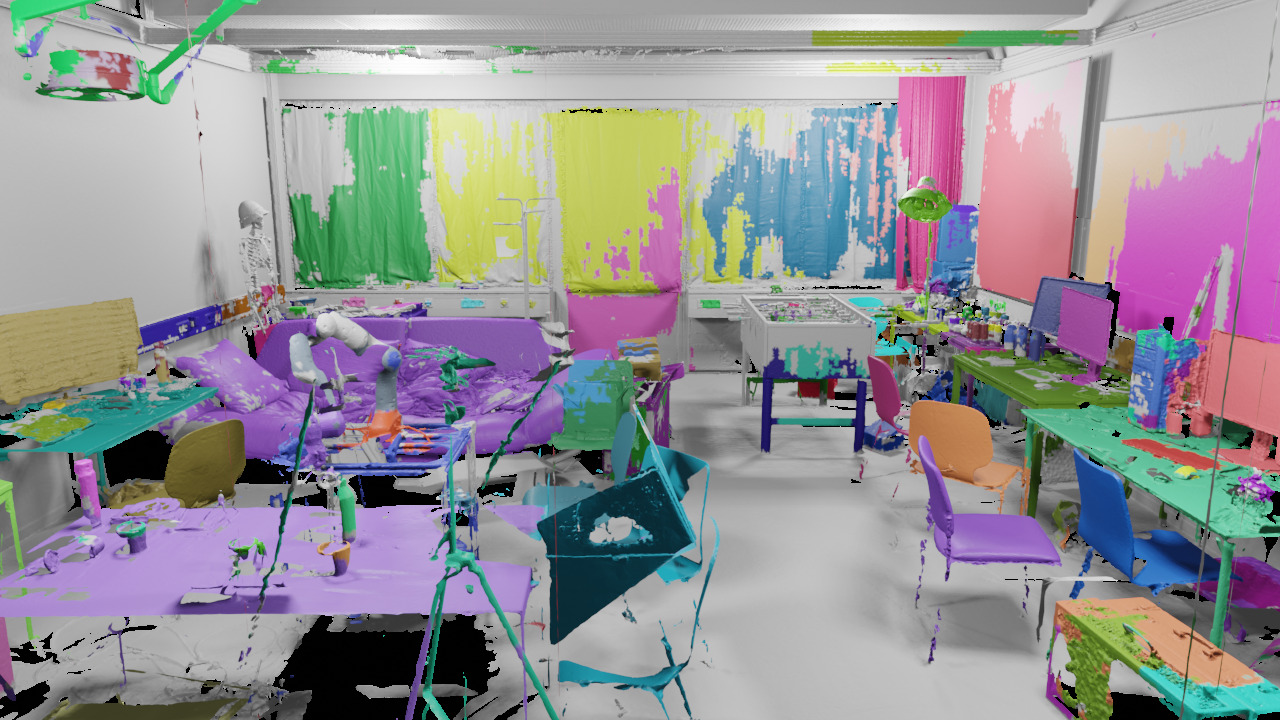} &
 \includegraphics[width=0.25\linewidth]{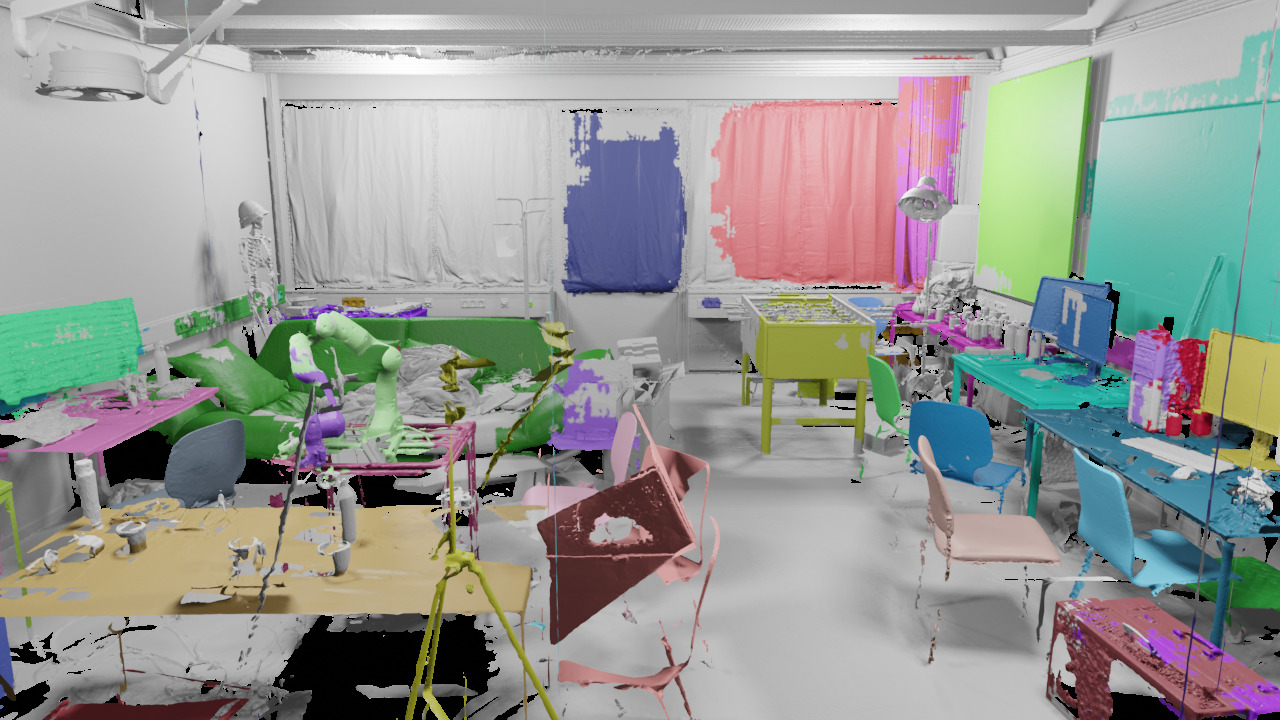} &
 \includegraphics[width=0.25\linewidth]{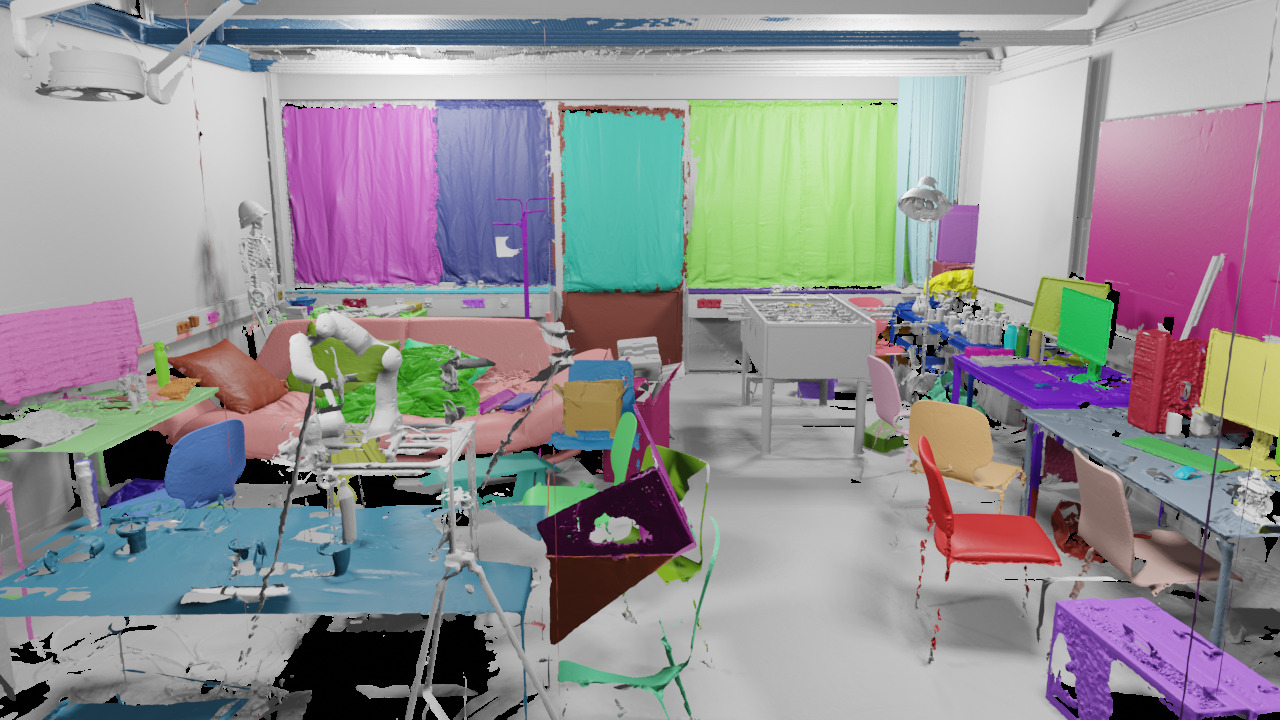} \\

 \includegraphics[width=0.25\linewidth]{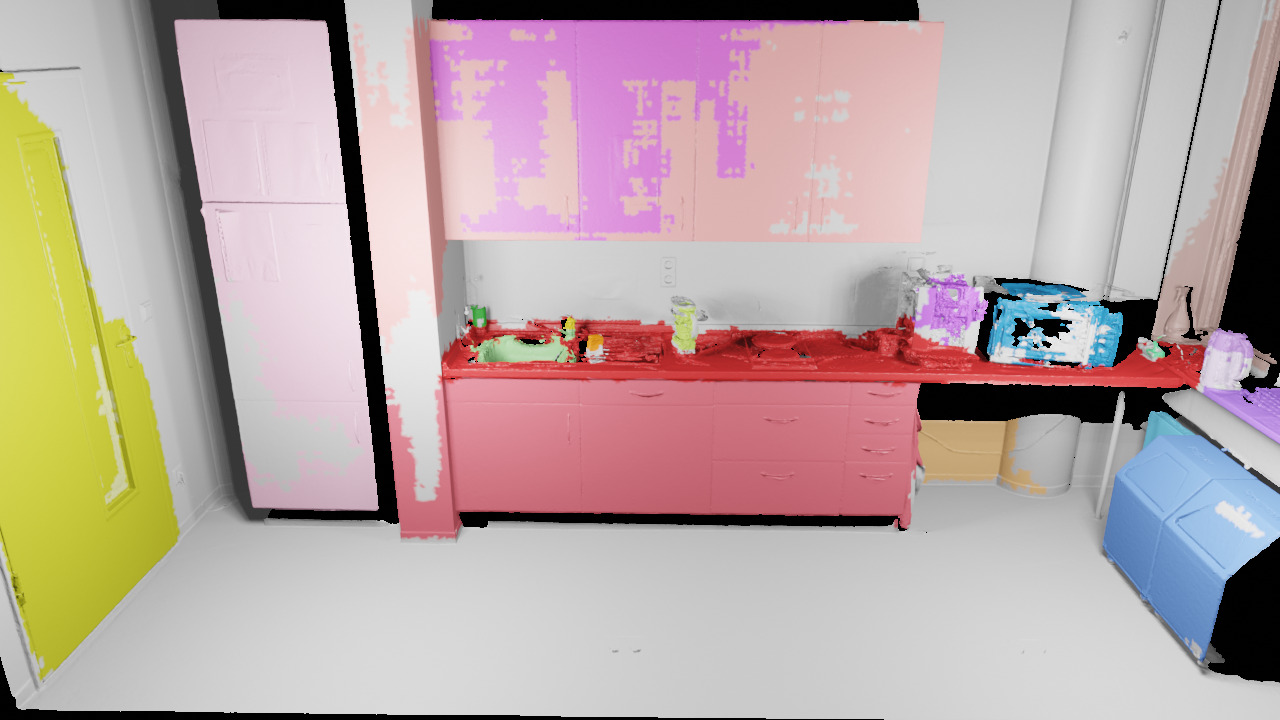} &
 \includegraphics[width=0.25\linewidth]{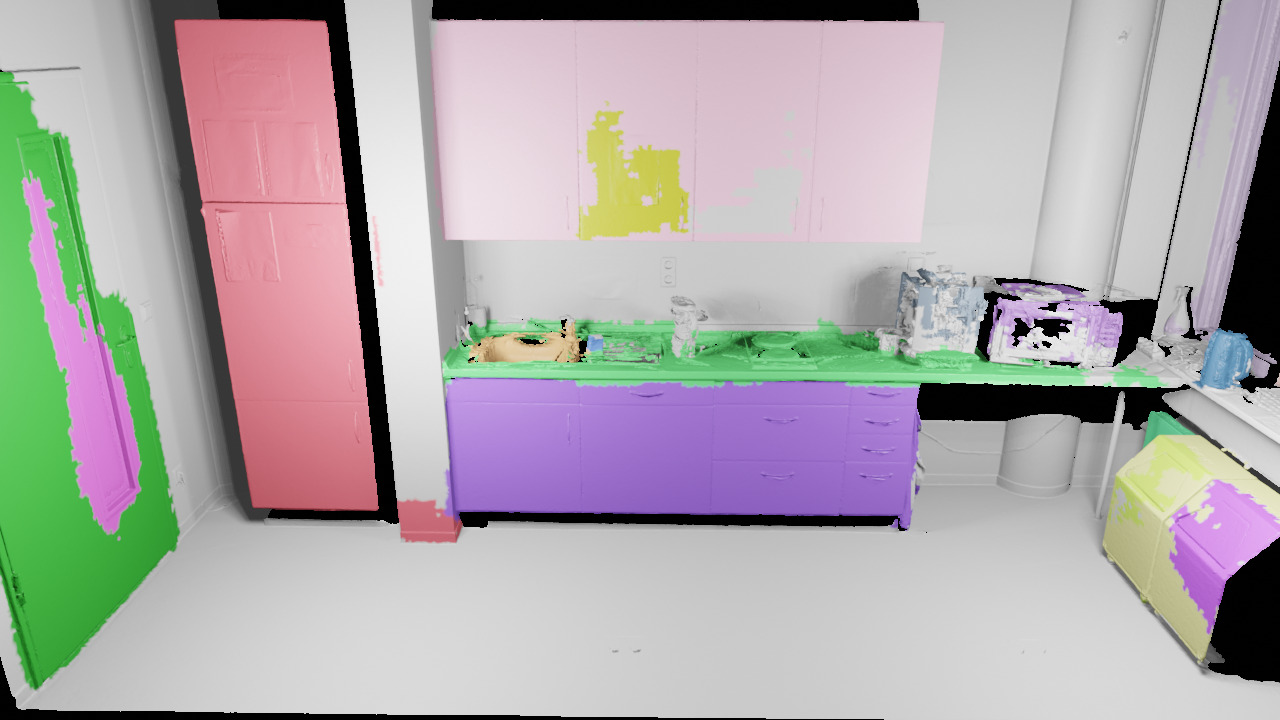} &
 \includegraphics[width=0.25\linewidth]{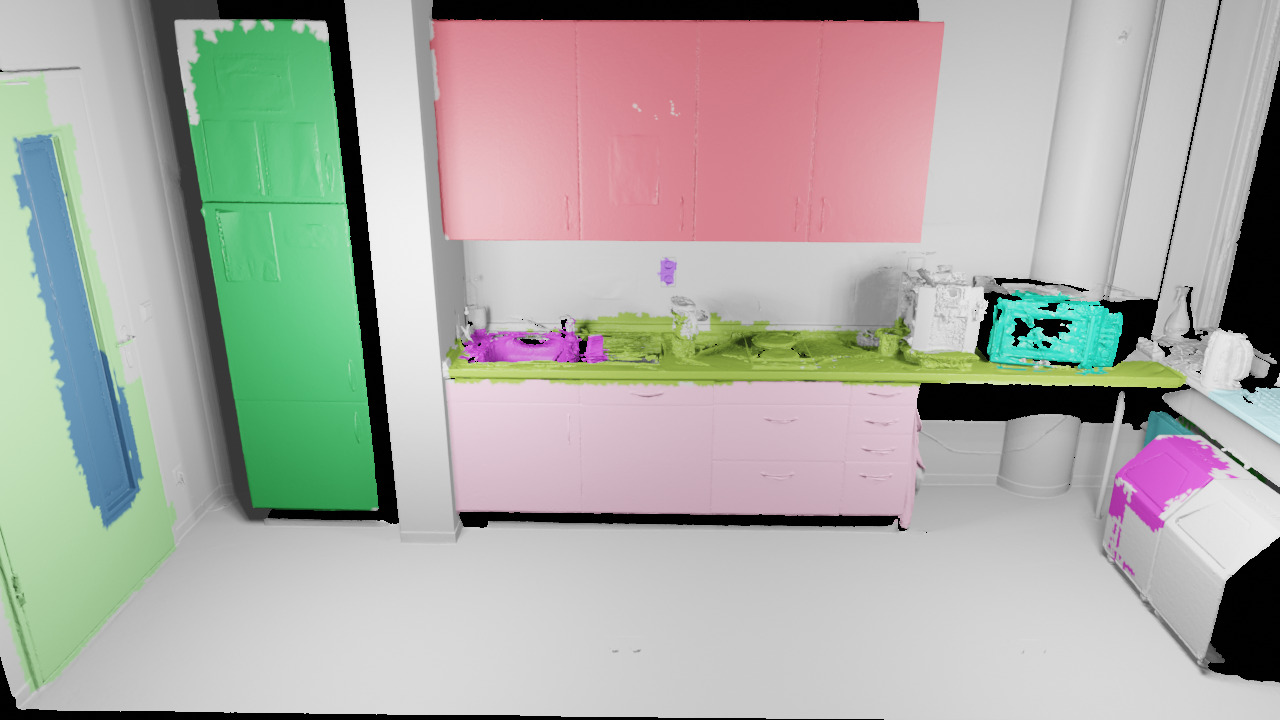} &
 \includegraphics[width=0.25\linewidth]{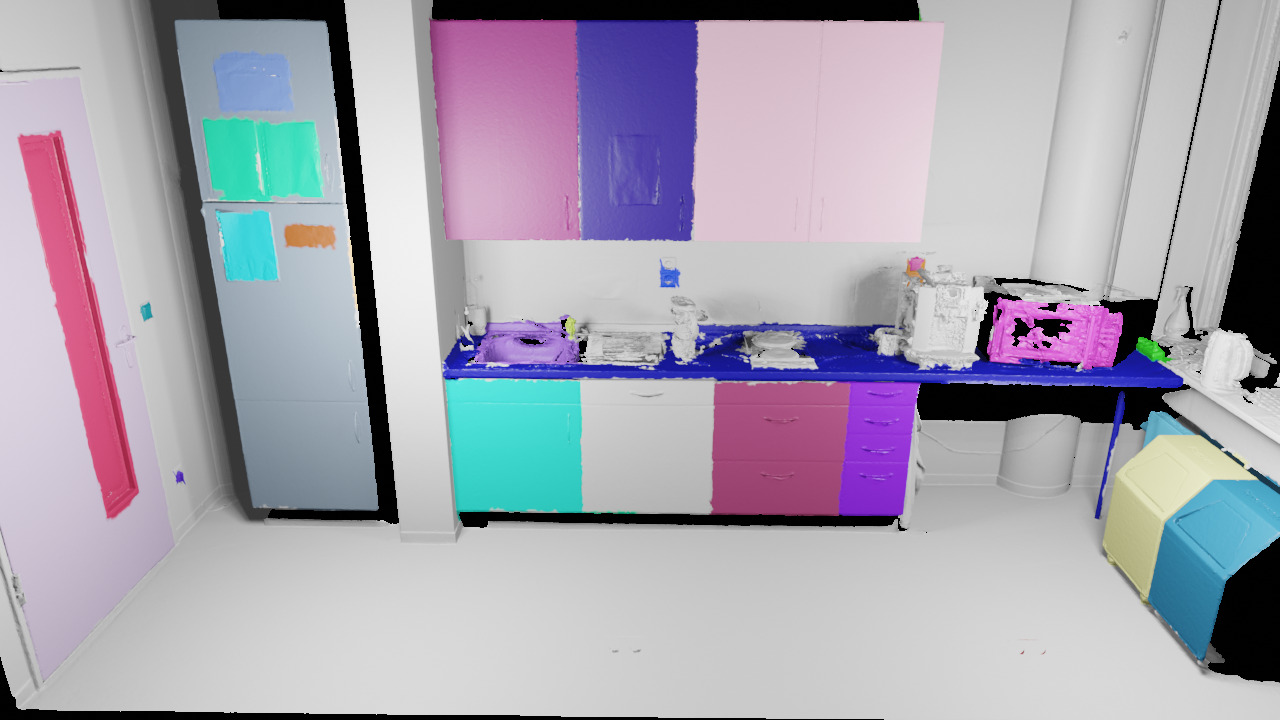} \\

\includegraphics[width=0.25\linewidth]{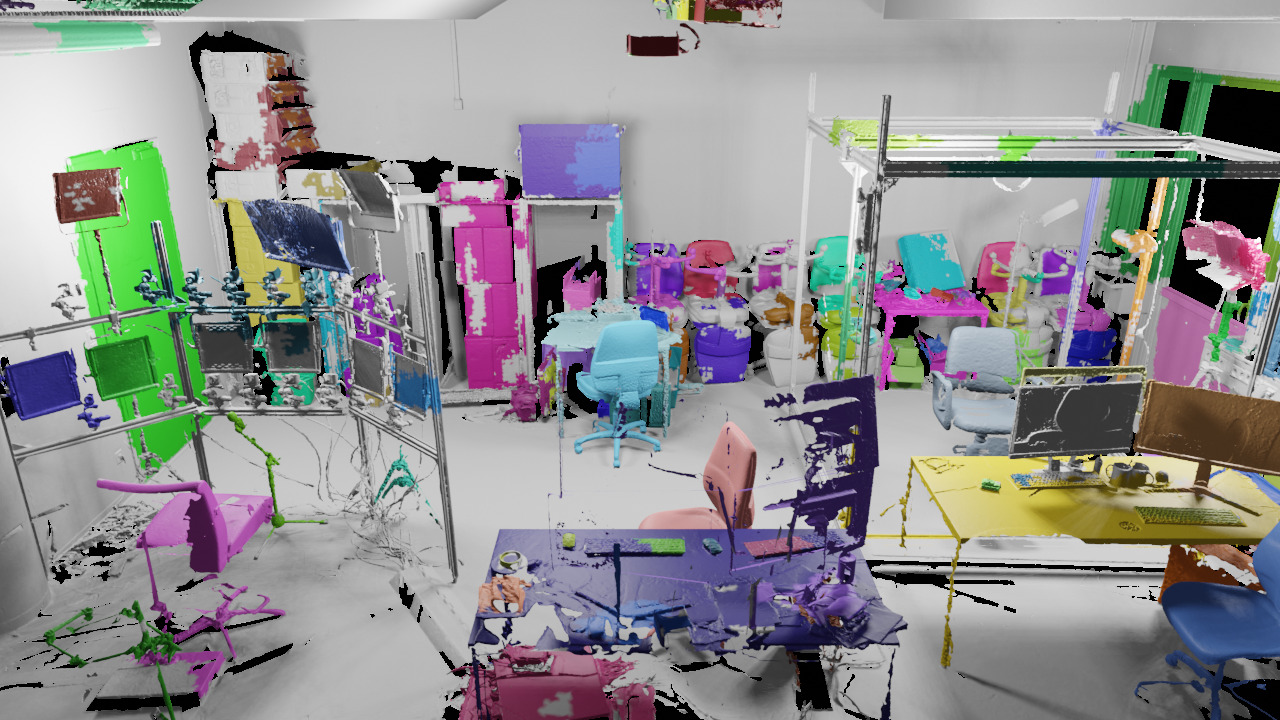} &
 \includegraphics[width=0.25\linewidth]{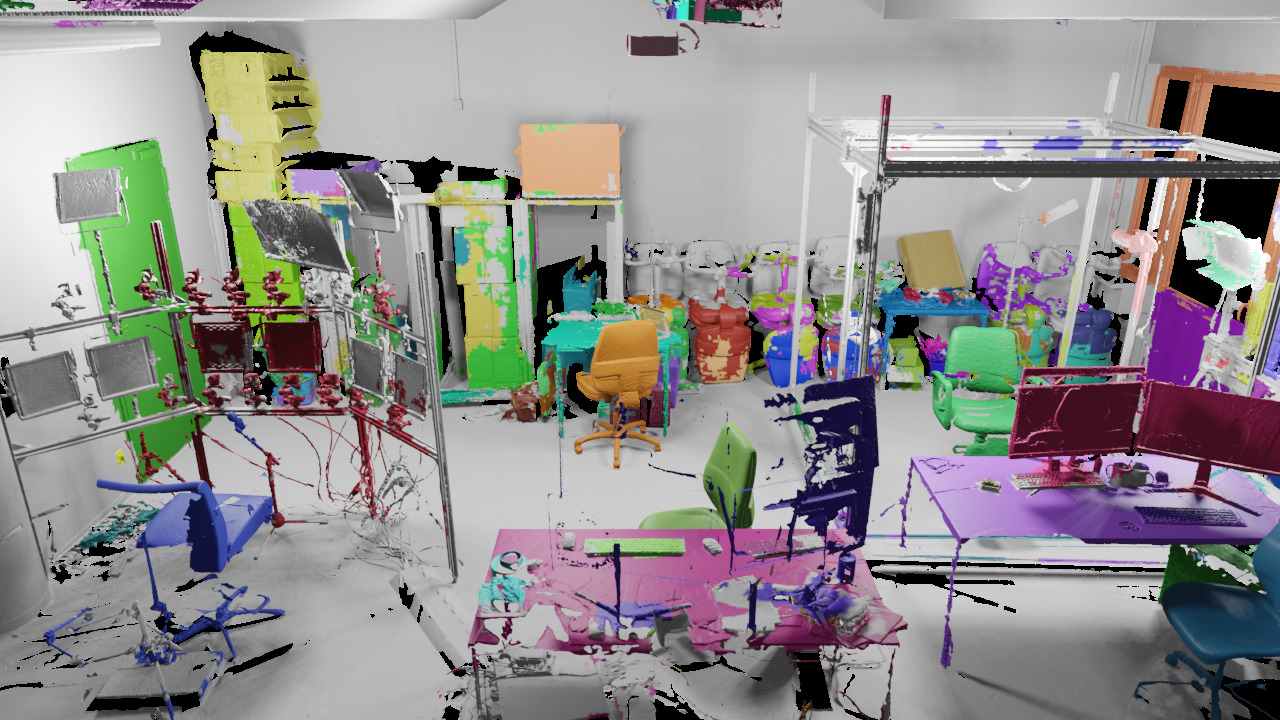} &
 \includegraphics[width=0.25\linewidth]{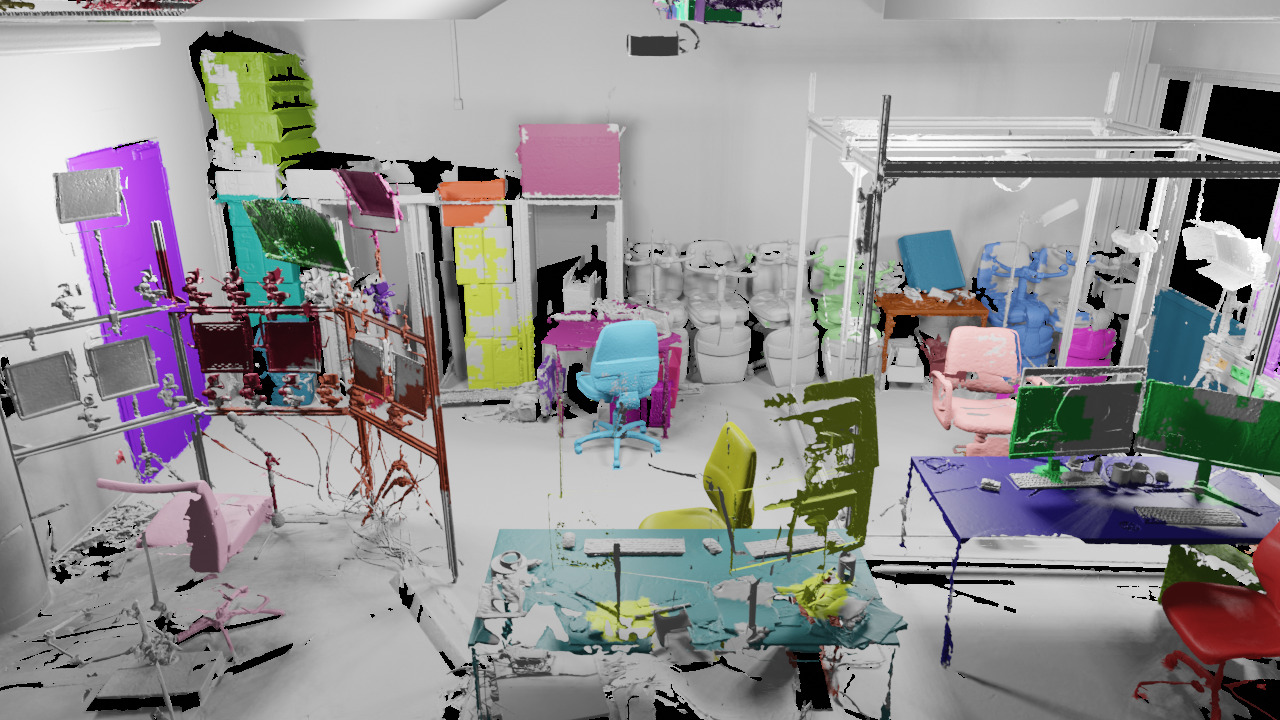} &
 \includegraphics[width=0.25\linewidth]{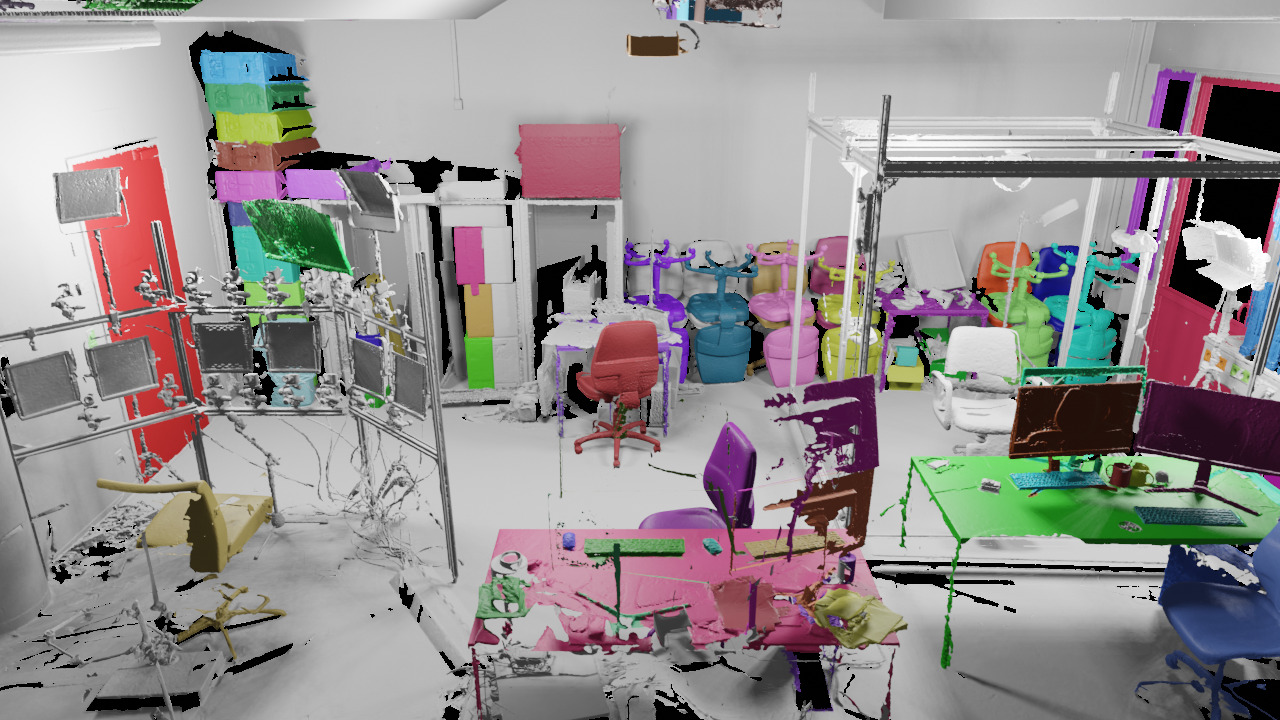} \\

 \includegraphics[width=0.25\linewidth]{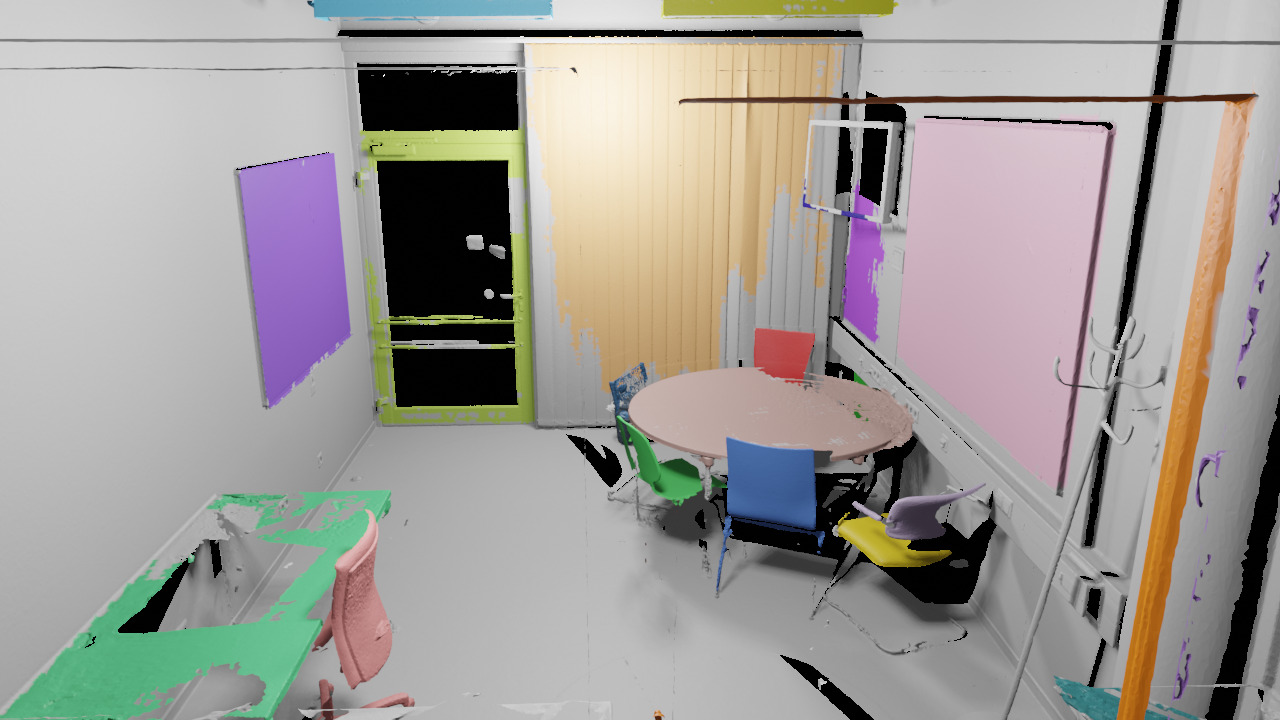} &
 \includegraphics[width=0.25\linewidth]{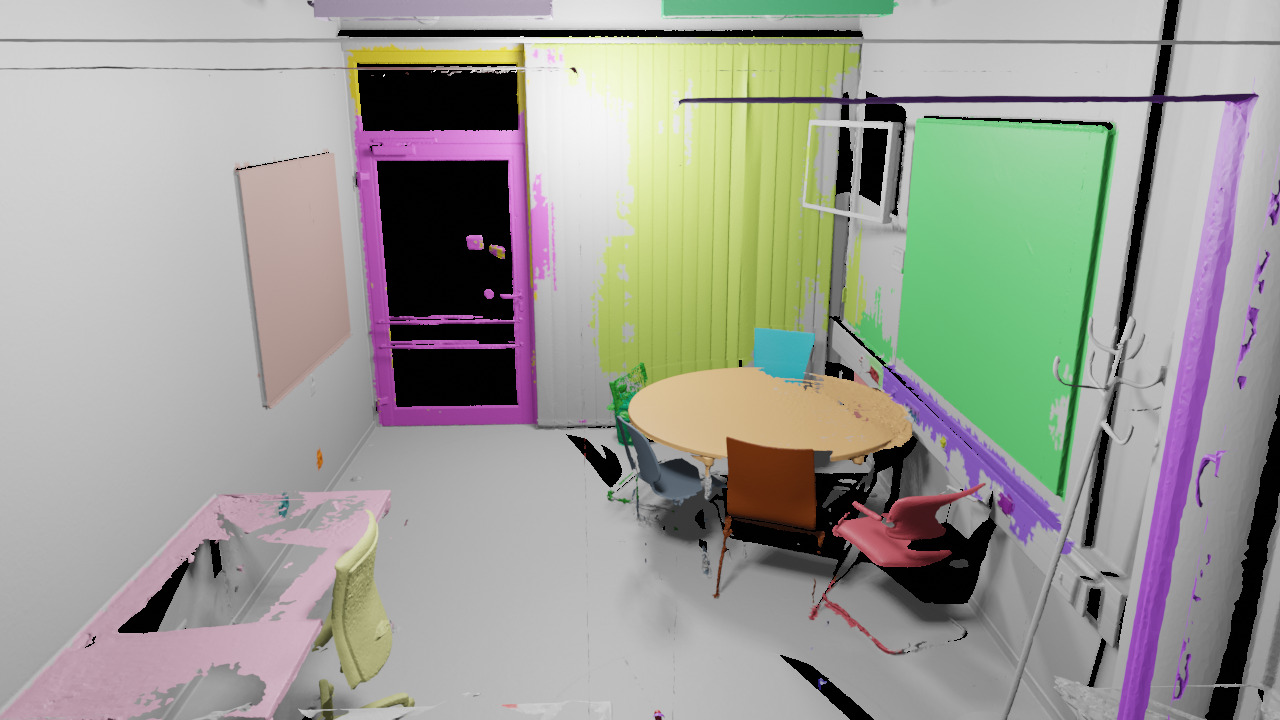} &
 \includegraphics[width=0.25\linewidth]{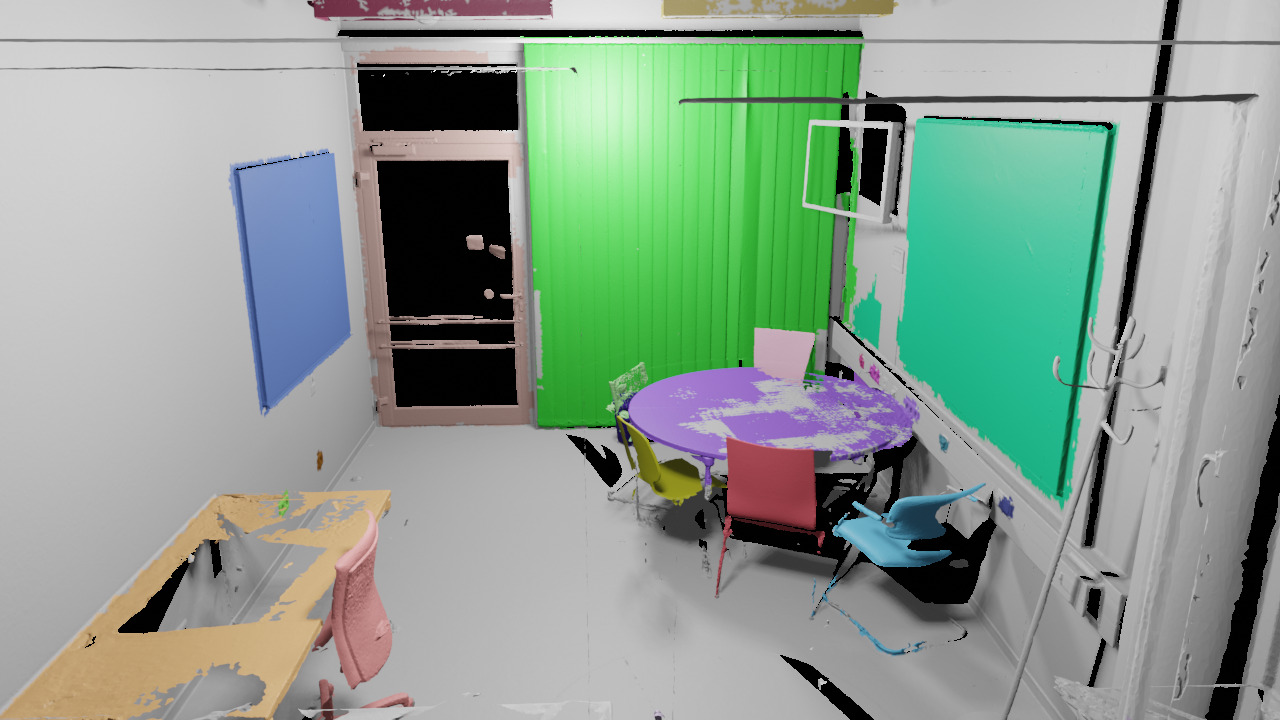} &
 \includegraphics[width=0.25\linewidth]{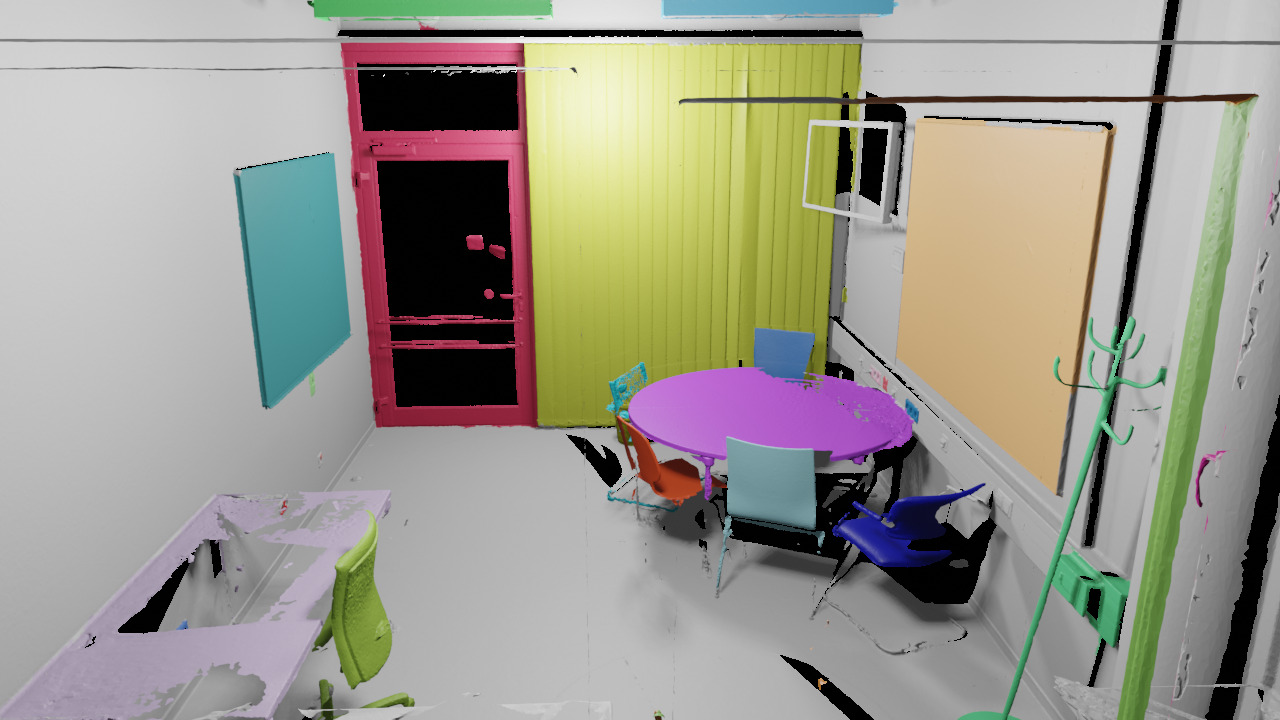} \\

\end{tabular}
\vspace{-0.25cm}
\caption{3D instance segmentation baselines.}
\end{subfigure}

\vspace{-0.2cm}
\caption{Qualitative results of 3D semantic and instance segmentation methods on the validation set of \OURS{}, showing diverse and cluttered scenes. All methods have notable room for improvement on small and ambiguous objects.}
\label{fig:semantic-qual-appendix}
\end{figure*}

\paragraph{Qualitative Results} Further qualitative results of semantic and instance segmentation baselines are shown in Fig. \ref{fig:semantic-qual-appendix}.

\section{Novel View Synthesis on iPhone Data}
\label{sec:app-nvs-iphone}
After training on iPhone data, we apply  color correction based on optimal transport between color distributions \cite{ferradans2014regularized,flamary2021pot} on the output in order to compare with the DSLR ground truth. The visual comparisons are shown in Fig.~\ref{fig:iphone-nerf}. 

%% file: 00_main_arxiv.bbl
\begin{thebibliography}{10}\itemsep=-1pt

\bibitem{barron2021mip}
Jonathan~T Barron, Ben Mildenhall, Matthew Tancik, Peter Hedman, Ricardo
  Martin-Brualla, and Pratul~P Srinivasan.
\newblock Mip-nerf: A multiscale representation for anti-aliasing neural
  radiance fields.
\newblock In {\em Proceedings of the IEEE/CVF International Conference on
  Computer Vision}, pages 5855--5864, 2021.

\bibitem{barron2022mip}
Jonathan~T Barron, Ben Mildenhall, Dor Verbin, Pratul~P Srinivasan, and Peter
  Hedman.
\newblock Mip-nerf 360: Unbounded anti-aliased neural radiance fields.
\newblock In {\em Proceedings of the IEEE/CVF Conference on Computer Vision and
  Pattern Recognition}, pages 5470--5479, 2022.

\bibitem{baruch2021arkitscenes}
Gilad Baruch, Zhuoyuan Chen, Afshin Dehghan, Tal Dimry, Yuri Feigin, Peter Fu,
  Thomas Gebauer, Brandon Joffe, Daniel Kurz, Arik Schwartz, et~al.
\newblock Arkitscenes--a diverse real-world dataset for 3d indoor scene
  understanding using mobile rgb-d data.
\newblock {\em arXiv preprint arXiv:2111.08897}, 2021.

\bibitem{chang2017matterport3d}
Angel Chang, Angela Dai, Thomas Funkhouser, Maciej Halber, Matthias Niessner,
  Manolis Savva, Shuran Song, Andy Zeng, and Yinda Zhang.
\newblock Matterport3d: Learning from rgb-d data in indoor environments.
\newblock {\em arXiv preprint arXiv:1709.06158}, 2017.

\bibitem{chen2022tensorf}
Anpei Chen, Zexiang Xu, Andreas Geiger, Jingyi Yu, and Hao Su.
\newblock Tensorf: Tensorial radiance fields.
\newblock In {\em Computer Vision--ECCV 2022: 17th European Conference, Tel
  Aviv, Israel, October 23--27, 2022, Proceedings, Part XXXII}, pages 333--350.
  Springer, 2022.

\bibitem{chen2021mvsnerf}
Anpei Chen, Zexiang Xu, Fuqiang Zhao, Xiaoshuai Zhang, Fanbo Xiang, Jingyi Yu,
  and Hao Su.
\newblock Mvsnerf: Fast generalizable radiance field reconstruction from
  multi-view stereo.
\newblock In {\em Proceedings of the IEEE/CVF International Conference on
  Computer Vision}, pages 14124--14133, 2021.

\bibitem{chen2021hierarchical}
Shaoyu Chen, Jiemin Fang, Qian Zhang, Wenyu Liu, and Xinggang Wang.
\newblock Hierarchical aggregation for 3d instance segmentation.
\newblock In {\em Proceedings of the IEEE/CVF International Conference on
  Computer Vision}, pages 15467--15476, 2021.

\bibitem{choy20194d}
Christopher Choy, JunYoung Gwak, and Silvio Savarese.
\newblock 4d spatio-temporal convnets: Minkowski convolutional neural networks.
\newblock In {\em Proceedings of the IEEE Conference on Computer Vision and
  Pattern Recognition}, pages 3075--3084, 2019.

\bibitem{dai2017scannet}
Angela Dai, Angel~X Chang, Manolis Savva, Maciej Halber, Thomas Funkhouser, and
  Matthias Nie{\ss}ner.
\newblock Scannet: Richly-annotated 3d reconstructions of indoor scenes.
\newblock In {\em Proceedings of the IEEE conference on computer vision and
  pattern recognition}, pages 5828--5839, 2017.

\bibitem{felzenszwalb2004efficient}
Pedro~F Felzenszwalb and Daniel~P Huttenlocher.
\newblock Efficient graph-based image segmentation.
\newblock {\em International journal of computer vision}, 59(2):167--181, 2004.

\bibitem{ferradans2014regularized}
Sira Ferradans, Nicolas Papadakis, Gabriel Peyr{\'e}, and Jean-Fran{\c{c}}ois
  Aujol.
\newblock Regularized discrete optimal transport.
\newblock {\em SIAM Journal on Imaging Sciences}, 7(3):1853--1882, 2014.

\bibitem{flamary2021pot}
R{\'e}mi Flamary, Nicolas Courty, Alexandre Gramfort, Mokhtar~Z. Alaya,
  Aur{\'e}lie Boisbunon, Stanislas Chambon, Laetitia Chapel, Adrien Corenflos,
  Kilian Fatras, Nemo Fournier, L{\'e}o Gautheron, Nathalie~T.H. Gayraud,
  Hicham Janati, Alain Rakotomamonjy, Ievgen Redko, Antoine Rolet, Antony
  Schutz, Vivien Seguy, Danica~J. Sutherland, Romain Tavenard, Alexander Tong,
  and Titouan Vayer.
\newblock Pot: Python optimal transport.
\newblock {\em Journal of Machine Learning Research}, 22(78):1--8, 2021.

\bibitem{fridovich2022plenoxels}
Sara Fridovich-Keil, Alex Yu, Matthew Tancik, Qinhong Chen, Benjamin Recht, and
  Angjoo Kanazawa.
\newblock Plenoxels: Radiance fields without neural networks.
\newblock In {\em Proceedings of the IEEE/CVF Conference on Computer Vision and
  Pattern Recognition}, pages 5501--5510, 2022.

\bibitem{fu2022panoptic}
Xiao Fu, Shangzhan Zhang, Tianrun Chen, Yichong Lu, Lanyun Zhu, Xiaowei Zhou,
  Andreas Geiger, and Yiyi Liao.
\newblock Panoptic nerf: 3d-to-2d label transfer for panoptic urban scene
  segmentation.
\newblock {\em arXiv preprint arXiv:2203.15224}, 2022.

\bibitem{garland1997surface}
Michael Garland and Paul~S Heckbert.
\newblock Surface simplification using quadric error metrics.
\newblock In {\em Proceedings of the 24th annual conference on Computer
  graphics and interactive techniques}, pages 209--216, 1997.

\bibitem{pix2pix2017}
Phillip Isola, Jun-Yan Zhu, Tinghui Zhou, and Alexei~A Efros.
\newblock Image-to-image translation with conditional adversarial networks.
\newblock {\em CVPR}, 2017.

\bibitem{jensen2014large}
Rasmus Jensen, Anders Dahl, George Vogiatzis, Engin Tola, and Henrik Aan{\ae}s.
\newblock Large scale multi-view stereopsis evaluation.
\newblock In {\em Proceedings of the IEEE conference on computer vision and
  pattern recognition}, pages 406--413, 2014.

\bibitem{jiang2020pointgroup}
Li Jiang, Hengshuang Zhao, Shaoshuai Shi, Shu Liu, Chi-Wing Fu, and Jiaya Jia.
\newblock Pointgroup: Dual-set point grouping for 3d instance segmentation.
\newblock In {\em Proceedings of the IEEE/CVF conference on computer vision and
  Pattern recognition}, pages 4867--4876, 2020.

\bibitem{johari2022geonerf}
Mohammad~Mahdi Johari, Yann Lepoittevin, and Fran{\c{c}}ois Fleuret.
\newblock Geonerf: Generalizing nerf with geometry priors.
\newblock In {\em Proceedings of the IEEE/CVF Conference on Computer Vision and
  Pattern Recognition}, pages 18365--18375, 2022.

\bibitem{kazhdan2006poisson}
Michael Kazhdan, Matthew Bolitho, and Hugues Hoppe.
\newblock Poisson surface reconstruction.
\newblock In {\em Proceedings of the fourth Eurographics symposium on Geometry
  processing}, volume~7, page~0, 2006.

\bibitem{kazhdan2013screened}
Michael Kazhdan and Hugues Hoppe.
\newblock Screened poisson surface reconstruction.
\newblock {\em ACM Transactions on Graphics (ToG)}, 32(3):1--13, 2013.

\bibitem{knapitsch2017tanks}
Arno Knapitsch, Jaesik Park, Qian-Yi Zhou, and Vladlen Koltun.
\newblock Tanksf and temples: Benchmarking large-scale scene reconstruction.
\newblock {\em ACM Transactions on Graphics (ToG)}, 36(4):1--13, 2017.

\bibitem{kundu2022panoptic}
Abhijit Kundu, Kyle Genova, Xiaoqi Yin, Alireza Fathi, Caroline Pantofaru,
  Leonidas~J Guibas, Andrea Tagliasacchi, Frank Dellaert, and Thomas
  Funkhouser.
\newblock Panoptic neural fields: A semantic object-aware neural scene
  representation.
\newblock In {\em Proceedings of the IEEE/CVF Conference on Computer Vision and
  Pattern Recognition}, pages 12871--12881, 2022.

\bibitem{lin2014microsoft}
Tsung-Yi Lin, Michael Maire, Serge Belongie, James Hays, Pietro Perona, Deva
  Ramanan, Piotr Doll{\'a}r, and C~Lawrence Zitnick.
\newblock Microsoft coco: Common objects in context.
\newblock In {\em Computer Vision--ECCV 2014: 13th European Conference, Zurich,
  Switzerland, September 6-12, 2014, Proceedings, Part V 13}, pages 740--755.
  Springer, 2014.

\bibitem{liu2020neural}
Lingjie Liu, Jiatao Gu, Kyaw Zaw~Lin, Tat-Seng Chua, and Christian Theobalt.
\newblock Neural sparse voxel fields.
\newblock {\em Advances in Neural Information Processing Systems},
  33:15651--15663, 2020.

\bibitem{liu2022neural}
Yuan Liu, Sida Peng, Lingjie Liu, Qianqian Wang, Peng Wang, Christian Theobalt,
  Xiaowei Zhou, and Wenping Wang.
\newblock Neural rays for occlusion-aware image-based rendering.
\newblock In {\em Proceedings of the IEEE/CVF Conference on Computer Vision and
  Pattern Recognition}, pages 7824--7833, 2022.

\bibitem{martin2021nerf}
Ricardo Martin-Brualla, Noha Radwan, Mehdi~SM Sajjadi, Jonathan~T Barron,
  Alexey Dosovitskiy, and Daniel Duckworth.
\newblock Nerf in the wild: Neural radiance fields for unconstrained photo
  collections.
\newblock In {\em Proceedings of the IEEE/CVF Conference on Computer Vision and
  Pattern Recognition}, pages 7210--7219, 2021.

\bibitem{mildenhall2019llff}
Ben Mildenhall, Pratul~P. Srinivasan, Rodrigo Ortiz-Cayon, Nima~Khademi
  Kalantari, Ravi Ramamoorthi, Ren Ng, and Abhishek Kar.
\newblock Local light field fusion: Practical view synthesis with prescriptive
  sampling guidelines.
\newblock {\em ACM Transactions on Graphics (TOG)}, 2019.

\bibitem{mildenhall2020nerf}
Ben Mildenhall, Pratul~P. Srinivasan, Matthew Tancik, Jonathan~T. Barron, Ravi
  Ramamoorthi, and Ren Ng.
\newblock Nerf: Representing scenes as neural radiance fields for view
  synthesis.
\newblock In {\em ECCV}, 2020.

\bibitem{muller2022diffrf}
Norman M{\"u}ller, Yawar Siddiqui, Lorenzo Porzi, Samuel~Rota Bul{\`o}, Peter
  Kontschieder, and Matthias Nie{\ss}ner.
\newblock Diffrf: Rendering-guided 3d radiance field diffusion.
\newblock {\em arXiv preprint arXiv:2212.01206}, 2022.

\bibitem{muller2022instant}
Thomas M{\"u}ller, Alex Evans, Christoph Schied, and Alexander Keller.
\newblock Instant neural graphics primitives with a multiresolution hash
  encoding.
\newblock {\em ACM Transactions on Graphics (ToG)}, 41(4):1--15, 2022.

\bibitem{qi2017pointnet}
Charles~R Qi, Hao Su, Kaichun Mo, and Leonidas~J Guibas.
\newblock Pointnet: Deep learning on point sets for 3d classification and
  segmentation.
\newblock In {\em Proceedings of the IEEE conference on computer vision and
  pattern recognition}, pages 652--660, 2017.

\bibitem{qi2017pointnet++}
Charles~Ruizhongtai Qi, Li Yi, Hao Su, and Leonidas~J Guibas.
\newblock Pointnet++: Deep hierarchical feature learning on point sets in a
  metric space.
\newblock {\em Advances in neural information processing systems}, 30, 2017.

\bibitem{rozenberszki2022language}
David Rozenberszki, Or Litany, and Angela Dai.
\newblock Language-grounded indoor 3d semantic segmentation in the wild.
\newblock In {\em Computer Vision--ECCV 2022: 17th European Conference, Tel
  Aviv, Israel, October 23--27, 2022, Proceedings, Part XXXIII}, pages
  125--141. Springer, 2022.

\bibitem{schoenberger2016sfm}
Johannes~Lutz Sch\"{o}nberger and Jan-Michael Frahm.
\newblock Structure-from-motion revisited.
\newblock In {\em Conference on Computer Vision and Pattern Recognition
  (CVPR)}, 2016.

\bibitem{schoenberger2016mvs}
Johannes~Lutz Sch\"{o}nberger, Enliang Zheng, Marc Pollefeys, and Jan-Michael
  Frahm.
\newblock Pixelwise view selection for unstructured multi-view stereo.
\newblock In {\em European Conference on Computer Vision (ECCV)}, 2016.

\bibitem{eth3d}
Thomas Schops, Johannes~L Schonberger, Silvano Galliani, Torsten Sattler,
  Konrad Schindler, Marc Pollefeys, and Andreas Geiger.
\newblock A multi-view stereo benchmark with high-resolution images and
  multi-camera videos.
\newblock In {\em Proceedings of the IEEE Conference on Computer Vision and
  Pattern Recognition}, pages 3260--3269, 2017.

\bibitem{siddiqui2022panoptic}
Yawar Siddiqui, Lorenzo Porzi, Samuel~Rota Bul{\'o}, Norman M{\"u}ller,
  Matthias Nie{\ss}ner, Angela Dai, and Peter Kontschieder.
\newblock Panoptic lifting for 3d scene understanding with neural fields.
\newblock {\em arXiv preprint arXiv:2212.09802}, 2022.

\bibitem{silberman2011indoor}
Nathan Silberman and Rob Fergus.
\newblock Indoor scene segmentation using a structured light sensor.
\newblock In {\em 2011 IEEE international conference on computer vision
  workshops (ICCV workshops)}, pages 601--608. IEEE, 2011.

\bibitem{song2015sun}
Shuran Song, Samuel~P Lichtenberg, and Jianxiong Xiao.
\newblock Sun rgb-d: A rgb-d scene understanding benchmark suite.
\newblock In {\em Proceedings of the IEEE conference on computer vision and
  pattern recognition}, pages 567--576, 2015.

\bibitem{suhail2022generalizable}
Mohammed Suhail, Carlos Esteves, Leonid Sigal, and Ameesh Makadia.
\newblock Generalizable patch-based neural rendering.
\newblock In {\em Computer Vision--ECCV 2022: 17th European Conference, Tel
  Aviv, Israel, October 23--27, 2022, Proceedings, Part XXXII}, pages 156--174.
  Springer, 2022.

\bibitem{sun2022direct}
Cheng Sun, Min Sun, and Hwann-Tzong Chen.
\newblock Direct voxel grid optimization: Super-fast convergence for radiance
  fields reconstruction.
\newblock In {\em Proceedings of the IEEE/CVF Conference on Computer Vision and
  Pattern Recognition}, pages 5459--5469, 2022.

\bibitem{nerfstudio}
Matthew Tancik, Ethan Weber, Evonne Ng, Ruilong Li, Brent Yi, Justin Kerr,
  Terrance Wang, Alexander Kristoffersen, Jake Austin, Kamyar Salahi, Abhik
  Ahuja, David McAllister, and Angjoo Kanazawa.
\newblock Nerfstudio: A modular framework for neural radiance field
  development.
\newblock {\em arXiv preprint arXiv:2302.04264}, 2023.

\bibitem{thomas2019kpconv}
Hugues Thomas, Charles~R Qi, Jean-Emmanuel Deschaud, Beatriz Marcotegui,
  Fran{\c{c}}ois Goulette, and Leonidas~J Guibas.
\newblock Kpconv: Flexible and deformable convolution for point clouds.
\newblock In {\em Proceedings of the IEEE/CVF international conference on
  computer vision}, pages 6411--6420, 2019.

\bibitem{vora2021nesf}
Suhani Vora, Noha Radwan, Klaus Greff, Henning Meyer, Kyle Genova, Mehdi~SM
  Sajjadi, Etienne Pot, Andrea Tagliasacchi, and Daniel Duckworth.
\newblock Nesf: Neural semantic fields for generalizable semantic segmentation
  of 3d scenes.
\newblock {\em arXiv preprint arXiv:2111.13260}, 2021.

\bibitem{vu2022softgroup}
Thang Vu, Kookhoi Kim, Tung~M Luu, Thanh Nguyen, and Chang~D Yoo.
\newblock Softgroup for 3d instance segmentation on point clouds.
\newblock In {\em Proceedings of the IEEE/CVF Conference on Computer Vision and
  Pattern Recognition}, pages 2708--2717, 2022.

\bibitem{wang2021ibrnet}
Qianqian Wang, Zhicheng Wang, Kyle Genova, Pratul~P Srinivasan, Howard Zhou,
  Jonathan~T Barron, Ricardo Martin-Brualla, Noah Snavely, and Thomas
  Funkhouser.
\newblock Ibrnet: Learning multi-view image-based rendering.
\newblock In {\em Proceedings of the IEEE/CVF Conference on Computer Vision and
  Pattern Recognition}, pages 4690--4699, 2021.

\bibitem{xu2022point}
Qiangeng Xu, Zexiang Xu, Julien Philip, Sai Bi, Zhixin Shu, Kalyan Sunkavalli,
  and Ulrich Neumann.
\newblock Point-nerf: Point-based neural radiance fields.
\newblock In {\em Proceedings of the IEEE/CVF Conference on Computer Vision and
  Pattern Recognition}, pages 5438--5448, 2022.

\bibitem{yao2020blendedmvs}
Yao Yao, Zixin Luo, Shiwei Li, Jingyang Zhang, Yufan Ren, Lei Zhou, Tian Fang,
  and Long Quan.
\newblock Blendedmvs: A large-scale dataset for generalized multi-view stereo
  networks.
\newblock {\em Computer Vision and Pattern Recognition (CVPR)}, 2020.

\bibitem{yu2021plenoctrees}
Alex Yu, Ruilong Li, Matthew Tancik, Hao Li, Ren Ng, and Angjoo Kanazawa.
\newblock Plenoctrees for real-time rendering of neural radiance fields.
\newblock In {\em Proceedings of the IEEE/CVF International Conference on
  Computer Vision}, pages 5752--5761, 2021.

\bibitem{yu2021pixelnerf}
Alex Yu, Vickie Ye, Matthew Tancik, and Angjoo Kanazawa.
\newblock pixelnerf: Neural radiance fields from one or few images.
\newblock In {\em Proceedings of the IEEE/CVF Conference on Computer Vision and
  Pattern Recognition}, pages 4578--4587, 2021.

\bibitem{bdd100k}
Fisher Yu, Haofeng Chen, Xin Wang, Wenqi Xian, Yingying Chen, Fangchen Liu,
  Vashisht Madhavan, and Trevor Darrell.
\newblock Bdd100k: A diverse driving dataset for heterogeneous multitask
  learning.
\newblock In {\em IEEE/CVF Conference on Computer Vision and Pattern
  Recognition (CVPR)}, June 2020.

\bibitem{zhang2022nerfusion}
Xiaoshuai Zhang, Sai Bi, Kalyan Sunkavalli, Hao Su, and Zexiang Xu.
\newblock Nerfusion: Fusing radiance fields for large-scale scene
  reconstruction.
\newblock In {\em Proceedings of the IEEE/CVF Conference on Computer Vision and
  Pattern Recognition}, pages 5449--5458, 2022.

\bibitem{zhi2021place}
Shuaifeng Zhi, Tristan Laidlow, Stefan Leutenegger, and Andrew~J Davison.
\newblock In-place scene labelling and understanding with implicit scene
  representation.
\newblock In {\em Proceedings of the IEEE/CVF International Conference on
  Computer Vision}, pages 15838--15847, 2021.

\bibitem{zhou2014color}
Qian-Yi Zhou and Vladlen Koltun.
\newblock Color map optimization for 3d reconstruction with consumer depth
  cameras.
\newblock {\em ACM Transactions on Graphics (ToG)}, 33(4):1--10, 2014.

\end{thebibliography}
